\documentclass[letterpaper, twocolumn, 10pt]{article}
\usepackage[ total={7.0in, 9.5in}]{geometry}
\usepackage{moreverb,url}

\usepackage[colorlinks,bookmarksopen,bookmarksnumbered,citecolor=blue,urlcolor=blue]{hyperref}

\usepackage{import}
\usepackage{extramarks}
\usepackage[plain]{algorithm}
\usepackage{algpseudocode}
\usepackage{graphicx}
\usepackage{amsmath}
\usepackage{amsfonts}
\usepackage{amssymb}
\usepackage{wasysym}
\usepackage{nomencl}
\makeglossary
\usepackage{listings}
\usepackage{xcolor}
\usepackage{textcomp}
\usepackage{cases}
\usepackage{setspace}
\usepackage[final]{pdfpages}
\usepackage{epstopdf}
\usepackage{multicol}
\usepackage{multirow}
\usepackage{makecell}
\usepackage{fancyhdr}
\usepackage{url}
\usepackage{lscape}
\usepackage[bottom]{footmisc}
\usepackage{rotating}
\usepackage{wrapfig}
\usepackage{epsfig}
\usepackage{fmtcount}
\usepackage{booktabs}
\usepackage{longtable}
\usepackage{parskip}
\usepackage{eqnarray}
\usepackage{mdframed}
\usepackage{enumitem}
\usepackage{transparent}
\usepackage{setspace}
\usepackage{bm}
\usepackage{mathtools}
\usepackage{cuted}
\usepackage{cancel}
\usepackage{siunitx}
\usepackage[export]{adjustbox}
\usepackage{tikz,pgfplots,pgfmath,tikzscale,wrapfig}
\usepackage{pgfplotstable}
\usepackage{caption}
\usepackage{subcaption}
\usetikzlibrary{shapes.misc}
\usetikzlibrary{arrows, arrows.meta}
\usetikzlibrary{shapes.geometric,calc,positioning}
\usetikzlibrary{decorations.markings}
\pgfplotsset{compat=newest}
\usetikzlibrary{positioning}

\usepackage[thinlines]{easytable}

\colorlet{bgcolor}{white}
\usepackage{longtable,tabularx}
\definecolor{light-gray}{gray}{0.85}
\definecolor{dark-green}{rgb}{0.1328,0.5430,0.1328}

\let\vec\undefined

\newcommand{\harpoonvec}[2]{%
	\ifx\displaystyle#1\doalign{$\harpvecsign$}{#1#2}\fi
	\ifx\textstyle#1\doalign{$\harpvecsign$}{#1#2}\fi
	\ifx\scriptstyle#1\doalign{\scalebox{.6}[.9]{$\harpvecsign$}}{#1#2}\fi
	\ifx\scriptscriptstyle#1\doalign{\scalebox{.5}[.8]{$\harpvecsign$}}{#1#2}\fi
}
\newcommand{\harpvecsign}{\scriptscriptstyle\rightharpoonup}

\newcommand{\harp}[1]{\mathpalette\harpoonvec{#1}}

\newcommand{\doalign}[2]{%
	{\vbox{\offinterlineskip\ialign{\hfil##\hfil\cr#1\cr$#2$\cr}}}%
}

\newcommand{\vec}[1]{\bm{\mathrm{#1}}}
\newcommand{\uvec}[1]{\harp{\vec{#1}}}

\newcommand{\dvec}[1]{\dot{\vec{#1}}}

\newcommand{\diff}{\mathrm{d}}

\newcommand{\fddt}[2]{ \prescript{\tiny{#1}}{}{\negmedspace\frac{ \diff}{\diff t}} \left(#2\right) }

\newcommand{\xin}[2]{  #1^{#2} }
\newcommand{\E}[1]{\mathbb{E}\left[#1\right]}
\newcommand{\sk}[1]{  {[#1]^\wedge} }
\newcommand{\desk}[1]{ {\left[#1\right]^\vee}}

\newcommand{\oS}{\mathrm{S}}

\newcommand{\oO}{\mathrm{O}}

\newcommand{\oA}{\mathrm{A}}

\newcommand{\oE}{\mathrm{E}}
\newcommand{\oG}{\mathrm{G}}
\newcommand{\oI}{\mathrm{O}}

\newcommand{\fA}{\mathcal{A}}
\newcommand{\fG}{\mathcal{G}}
\newcommand{\fS}{\mathcal{S}}
\newcommand{\fI}{\mathcal{I}}
\newcommand{\fC}{\mathcal{C}}

\newcommand{\Gx}{\uvec{g}_1}
\newcommand{\Gy}{\uvec{g}_2}
\newcommand{\Gz}{\uvec{g}_3}

\definecolor{black}{RGB}{0,0,0}
\definecolor{light-gray}{gray}{0.95}
\definecolor{dark-gray}{gray}{0.20}
\definecolor{goldenrod}{RGB}{254,236,167}
\definecolor{cadetblue}{RGB}{137,136,152}
\definecolor{navyblue}{RGB}{25,114,180}
\definecolor{forestgreen}{RGB}{0,102,0}
\definecolor{pinegreen}{RGB}{18,138,114}
\definecolor{springgreen}{RGB}{198,218,110}
\definecolor{maroon}{RGB}{153,0,0}
\definecolor{brown}{RGB}{153,76,0}
\definecolor{orange}{RGB}{255,128,0}
\definecolor{denim}{RGB}{0,102,204}
\definecolor{darkblue}{RGB}{0,0,102}
\definecolor{blue}{RGB}{0,0,255}
\definecolor{indigo}{RGB}{102,0,204}
\definecolor{purple}{RGB}{204,0,204}
\definecolor{dark-yellow}{RGB}{102,102,0}
\definecolor{dark-teal}{RGB}{0,153,153}
\definecolor{green}{RGB}{0,204,0}

\newcommand{\xR}{ \textcolor{black}{R} }
\newcommand{\xQ}{ \textcolor{black}{Q} }
\newcommand{\xs}{ \textcolor{black}{\vec{s}} }

\newcommand{\xw}{ \textcolor{black}{\vec{w}} }
\newcommand{\xr}{ \textcolor{black}{\vec{r}} }
\newcommand{\xv}{ \textcolor{black}{\vec{v}} }

\newcommand{\txQk}[1]{\textcolor{black}{\tilde{Q}_{#1}}}

\newcommand{\txQki}[1]{\textcolor{black}{\tilde{Q}_{k}^{(#1)}}}

\newcommand{\txwk}[1]{\textcolor{black}{\tilde{\vec{w}}_{#1}}}

\newcommand{\txwki}[1]{\textcolor{black}{\tilde{\vec{w}}_{k}^{(#1)}}}

\newcommand{\txrk}[1]{\textcolor{black}{\tilde{\vec{r}}_{#1}}}

\newcommand{\txrki}[1]{\textcolor{black}{\tilde{\vec{r}}_{k}^{(#1)}}}

\newcommand{\txvki}[1]{\textcolor{black}{\tilde{\vec{v}}_{k}^{(#1)}}}

\newcommand{\txvk}[1]{\textcolor{black}{\tilde{\vec{v}}_{#1}}}

\DeclareRobustCommand{\homo}[1]{\protect{\underbar{$#1$}}}

\newcommand{\Exp}[1]{\exp{( \sk{#1})}}

\newcommand{\Log}[1]{\desk{\log\left(#1\right) }}

\tikzset{
	overdraw/.style={preaction={draw,bgcolor,line width=#1}},
	overdraw/.default=3pt
}

\everymath{\textstyle}

\newcommand{\aij}[2]{\mathrm{a}_{#1 #2}}
\newcommand{\bi}[1]{\mathrm{b}_{#1}}
\newcommand{\ci}[1]{\mathrm{c}_{#1}}

\newtheorem{fact}{Fact}

\usepackage{xcolor}

\title{AstroSLAM: Autonomous Monocular Navigation in the Vicinity of a Celestial Small Body - Theory and Experiments}

\author{Mehregan Dor\textsuperscript{1},
	Travis Driver\textsuperscript{1},
	Kenneth Getzandanner\textsuperscript{2} and
	Panagiotis Tsiotras\textsuperscript{1}}

\date{\textsuperscript{1}Georgia Institute of Technology, Atlanta, GA, USA \\
	\texttt{\{mehregan.dor, travisdriver, tsiotras\}@gatech.edu} \\ [1ex]
      \textsuperscript{2}NASA Goddard Space Flight Center, Greenbelt, MD, USA \\
      \texttt{kenneth.getzandanner@nasa.gov}}
\begin{document}
\maketitle


\begin{abstract}
We propose AstroSLAM, a standalone vision-based solution for autonomous online navigation around an unknown target small celestial body.
AstroSLAM is predicated on the formulation of the SLAM problem as an incrementally growing factor graph, facilitated by the use of the GTSAM library and the iSAM2 engine.
By combining sensor fusion with orbital motion priors, we achieve improved performance over a baseline SLAM solution.
We incorporate orbital motion constraints into the factor graph by devising a novel relative dynamics factor, which links the relative pose of the spacecraft to the problem of predicting trajectories stemming from the motion of the spacecraft in the vicinity of the small body.
We demonstrate the excellent performance of AstroSLAM using both real legacy mission imagery and trajectory data courtesy of NASA's Planetary Data System, as well as real in-lab imagery data generated on a 3 degree-of-freedom spacecraft simulator test-bed. 
\end{abstract}

\section{Introduction}
\label{sec:introduction}

Precise relative navigation techniques, incorporating increased levels of autonomy, will be a key enabling element of future small-body orbiter missions~\cite{christian2012onboard,delpech2015vision, nesnas2021}.
Firstly, good navigation can inform safe and efficient path planning, control execution, and maneuvering~\cite{bhaskaran2014closed}.
In near small-body missions, achieving fuel-efficiency during non-critical maneuvers and guaranteeing execution of safety-critical maneuvers requires precise knowledge of the relative position and orientation of the spacecraft with respect to the small-body.
Secondly, precise navigation situates the acquired science data. 
Indeed, scientists and mission planners design science acquisition phases based on the expected scientific value of instrument data acquired at predetermined times, on specific orbits and with specific spacecraft orientations~\cite{MILLER}.
Thirdly, precise navigation facilitates the detailed mapping and shape reconstruction of the target small-body, since good knowledge of the spacecraft relative position and orientation with respect to the target, as well as a good knowledge of the Sun light direction, are crucial in commonly used shape reconstruction solutions~\cite{gaskell2008characterizing}.
Finally, good estimates of the spacecraft state enable precise characterization of the target small-body's spin state, mass moment values and gravitational model~\cite{MILLER}.

In recent years, with ever improving navigation solutions, space missions have successfully performed daring firsts in navigation around small celestial bodies. 
Orbiter Near-Earth Asteroid Rendezvous (NEAR) Shoemaker's controlled asteroid touchdown (1996)~\cite{PROCKTER2002491}, Hayabusa I \& II's touchdown and successful sample return (2003)~\cite{yoshikawa2015hayabusa, terui2020guidance}, Dawn's orbiting of two celestial bodies in a single mission (2007)~\cite{konopliv2014vesta} and the recent Origins, Spectral Interpretation, Resource Identification, Security -- Regolith Explorer's (OSIRIS-REx's) Touch-and-Go (TAG) operation leveraging Natural Feature Tracking (NFT) relying on high navigation solution accuracy during descent~\cite{lauretta2021tag, berry2022tag}, are only few of the most notable feats accomplished thanks to autonomous navigation.

OSIRIS-REx proximity operations at the near-Earth asteroid (101955) Bennu, in particular, pushed the boundaries of what can be accomplished using primarily ground-in-the-loop navigation techniques~\cite{antreasian2022prox}.
Like similar small-body missions, OSIRIS-REx's proximity to Bennu, as well as the asteroid's small size and low gravitational attraction relative to perturbing forces, drove the need for frequent and timely navigation updates in order to achieve mission objectives~\cite{leonard2022odpred}.
These updates drove operations complexity and cadence, challenging the flight team, and heavily utilizing Deep Space Network (DSN) assets.
OSIRIS-REx proximity operations navigation and TAG also relied on detailed local and global topographic maps constructed from image and LiDAR data, with ground sample distances ranging from 75 cm down to 8 mm~\cite{barnouin2020digital}.
Building these maps required dedicated, months-long observation and data collection campaigns and a substantial amount of effort by the Altimetry Working Group on the ground, as well as multiple iterations with the navigation team~\cite{leonard2020shape}.
It also required downlinking tens of thousands of images and hundreds of gigabytes of LiDAR data from the spacecraft through the DSN.

It is recognized that the high-risk nature of missions in proximity of small celestial bodies, along with a lack of autonomy in current mission procedures, severely limits the possibilities in mission design~\cite{Starek2016}.
Indeed, ground-segment operators are intimately involved in all in-situ tasks, which ultimately rely on extensive human-in-the-loop verification, as well as ground-based computations for estimation, guidance, and control~\cite{williams2002, nesnas2021}.
In addition, long round-trip light times and severely limited bit-rate in communications render ground-in-the-loop processes extremely tedious.
In tandem, we acknowledge that the incorporation of autonomous capabilities has the potential to improve navigation performance and reduce operational complexity for future missions~\cite{getzandanner2022ll, nesnas2021}.

In this paper, we build on our previous work~\cite{dor2021visual} and present a holistic application of a factor graph-based incremental smoothing solution for monocular simultaneous localization and mapping (SLAM) around a small celestial body, referred to as \texttt{AstroSLAM}. 
Using the \texttt{GTSAM} library~\cite{dellaert2017factor} and the \texttt{iSAM2} solver~\cite{kaess2012isam2}, 
we perform multi-sensor fusion and constraint enforcement on-the-fly.
We incorporate inertial attitude measurements from a star tracker and Earth-relative DSN radiometric data to obtain an initial pose prior, and then we leverage image-based measurements and dynamical motion constraints for subsequent trajectory estimation.
Crucially, we incorporate the equations of the relative motion between the spacecraft and the small-body into the problem, given that these constraints provide a strong motion prior.
We do so by writing and implementing a new factor node called \texttt{RelDyn}, which encodes the constraints derived from the relative dynamics and kinematics.

In order to demonstrate the improvement in performance brought about by incorporating the relative motion dynamics in the SLAM problem, which usually suffers from incorrect data association in real-world applications, we test \texttt{AstroSLAM} using \textit{real-world} imagery.
%
We test our algorithm on an image sequence from NASA's Dawn mission, as well as on a sequence of images of a realistic mock-up asteroid produced in the lab. 
We compare the estimated solution against archived navigational data from Dawn for the first demonstration, and also compare the estimated solution to ground-truth data from the in-lab sequence. 
It is shown that \texttt{AstroSLAM} demonstrates excellent performance in both cases.

\subsection{Related work}
\label{sec:background}

Filter-based methods~\cite{nakath2018multi}, such as the Extended Kalman Filter (EKF), have traditionally been applied to perform on-the-fly multi-sensor fusion for precise navigation purposes.
\cite{bercovici2019} proposed a Flash-LiDAR-based pose estimation and shape reconstruction approach, by solving a maximum likelihood estimation problem via particle-swarm optimization, followed by a least-squares filter providing measurements for the spacecraft position and orientation in the small-body frame coordinates. 
Other recent works in the field have established proof-of-concepts for online implementation of batch optimization and graph-based approaches for precise near-small-body navigation, like real-time SLAM. 
Notably, \cite{nakath2019active} presents an active SLAM framework which also employs Flash-LiDAR as the base measurement of the SLAM formulation, with sensor fusion of data from an inertial measurement unit and star tracker, tested with simulated data. 
However, the limited range of Flash-LiDAR instruments restricts the spacecraft's orbit to unrealistically small radii, reducing the use scenarios to either navigation near very small small-bodies or to the touchdown phase for larger target small-bodies. 
For example, the OSIRIS-REx Guidance, Navigation, and Control (GNC) Flash-LiDAR, which is mentioned by both Nakath and Bercovici, has a reliable maximum range of approximately 1~km~\cite{church2020LiDAR, leonard2022LiDAR}. 
In contrast, an approach that uses long range optical imagery, like the one we propose in this paper, enables detailed characterization of the small-body early in the approach phase of the mission, at which point knowledge about the target small-body may still be poor.

Several prior works have applied visual SLAM solutions for spacecraft relative navigation.
However, much fewer works have directly applied visual SLAM to the small-body navigation problem.
Among the most interesting works in this area, we note \cite{cocaud2010surf}, which initially leverages SURF~\cite{bay2006surf} visual cues and range measurements, and \cite{cocaud2012autonomous}, which focuses on image feature-only formulation, and solves the relative pose estimation problem using a Rao-Blackwellized particle filter. 
However, the latter works only tested the algorithm on simulated imagery of asteroid Itokawa.
Additionnally, particle filters are notoriously compute-intensive, and not directly amenable to on-the-fly implementation.
Similarly, \cite{baldini2018autonomous} implemented \texttt{OpenSFM} on simulated images of comet 67P/Churyuomov-Gerasimenko, while
\cite{takeishi2015simultaneous} performed a particle filter minimization of the observation error and used both simulated landmarks and SIFT~\cite{lowe2004distinctive} features extracted and tracked across a sequence of real images of a simple asteroid mock-up, with albeit unrealistic motion.
Most recently, \cite{villa2022} successfully implemented an autonomous navigation and dense reconstruction predicated on visual-only landmark observation batch optimization and stereophotogrammetry.
However, the implied batch optimization methods require intensive on-board compute power, which are not easily amenable to on-the-fly autonomous navigation.

The SLAM problem can be assimilated to a discrete-time sequential state estimation and static scene mapping problem, and it is known that the inclusion of motion priors allows for some ``smoothness" to be worked into the SLAM solution.
However, the ability of the motion prior to improve the SLAM solution depends directly on the validity of the motion model in the application case and on the uncertainties associated with perturbations and un-modelled effects.

We postulate that a motion prior should indeed be incorporated into the near-small-body monocular SLAM navigation problem, but that a simplistic motion model is not enough to validate the algorithm on real data.
The motion prior should, instead, be based on a high-fidelity description of the dynamics of the spacecraft-small-body system.
Indeed, the orbital motion of the spacecraft in the vicinity of a small-body can be modelled with very high fidelity, owing to careful and accurate modelling of the perturbing forces at play.
Additionally, the orbital motion of a typical small-body in the solar system targeted for further probing is typically estimated with high precision due to tracking and Orbit Determination (OD) throughout a long period via ground observations, as well as during the approach phase via Optical Navigation (OpNav).
It follows that an accurate and highly certain relative motion model between the spacecraft and the small-body can be derived and used in formulating a \textit{strong motion prior}.
Nevertheless, parameters that affect the description of the relative position of the spacecraft with respect to the small-body, such as the true size of the small-body, the relative distance to the its center of mass, its gravitational potential, its spin state, or the forces on the spacecraft due to its albedo radiation, are not well known a-priori, and must be estimated in-situ.
For the purpose of this work, these parameters are considered to be known with reasonably low uncertainty pre-encounter.
Future work will incorporate additional measurement modalities to estimate these parameters on-the-fly 
in an autonomous fashion.

Modern SLAM solutions are predicated on the formulation of the estimation or smoothing problems 
using a \textit{factor graph}~\cite{dellaert2021factor}, which we discuss in deeper detail in Section~\ref{sec:Bayesian_description}.
As such, several works have incorporated dynamics-derived constraints into the SLAM factor graph problem. 
Most works incorporate a constant velocity with white noise motion prior~\cite{matsuzaki2000, anderson2015} or a linear dynamical model with white noise motion prior~\cite{anderson2015batch}, modelled as a Gaussian process. 
These approaches are often referred to as Simultaneous Trajectory Estimation And Mapping (STEAM), and fall within the realm of batch optimization methods, which for large problems can be compute-intensive.
Notably,~\cite{yan2017incremental} provides an extension by transforming the batch STEAM optimization into an incremental method using efficient variable re-ordering at every optimization step, while still exploiting a continuous-time Gaussian process motion prior.
In these works, the estimated Gaussian process provides a time-based support to evaluate the trajectory at any desired query time within the sampling interval.
However, the accuracy of the solution is predicated on interpolation of the estimated state and covariance between the selected optimization times using the Gaussian process model in an iterative fashion.
Therefore, it is required that the timesteps for optimization be chosen close to each other, thus increasing the computation task.
This disadvantage offsets the advantage of the use of a continuous Gaussian process-based method for the purpose of small-body navigation, since optimization times can be significantly spread apart along the trajectory around the small-body given the limited on-board resources in a real mission.
Instead, we propose to use a high fidelity dynamical model, paired with an on-manifold integration method, to obtain accurate predictions of the state with large time-steps, thus reducing the density of the selected optimization points in time.

As opposed to the aforementioned works, where the factor encodes the error of the dynamical prediction by exploiting the solution to the piecewise-constant input locally-linearized model,~\cite{xie2020factor} formulates the dynamics factor using the non-linear differential equation directly.
It follows that the resulting factor graph has ``non-state'' variables, such as linear and angular accelerations, as well as wrenches.
To avoid an underconstrained problem, these dynamics-related accelerations each need an individual measurement or prior factor.
This solution, although simple and useful for reference trajectory planning, is unrealistic for the purpose of estimation, since real-world measured acceleration and wrench signals are generally fraught with high-frequency noise and poor signal-to-noise ratio.

The factor graph-based formulation using stereo SLAM implemented on the SPHERES platforms~\cite{Tweddle2013} is most closely related to our work.
Indeed, Tweddle incorporated non-specific dynamical constraints into the problem by implementing a factor that captures the residual of the integrated equations of motion.
It is noteworthy that in the latter work, the observing agent has no inertial motion and has a static viewing direction.
In contrast, in our work, both the observing spacecraft and the target have inertial translational and rotational motion, and are subject to specific dynamics as a consequence of the driving forces and torques. 
We capture these constraints in the form of relative kinematics and relative dynamics yielding equations describing the relative motion between the spacecraft and the small-body.

Several works incorporate some form of dynamics-based modelling in SLAM for the purpose of improving the baseline SLAM solution in near-small-body navigation. 
For example, \cite{delpech2015vision} propose to perform an EKF step integrating the inertial equations of motion in between steps involving bundle adjustment for SLAM, in an alternating fashion. 
Most notably, the work by~\cite{rathinam2017}, which has important parallels to our own, incorporates orbital motion priors as factors into the SLAM smoothing problem factor graph directly, arguably for the purpose of estimating the relative pose with respect to a small-body.
Nevertheless, there are several key differences and shortcomings with respect to our work, which we detail next.

The first difference lies in the fidelity and accuracy of the modelled spacecraft-small-body system dynamics with respect to the real mission setting. 
In~\cite{rathinam2017} the inertial equations of motion of the spacecraft and of the small-body are implemented separately, whereby the spacecraft is subject to a massive central body gravitational force and the small-body is subject to only a zero-mean perturbation force.
However, it is generally known that, due to the relatively small gravitational force of the small-body, the solar gravitational force and the solar radiation pressure forces intervene in a significant way to affect the trajectory of the spacecraft~\cite{scheeres2016orbital} around the small-body, and therefore these forces should not be neglected during modelling.
This is true for most missions to small-bodies of interest in the solar system, especially smaller small-bodies, such as Itokawa, which~\cite{rathinam2017} specifically consider in their work as an illustrative example.
In addition,~\cite{rathinam2017} omit to incorporate spacecraft control forces in the modelling, as well as any spacecraft inertial attitude and angular velocity measurements. 
Such a choice severely restricts the possible use cases of their formulation.
We chose instead to model the \textit{relative} kinematics and dynamics between the spacecraft and the small-body, predicated on the fact that in the SLAM problem, without added dynamics related factors, the measurement is innately a function of the relative position and attitude of the camera (here the spacecraft) with respect to the static scene (here the rigid small-body surface) through the 3D point projection measurement function.
It then suffices to propagate the relative position and orientation in the kinematics and dynamics.
In addition, we also include spacecraft control forces, spacecraft attitude and angular velocity measurements and the above-mentioned Sun-related perturbing forces in our modelling framework.

The second difference stems from the approach to encoding the chosen dynamics in factors to be included in the factor-graph.
Specifically,~\cite{rathinam2017} encode their dynamics into two separate factors without any specific discussion about the identification or quantification of the disturbances on which the factors' residual error function is predicated.
Yet, it is obvious from their formulation that the dynamics of the spacecraft and of the small-body are actually coupled through the noise. 
Indeed, there is a link through the dynamics between the disturbance considered in the inertial motion model of the small-body and the small-body gravitational force affecting the spacecraft, which is itself a function of the relative position vector between the spacecraft and the small-body.
Hence, we postulate that the equations of motion of the spacecraft-small-body system cannot be decoupled and, consequently, should not be encoded by two separate factors.
As the quantification of the noise sources and the way they enter the equations of motion impact the feasibility of smoothing, based on the concept of \textit{smoothability}~\cite{gelb1974applied}, we expose the effect of these noise sources in a full development of the stochastic differential equations in Appendix~\ref{sec:EoM_Stochastic}. 
We discuss the factor error function in Section~\ref{sec:reldyn_factor_formulation} 
and assess their impact on smoothability in Section \ref{sec:smoothability}.

The third difference pertains to our use of a front-end system processing actual image data towards the SLAM solution, and their lack thereof.
In~\cite{rathinam2017}, randomly sampled idealized 3D points from a shape model are used to simulate camera feature point measurements.
However, this simulation fails to mimic real-life effects encountered in small-body surveying missions. 
These effects include, among others, small-body surface shadowing, landmark visibility restricted by view-cone or occlusions, local image quality variations and image blurring, etc., which all
affect the number of tracked surface features, the reliability of the matched features and, in turn, the overall error in the SLAM solution.
Crucially, the use of a real front-end system in our work allows us to quantify the real improvement brought about by the incorporation of  motion priors.
Indeed, the ability to appropriately match features directly depends on knowledge of the relative pose, which, in turn, is improved by the use of motion priors, a result we detail in Section \ref{sec:experiments}.

Finally, we use real mission imagery and trajectory data to validate our overall system, demonstrating impressive performance, further supporting our choice to model the forces missing from~\cite{rathinam2017}.

\subsection{Contributions}
\label{sec:contributions}

In contrast to the traditional ground-in-the-loop mindset, this paper proposes \texttt{AstroSLAM}, a viable autonomous navigation approach for near-small-body operations based on small celestial body imagery collected by the spacecraft on-board cameras.
Specifically, the contributions of this work are the following: (a) we formulate a precise autonomous vision-based navigation scheme based on SLAM and sensor fusion in orbit; (b) we model and incorporate orbital motion constraints specific to the small-body circumnavigation problem to make the SLAM solution more robust to outliers and drift; (c) we demonstrate our algorithm on real imagery obtained in-situ from a previously flown small-body orbiter mission; (d) we detail the elaboration of an in-lab hardware test setup designed to verify and validate the algorithm with realistic setting and accurate ground-truth capture; (e) we demonstrate our algorithm's performance using imagery and trajectory data produced in our experimental lab facility---
such a real-word validation is, to our knowledge, a first among works concerning near-small-body SLAM algorithms and constitutes a significant novelty of this paper.

This paper is organized as follows:
Section~\ref{sec:theory} introduces the problem and summarizes the notation used.
Section~\ref{sec:Bayesian_description} introduces the main mathematical tool used for estimating the unknown relative pose of the spacecraft in this paper, and
and Section~\ref{sec:incorp_of_motion_priors} focuses on the novel relative orbital motion prior calculations utilized in the factor graph estimation engine.
Details of the technical approach in terms of implementation are given in
Section~\ref{sec:methods}.
Section~\ref{sec:experiments} validates the algorithm against real asteroid imagery from prior NASA missions, and
Section~\ref{sec:valid_astros}
presents the results from experiments carried out in a realistic laboratory facility where the ground truth relative pose is available.
Lastly, Section~\ref{sec:conclusion} provides some conclusions along with possible avenues of future work.
To keep the narrative accessible, many of of the more technical derivations are given in the Appendix.

\section{Problem statement}
\label{sec:theory}

In this section, we discuss the relevant theory, and establish the problem statement pertaining to the asteroid relative navigation problem incorporating monocular SLAM, an appropriate motion prior and sensor fusion.

Firstly, we summarize the notation conventions used throughout the paper in Section~\ref{sec:notation}.
Secondly, we contextualize the defined notation within the problem of a spacecraft navigating around an asteroid in Section~\ref{subsec:asteroid_nav_defs}.

\subsection{Notation}
\label{sec:notation}

Given the affine space $(e.g., \mathbb{E}^3, \mathbb{R}^3)$,
the translation vector between any two points $\mathrm{X}, \mathrm{Y} \in \mathbb{E}^3$ is denoted $\vec{r}_{\mathrm{Y}\mathrm{X}} \triangleq \left(\mathrm{Y} - \mathrm{X}\right)  \in \mathbb{R}^3$, read ``from $\mathrm{X}$ to $\mathrm{Y}$". 
Any frame $\mathcal{X}$ is a tuple $(\mathrm{X}, \left\lbrace\uvec{x}_i\right\rbrace_{i=1}^3)$ where the point $\mathrm{X}\in \mathbb{E}^3$ denotes the origin of the frame and the set of unit directions $ \left\lbrace\uvec{x}_i \right\rbrace_{i=1}^3$, where $\uvec{x}_i \in \mathbb{S}^2,\, i=1,\ldots,3$, such that for each $i,j=1,\ldots,3,\, \uvec{x}_i \cdot \uvec{x}_j = \delta_{ij}$,  and that $\uvec{x}_i \cdot \left( \uvec{x}_j \times \uvec{x}_k \right) = \epsilon_{ijk}, (i,j,k=1,\ldots,3)$, constitutes the right-handed orthonormal basis of the frame.
The expression of any vector $\vec{v} \in \mathbb{R}^3$ in a given frame $\mathcal{X} = (\cdot,\left\lbrace\uvec{x}_i\right\rbrace_{i=1}^3)$ is denoted by $\xin{\vec{v}}{\mathcal{X}} \triangleq \begin{bmatrix} \vec{v} \cdot \uvec{x}_1 & \vec{v} \cdot \uvec{x}_2 & \vec{v} \cdot \uvec{x}_3 \end{bmatrix}^\top \in \mathbb{R}^3$. 
For any two frames $\mathcal{X} = (\cdot,\left\lbrace\uvec{x}_i\right\rbrace_{i=1}^3)$ and $\mathcal{Y} = (\cdot,\left\lbrace\uvec{y}_i\right\rbrace_{i=1}^3)$, we denote the rotation $R_{\mathcal{X}\mathcal{Y}} \triangleq \begin{bmatrix}\uvec{y}_1^\mathcal{X}& \uvec{y}_2^\mathcal{X} & \uvec{y}_3^\mathcal{X}\end{bmatrix} \in \mathrm{SO}(3)$ such that $\vec{v}^\mathcal{X} = R_{\mathcal{X}\mathcal{Y}} \vec{v}^\mathcal{Y}$ for any $\vec{v} \in \mathbb{R}^3$. 
This notation is consistent with the composition rule $R_{\mathcal{X}\mathcal{Z}} = R_{\mathcal{X}\mathcal{Y}}R_{\mathcal{Y}\mathcal{Z}}$ for any three frames $\mathcal{X}, \mathcal{Y}, \mathcal{Z}$.
Given a coordinate vector $\xin{\vec{v}}{\mathcal{X}} \in \mathbb{R}^3$ expressed in frame $\mathcal{X}$, we denote the corresponding homogeneous coordinates $\xin{\homo{\vec{v}}}{\mathcal{X}} \triangleq \begin{bmatrix}
	(\xin{\vec{v}}{\mathcal{X}})^\top & 1 
\end{bmatrix}^\top \in \mathbb{P}^3$.
For any two frames $\mathcal{X}=(\mathrm{X}, \left\lbrace\uvec{x}_i\right\rbrace_{i=1}^3)$ and $\mathcal{Y} = (\mathrm{Y}, \left\lbrace\uvec{y}_i\right\rbrace_{i=1}^3)$, we denote the homogeneous transformation $T_{\mathcal{X}\mathcal{Y}} \in \mathrm{SE}(3)$ by
\begin{align}
	T_{\mathcal{X}\mathcal{Y}} \triangleq \begin{bmatrix}
		R_{\mathcal{X}\mathcal{Y}} & \vec{r}_{\mathrm{Y}\mathrm{X}}^\mathcal{X} \\ 0_{1\times 3} & 1
	\end{bmatrix},
\end{align}
such that $\homo{\vec{r}}_{\mathrm{P}\mathrm{X}}^\mathcal{X} = T_{\mathcal{X}\mathcal{Y}} \homo{\vec{r}}_{\mathrm{P}\mathrm{Y}}^\mathcal{Y}$ for any point $\mathrm{P} \in \mathbb{E}^3$. This notation is consistent with the composition rule $T_{\mathcal{X}\mathcal{Z}} = T_{\mathcal{X}\mathcal{Y}} T_{\mathcal{Y}\mathcal{Z}}$ for any three frames $\mathcal{X},\mathcal{Y},\mathcal{Z}$.

Given a Lie algebra $\mathfrak{g}$ of dimension $n$ associated (at the identity) to a matrix Lie group $\mathfrak{G}$, we denote the \textit{hat} operator $\sk{\cdot} : \mathbb{R}^n \rightarrow \mathfrak{g}$ which maps any $n$-vector $\vec{x} = \begin{bmatrix}x_1 & \ldots & x_n\end{bmatrix}^\top \in \mathbb{R}^n$ to an element $\sk{\vec{x}}\in \mathfrak{g}$, by $\sk{\vec{x}} = \sum_{i=1}^{n} x_i E_i$, where $E_i, \, i=1,\ldots,n$ are the basis vectors of the matrix Lie algebra $\mathfrak{g}(n)$~\cite{chirikjian2011stochastic}. 
We denote its inverse \textit{vee} operator $\desk{\cdot} : \mathfrak{g} \rightarrow \mathbb{R}^n$, which extracts the coordinates $\vec{x} = \begin{bmatrix}x_1 & \ldots & x_n\end{bmatrix}^\top$ from $\sk{\vec{x}} = \sum_{i=1}^{n} x_i E_i \in \mathfrak{g}(n)$ in terms of $E_i$.
We denote the exponential map $\exp : \mathfrak{g} \rightarrow \mathfrak{G}$, mapping an element $\sk{\vec{x}} \in \mathfrak{g}$ to the element $\Exp{\vec{x}} \in \mathfrak{G}$ in the neighborhood of the identity element.
Specifically, for the $\mathrm{SO}(3)$ group of rotations, we denote the exponential map at the identity $\exp : \mathrm{so}(3) \rightarrow \mathrm{SO}(3)$ which associates any tangent vector (skew-symmetric matrix) $\sk{\vec{x}}\in \mathrm{so}(3)$ to a 3D rotation as the matrix exponential, given by
\begin{align}
	&\Exp{\vec{x}} \nonumber \\
	&=  \mathrm{I}_3 + \sk{\vec{x}} + \frac{1}{2}\left(\sk{\vec{x}}\right)^2 + \frac{1}{6}\left(\sk{\vec{x}}\right)^3 + \ldots \nonumber \\
	&= \mathrm{I}_3 + \frac{\sin \|\vec{x}\|}{\|\vec{x}\|} \sk{\vec{x}} + \frac{1-\cos \|\vec{x}\|}{\|\vec{x}\|^2}     \left(\sk{\vec{x}}\right)^2.
	\label{eq:SO3_expmap}
\end{align}
Finally, we denote the logarithm map (at the identity) by $\log : \mathfrak{G} \rightarrow \mathfrak{g}$, which maps an element in the neighborhood of the identity element of group $\mathfrak{G}$ to an element in the associated Lie algebra $\mathfrak{g}$.
In the case when $\mathfrak{G}= \mathrm{SO}(3)$ and $\mathfrak{g} =\mathrm{so}(3) $, the logarithm map $\log: \mathrm{SO}(3)\rightarrow \mathrm{so}(3)$ is the inverse of the operation given in (\ref{eq:SO3_expmap}), and is a bijective mapping as long as $\|\vec{x}\| < \pi$.

The following facts are useful in deriving first-order approximations for the effect of noises in the stochastic differential equations presented in 
Appendix~\ref{sec:EoM_Stochastic}.
Note that Fact \ref{fact1} is a consequence of the rotational invariance of the cross product in $\mathbb{R}^3$, and its simple proof is left to the reader.
\begin{fact}
	Given the hat operator $\sk{\cdot} : \mathbb{R}^3 \rightarrow \mathrm{so}(3)$, for any $R\in \mathrm{SO}(3)$, and coordinate vector $\vec{x}\in \mathbb{R}^3$, 
	\begin{enumerate}
		\item $R\sk{\vec{x}}R^\top = \sk{R\vec{x}}$,
		\item $R\sk{\vec{x}} = \sk{R\vec{x}}R$,
		\item $\sk{\vec{x}}R = R\sk{R^\top\vec{x}}$.
	\end{enumerate}
	\label{fact1}
\end{fact}

\begin{fact}
	Given the exponential map $\exp : \mathrm{so}(3) \rightarrow \mathrm{SO}(3)$, for any $R\in \mathrm{SO}(3)$ and any coordinate vector $\vec{x} \in \mathbb{R}^3$, 
	\begin{enumerate}
		\item $R\Exp{\vec{x}}R^\top = \Exp{R\vec{x}}$,
		\item $R\Exp{\vec{x}} = \Exp{R\vec{x}}R$,
		\item $\Exp{\vec{x}}R = R\Exp{R^\top\vec{x}}$.
	\end{enumerate}
	\label{fact2}
	
	Note that Fact~\ref{fact2} is a direct consequence of Fact~\ref{fact1} applied to the matrix exponential expansion in (\ref{eq:SO3_expmap}).
	
\end{fact}
\begin{fact}
	Given the exponential map $\exp : \mathrm{so}(3) \rightarrow \mathrm{SO}(3)$, for any vector $\vec{x} \in \mathbb{R}^3$ having sufficiently small length, it follows that
	\begin{align}
		\Exp{\vec{x}} \simeq \mathrm{I}_3 + \sk{\vec{x}}.
	\end{align}
	\label{fact3}
\end{fact}

\subsection{Problem definitions}
\label{subsec:asteroid_nav_defs}

Let $\oA, \oS, \oI \in \mathbb{E}^3$ and assume that the point $\oA$ corresponds to the center of mass of the asteroid body, $\oS$ corresponds to the center of mass of the spacecraft, and $\oI$ corresponds to an inertial point in space, e.g., the barycenter of the solar system. 
Initially, we distinguish three frames of interest: the inertial frame $\fI \triangleq ( \oI, \left\lbrace\uvec{n}_i\right\rbrace_{i=1}^3)$, the asteroid principal axis frame $\fA \triangleq ( \oA, \left\lbrace\uvec{a}_i\right\rbrace_{i=1}^3)$, and the spacecraft body-fixed frame $\fS \triangleq ( \oS, \left\lbrace\uvec{s}_i\right\rbrace_{i=1}^3)$.
Notice that the definition of the frame $\fA$ is based on the a-priori unknown mass moments of the asteroid, i.e., its center of mass and the principal axes of its inertia tensor.
It is therefore necessary to initially consider an arbitrarily chosen asteroid-fixed frame $\fG \triangleq ( \oG, \left\lbrace\uvec{g}_i\right\rbrace_{i=1}^3)$, and  to then estimate the transformation $T_{\fG\fA} \in \mathrm{SE}(3)$ from frame $\fA$ to frame $\fG$.
Note that since frame $\fG$ is fixed with respect to asteroid frame $\fA$, the transformation $T_{\fG\fA}$ is fixed.
An estimate of $T_{\fG\fA}$ is obtained as a result of detailed analysis of the asteroid's shape~\cite{gaskell2008characterizing, driver2020carving}, typically performed at a later stage, and is therefore not a subject of study in this paper.
\begin{figure}[th]
	\centering
	\def\Elevation{30} 
	\def\Azimuth{125} 
	\includegraphics[width=3.4in, clip=true, trim=0em 0em 0em 1.5em]{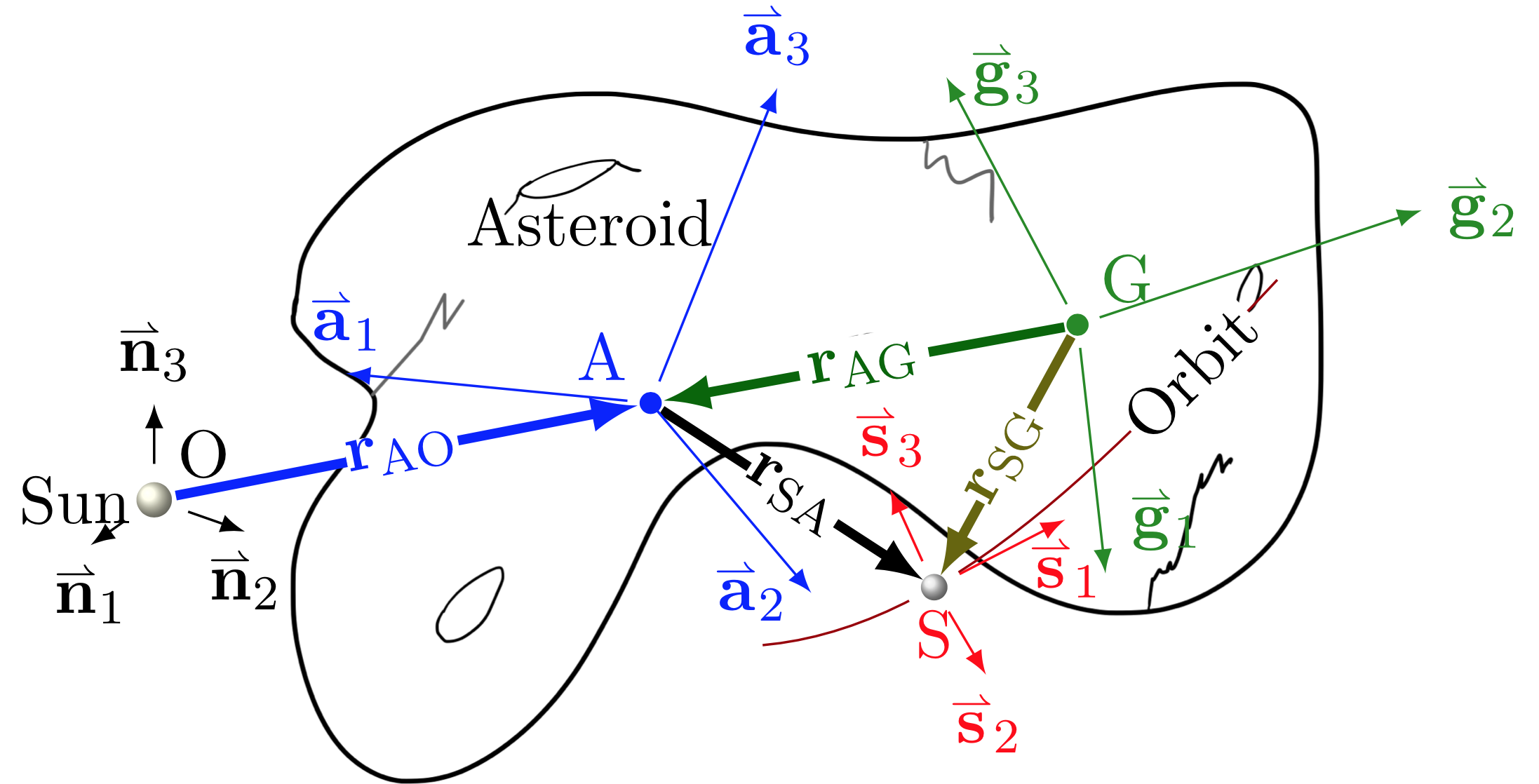}
	\caption{Relative navigation problem frame definitions and vector quantities.}
	\vspace{-1.0em}
\end{figure}
Any reference hereafter to the \textit{relative pose of the spacecraft}, for a given spacecraft frame $\fS = (\mathrm{S}, \left\lbrace\uvec{s}_i\right\rbrace_{i=1}^3)$, designates the transformation
\begin{align}
	T_{\fG\fS} = \begin{bmatrix}
		R_{\fG\fS} & \vec{r}_{\oS\oG}^{\fG} \\
		0_{1\times 3} & 1
	\end{bmatrix},
	\label{eq:SE3_def}
\end{align}
which encodes the relative rotation $R_{\fG\fS} = \begin{bmatrix}\uvec{s}_{1}^\fG & \uvec{s}_{2}^\fG & \uvec{s}_{3}^\fG \end{bmatrix}^\top \in \mathrm{SO}(3)$ of the spacecraft with respect to the $\fG$ frame and the coordinates $\vec{r}_{\oS\oG}^{\fG} = \begin{bmatrix} \vec{r}_{\oS\oG} \cdot \Gx & \vec{r}_{\oS\oG} \cdot \Gy & \vec{r}_{\oS\oG} \cdot \Gz \end{bmatrix} \in \mathbb{R}^3$ of the spacecraft position vector relative to the point $\oG$ as expressed in the $\fG$ frame.

Let ${t}_0$ be an initial time, let $t\geq {t}_0$,
let $(t_k)_{k=0}^n \subset [{t}_0, t]$ be the sequence of sensor acquisition times, and let $T_{\fG_k\fS_k} \triangleq T_{\fG\fS}(t_k)$ describe the pose of the spacecraft as expressed in the $\fG$ frame at time index $0\leq k \leq n$. 
Then, the sequence $( T_{\fG_k\fS_k})_{k=0}^n \in \Pi_{k=1}^{n} \mathrm{SE}(3)$ describes the discrete trajectory of the relative pose of the spacecraft.
We define the camera sensor frame $\fC = (\oS, \left\lbrace\uvec{c}_i\right\rbrace_{i=1}^3)$, with fixed pose $T_{\fS\fC}$ with respect to the spacecraft frame $\fS$, and obtain the sequence $\left(T_{\fG_k\fC_k}\right)_{k=1}^n$ of all camera poses, also known as \textit{frames}, where the relative camera pose is $T_{\fG_k\fC_k} = T_{\fG_k\fS_k}T_{\fS_k\fC_k} =  T_{\fG_k\fS_k}T_{\fS\fC}, \, (k=0,\ldots,n)$.

A \textit{landmark} $\mathrm{L} \in \mathbb{E}^3$ is defined as a notable 3D point in the scene, which is potentially triangulated during SLAM using camera observations.
We denote by $\Psi_k = \left\lbrace \mathrm{L}_i \in \mathbb{E}^3, \, i=1,\ldots,m_k \right\rbrace$ the set of all landmarks accumulated up until time index $ k = 0, \ldots, n$, also called the \textit{map} at time index $k$.
To each landmark $\mathrm{L} \in \Psi_n$ corresponds a position vector $\vec{r}_{\mathrm{L} \oG} \in \mathbb{R}^3$, whose coordinates in the $\fG$ frame, denoted $\vec{r}_{\mathrm{L} \oG}^{\fG}$, are fixed since the asteroid is presumed to be a rigid body.
Let $r_k$ be the total number of image feature points detected in the camera image captured at time $t_k$ for $k=0,\ldots,n$. 
We collect all detected feature points in the set $\Upsilon_k = \left\lbrace \mathrm{P}_{i} \in \mathbb{P}^2, \, i=1,\ldots,r_k\right\rbrace$, and to each $\mathrm{P}_i \in \Upsilon_k$ are associated the ideal 2D image coordinates $\vec{y}_{ik} \in \mathbb{R}^2$.

Assume a point $\mathrm{P}_i \in \Upsilon_k$ corresponds to the image projection of a scene landmark $\mathrm{L}_j \in \Psi_k$, as captured at time index $k$.
The 2D image coordinates $\vec{y}_{ik}$ relate to the 3D position coordinates $\vec{r}_{\mathrm{L}_j \oG}^{\fG}$ through the pinhole camera model relationship, given by
\begin{align}
	\begin{bmatrix}
		\lambda\vec{y}_{ik} \\ \lambda
	\end{bmatrix}
	= 
	\mathrm{K} T_{\fC_k\fG_k} \homo{\vec{r}}_{\mathrm{L}_j \oG}^{\fG},
	\label{eq:landmark_projection}
\end{align}
where $\lambda > 0 $ is a scaling factor, and where
\begin{align*}
	\mathrm{K} \triangleq \begin{bmatrix}
		\mathrm{f}_x & 0   & \mathrm{c}_x & 0 \\ 
		0   & \mathrm{f}_y & \mathrm{c}_y & 0\\
		0   & 0   & 1 & 0
	\end{bmatrix}
\end{align*}
is the camera intrinsic matrix, with $\mathrm{f}_x, \mathrm{f}_y, \mathrm{c}_x, \mathrm{c}_y$ scalars corresponding to the known camera focal lengths and optical center offsets along the two image dimensions.
The real measured feature point coordinates $\vec{y}_{ik}^{\mathrm{m}}$ are then defined such that 
\begin{align}
	\vec{y}_{ik}^{\mathrm{m}} = \vec{y}_{ik} + \vec{\nu}_{\vec{y}},
\end{align}
where $\vec{\nu}_{\vec{y}} \sim \mathcal{N}(\vec{0}, \Sigma_{\vec{y}}^{\mathrm{m}})$ corresponds to the feature point measurement noise, with associated $2 \times 2$ covariance matrix $\Sigma_{\vec{y}}^{\mathrm{m}}$.
We collect all real camera frame-landmark observations up until time index $k=0,\ldots,n$ in the set $\mathcal{Y}_k \triangleq \left\lbrace \vec{y}_{ij}^{\mathrm{m}} \in \mathbb{R}^2 : i=1,\ldots,r_j, j=0,\ldots,k \right\rbrace$.
We also define $\mathcal{X}_k \triangleq \left\lbrace T_{\fG_i \fS_i} \in \mathrm{SE}(3) : i=0,\ldots,k \right\rbrace$ as the set of all possible spacecraft relative poses discretized at times $(t_k)_{k=0}^{n}$, and define $\mathcal{L}_k \triangleq \left\lbrace \vec{r}_{\mathrm{L}\oG}^{\fG} \in \mathbb{R}^3 :  \mathrm{L} \in \Psi_k\right\rbrace$ as the set of all landmark coordinates mapped up until time index $k=0,\ldots,n$.

By exploiting the multi-view geometry constraints derived by capturing observations $\mathcal{Y}_k$ of the landmarks $\Psi_k$ at poses $\mathcal{X}_k$, as well as the constraints derived from the intrinsic motion of the spacecraft around the asteroid and other sensor measurements, further detailed in 
Section~\ref{sec:incorp_of_motion_priors},
we wish to find a solution to the trajectory $\Xi_k \in \mathcal{X}_k$ along with the set of mapped landmark coordinates $\Lambda_k \in \mathcal{L}_k$, on-the-fly, for $k=0,\ldots,n$ in a sequential and incremental manner. 
We detail the method employed to solve this problem in Section~\ref{sec:methods}.
First, we provide a brief primer on the Bayesian estimation, the SLAM problem formulated in the Bayesian framework, and the relevance of representing the structure of the problem using a factor graph.

\section{Bayesian estimation via factor graphs}
\label{sec:Bayesian_description}

In this section, we discuss the general framework of probabilistic inference applied to the incremental SLAM and sensor fusion problem for the purpose of asteroid circumnavigation. 

In modern renditions of SLAM~\cite{dellaert2017factor}, the problem stated in Section~\ref{subsec:asteroid_nav_defs} is formulated using a probabilistic inference framework, predicated on Bayesian estimation, which we evoke for our solution in this section.
The use of the factor graph formulation for modelling the asteroid navigation problem is motivated by the fact that the typical Bayesian estimation method used to solve the SLAM estimation problem is amenable to graph representation due to sparsity in the structure of the cost.
By modelling $\mathcal{X}_k$, $\mathcal{L}_k$ and $\mathcal{Y}_k$ at time index $k$  as random variables, we seek the Maximum A-posteriori Probability (MAP) solution, which maximizes the posterior probability of the spacecraft pose trajectories $\mathcal{X}_k$ and map landmark coordinates $\mathcal{L}_k$, given the known observations $\mathcal{Y}_k$, for $k=0,\ldots,n$.

To establish this solution, at any time step $k=0,\ldots,n$, we first define the conditional probability distribution $p(\mathcal{X}_k, \mathcal{L}_k|\mathcal{Y}_k)$, named the \textit{posterior}, as $p(\mathcal{X}_k, \mathcal{L}_k|\mathcal{Y}_k) \triangleq p(\mathcal{X}_k, \mathcal{L}_k,\mathcal{Y}_k)/ p(\mathcal{Y}_k)$, where $p(\mathcal{X}_k, \mathcal{L}_k,\mathcal{Y}_k)$ is a joint probability distribution on the set of all trajectories $\mathcal{X}_k$, mapped landmark coordinates $\mathcal{L}_k$ and observations $\mathcal{Y}_k$, and $p(\mathcal{Y}_k)$ is a prior probability distribution on observations $\mathcal{Y}_k$. 
Typically, in the format presented above, the posterior cannot be directly computed, since the distribution $p(\mathcal{X}_k, \mathcal{L}_k,\mathcal{Y}_k)$ is not known.
Instead, we apply Bayes' rule stated as $p(\mathcal{X}_k, \mathcal{L}_k,\mathcal{Y}_k) = p(\mathcal{Y}_k|\mathcal{X}_k, \mathcal{L}_k)p(\mathcal{X}_k, \mathcal{L}_k)$, so that we can equivalently write the posterior as $p(\mathcal{X}_k, \mathcal{L}_k|\mathcal{Y}_k) = p(\mathcal{Y}_k|\mathcal{X}_k, \mathcal{L}_k)p(\mathcal{X}_k, \mathcal{L}_k)/ p(\mathcal{Y}_k)$,
where the prior distribution $p(\mathcal{Y}_k)$ can be explicitly computed by performing the marginalization $p(\mathcal{Y}_k) = \int_{ \mathcal{L}_k}\int_{\mathcal{X}_k} p(\mathcal{Y}_k|\Xi,\Lambda)p(\Xi,\Lambda) \,\mathrm{d}\Xi \,\mathrm{d}\Lambda$ when tractable.
However, since the observations are \textit{given} in the SLAM problem, and the computation of the marginalization to obtain the denominator is generally not tractable, we can drop $p(\mathcal{Y}_k)$ to yield the proportionality relationship given by $p(\mathcal{X}_k|\mathcal{Y}_k) \propto p(\mathcal{Y}_k|\mathcal{X}_k, \mathcal{L}_{k})p(\mathcal{X}_k, \mathcal{L}_{k})$.
Independence relationships between observations, landmarks and poses, if they exist in the problem, can further be exploited to pick subsets of poses $\mathcal{X}_{ki} \subset \mathcal{X}_k$, subsets of landmarks $\mathcal{L}_{ki} \in \mathcal{L}_{k}$ and subsets of observations $\mathcal{Y}_{ki} \subset \mathcal{Y}_k$ such that the conditional probabilities $p(\mathcal{Y}_k|\mathcal{X}_k,\mathcal{L}_{k})$ and $p(\mathcal{X}_k,\mathcal{L}_{k})$ are factorized to yield $p(\mathcal{Y}_k|\mathcal{X}_k, \mathcal{L}_{k})p(\mathcal{X}_k, \mathcal{L}_{k}) = \prod_{i} p(\mathcal{Y}_{ki}|\mathcal{X}_{ki}, \mathcal{L}_{ki})p(\mathcal{X}_{ki})p(\mathcal{L}_{ki})$, while $\bigcup_i \mathcal{X}_{ki} = \mathcal{X}_k$, $\bigcup_i \mathcal{L}_{ki} = \mathcal{L}_k$ and $\bigcup_i \mathcal{Y}_{ki} = \mathcal{Y}_k$. 
By defining the likelihood function $\ell$ such that $\ell(\Xi_k, \lambda_k; \mathcal{Y}_k) \propto p(\mathcal{Y}_k|\Xi_k, \Lambda_k), \, \Xi_k \in \mathcal{X}_k,\, \Lambda_k \in \mathcal{L}_k$, we can now write the MAP solution pair $(\Xi_k^{*}, \Lambda_k^{*}) \in \mathcal{X}_k \times \mathcal{L}_k$ as the solution to the optimization problem given by
\begin{align}
	\max_{ \Xi_k \in \mathcal{X}_{k} , \Lambda_k \in \mathcal{L}_{k} } \prod_{i} \ell(\Xi_{ki}, \Lambda_{ki}; \mathcal{Y}_{ki})p(\Xi_{ki}),
	\label{eq:MAP_likelihood}
\end{align}
where $\bigcup_i \Xi_{ki} = \Xi_k$ and $\bigcup_i \Lambda_{ki} = \Lambda_k$.
Probabilistic graphical models, and specifically factor graphs, can be used to make explicit the factorization of the posterior described in (\ref{eq:MAP_likelihood})~\cite{koller2009probabilistic}.

Factor graphs are un-directed bi-partite probabilistic graphical models constituted of \textit{factor} nodes and \textit{variable} nodes, with \textit{edges} connecting variable-factor pairs~\cite{dellaert2021factor}.
The structure captured by the edges and nodes of the factor graph encodes the structure of the estimation problem's posterior probability density function, by exploiting the fact that the latter can be factorized as a product of many functions, each depending on a subset of the variables of the problem, as discussed above.
We define the factor functions $\phi_i : \mathcal{X}_{ki} \times \mathcal{L}_{ki} \rightarrow \mathbb{R}$, where $\left\lbrace\mathcal{X}_{ki}\right\rbrace, \left\lbrace\mathcal{L}_{ki}\right\rbrace$ are subsets solely specified by the variable adjacency $\mathcal{N}(\phi_i)$ of each factor in $\left\lbrace\phi_i\right\rbrace$, as encoded in the graph by factor-variable edges.
Consequently,  (\ref{eq:MAP_likelihood}) can be re-written as
\begin{align*}
	\Xi_k^{*}, \Lambda_k^{*} = \arg \max_{\mathcal{X}_{k}, \mathcal{L}_{k}} \prod_{i} \phi_i(\Xi_{ki}, \Lambda_{ki}),
\end{align*}
Each factor function thus encodes, up to a scalar, the value of an associated likelihood or prior distribution function on a subset of the problem variables.

By exploiting sparsity in the structure of the joint density function, the factor graph formulation can render very large estimation problems tractable in terms of computation.
Specifically, the factor graph, in contrast to a Bayes net, provides a framework for efficient variable elimination by first abstracting away causality.
Factor functions can be derived and emplaced in the factor graph based on the \textit{problem-specific constraints} which we wish to include.
Next, in Section~\ref{sec:incorp_of_motion_priors}, 
we devise the factor which serves as a motion prior in the asteroid circumnavigation problem.

\section{Relative orbit motion priors}
\label{sec:incorp_of_motion_priors}

In this section, we detail the theoretical approach for devising the \texttt{RelDyn} factor, which enforces a strong motion prior, namely, the equations governing the motion of the spacecraft relative to the small-body, and which incorporates fusion of other sensor measurements. 
We also discuss the relevant sources of noise and the propagated system of equations.

Similarly to~\cite{dong2016motion}, which implements STEAM using a Gaussian Process with a linear time-varying stochastic differential equation, we derive a relative dynamics factor based on the moments of the distribution describing the dispersion of the solution realizations of a system of nonlinear stochastic differential equations, evaluated at discrete times, for the purpose of improving the performance of baseline SLAM on small-body imagery.

\subsection{The \emph{\texttt{RelDyn}} factor formulation}
\label{sec:reldyn_factor_formulation}

Given a sequence of spacecraft states $\Xi_n = (x_k)_{k=0}^{n}$ at discrete times $(t_k)_{k=0}^{n}$, for $k=1,\ldots,n$, we want to devise factors which relate to the motion prior distribution $p(x_{k}, x_{k+1})$.
Note that, using Bayes' rule, the joint distribution $p(x_{k}, x_{k+1})$ can be rewritten as either $p(x_{k}, x_{k+1}) = p(x_{k+1}|x_k)p(x_k)$ or $p(x_{k}, x_{k+1}) = p(x_{k}|x_{k+1})p(x_{k+1})$, depending on which prior, $p(x_{k})$ or $p(x_{k+1})$, is readily available at the time of computation.
In our case, we typically know the state of the spacecraft and the distribution of the uncertainty of that state at the beginning of a segment, and therefore we pick $p(x_0) = \mathcal{N}(\hat{x}_0, \Sigma_0)$ as a known prior, where $\hat{x}_0$ is the known state mean and $\Sigma_0$ is the known state covariance.
We associate the factor $\phi_0^{\mathrm{prior}}(x_0)$ to the prior $p(x_0)$ accordingly.
Now, the \texttt{RelDyn} factor $\phi_k^{\mathrm{RelDyn}}$
can be formulated as
\begin{align}
	\phi_k^{\mathrm{RelDyn}} (x_k, x_{k+1}) \propto p(x_{k+1}|x_k) = \mathcal{N}(\vec{0},P_k),
	\label{eq:def_reldyn_factor}
\end{align}
where $P_k$ is a covariance matrix derived from the propagation of the moments of the distribution describing the realizations of the stochastic differential equations.

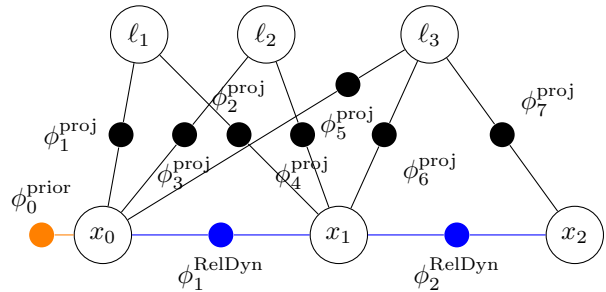
\begin{figure}[ht]
	\centering
	\begin{tikzpicture}[xscale=0.95, yscale=0.7]
		\node[shape=circle, draw=black, minimum size = 0.3in] (T0) at (-1.8 in,0 in) {$x_0$};
		
		\node[shape=circle, draw=black, minimum size = 0.3in] (T1) at (-0.5 in,0 in) {$x_1$};
		\node[shape=circle, draw=black, minimum size = 0.3in, text=black] (T2) at (0.8 in,0 in) {$x_2$};
		
		\node[shape=circle, draw=black, minimum size = 0.3in] (x2) at (-1.6in , 1.5 in) {$\ell_1$};
		\node[shape=circle, draw=black, minimum size = 0.3in] (x3) at (-0.9 in, 1.5 in) {$\ell_2$};
		\node[shape=circle, draw=black, minimum size = 0.3in, text=black] (x4) at (0.0 in, 1.5 in) {$\ell_3$};
		
		\node[shape=circle, fill=black, label=left:$\phi_1^{\text{proj}}$] (m1) at ($ 0.5*(T0) + 0.5*(x2) $ ) {};
		\node[shape=circle, fill=black, label=above:$\phi_2^{\text{proj}}$] (m2) at ($ 0.5*(x2) + 0.5*(T1) $ ) {};
		\node[shape=circle, fill=black, label=below :$\phi_3^{\text{proj}}$] (m3) at ($ 0.5*(T0) + 0.5*(x3) $ ) {};
		\node[shape=circle, fill=black,, label=below:$\phi_5^{\text{proj}}$] (m5) at ($ 0.75*(x4) + 0.25*(T0) $ ) {};
		\node[shape=circle, fill=black, label=below:$\phi_4^{\text{proj}}$] (m4) at ($ 0.5*(x3) + 0.5*(T1) $ ) {};
		\node[shape=circle, fill=black, label=below right:$\phi_6^{\text{proj}}$] (m6) at ($ 0.5*(x4) + 0.5*(T1) $ ) {};
		\node[shape=circle, fill=black, label=above right:$\phi_7^{\text{proj}}$] (m7) at ($ 0.5*(x4) + 0.5*(T2) $ ) {};
		
		\node[shape=circle, fill=orange, label=above:$\phi_0^{\mathrm{prior}}$, left=0.1in of T0] (pT0) {};
		
		\node[shape=circle, fill=blue, label=below:$\phi^{\text{RelDyn}}_1$] (f1) at ($ 0.5*(T0) + 0.5*(T1) + (0in, 0.0in)$ ) {};
		
		\path[-,draw=blue] (f1) edge (T0);
		
		\path[-,draw=blue] (f1) edge (T1);
		
		\node[shape=circle, fill=blue, label=below:$\phi^{\text{RelDyn}}_2$] (f2) at ($ 0.5*(T1) + 0.5*(T2) + (0in, 0.0in)$ ) {};
		
		\path[-,draw=blue] (f2) edge (T1);
		\path[-,draw=blue] (f2) edge (T2);

		\path[-, draw=orange] (pT0) edge (T0);
		\path[-] (x2) edge (m1);
		\path[-] (x3) edge (m3);
		\path[-] (x2) edge (m2);
		
		\path[-,draw=black] (m1) edge (T0);
		\path[-,draw=black] (m3) edge (T0);
		\path[draw=black,-] (m5) edge (T0);
		\path[draw=black,-] (m5) edge (x4);
		\path[-] (m2) edge (T1);
		\path[-] (m4) edge (T1);
		\path[draw=black,-] (m6) edge (T1);
		\path[draw=black,-] (x4) edge (m7);
		\path[draw=black,-] (m7) edge (T2);
		\path[-] (x3) edge (m4);
		\path[draw=black,-] (x4) edge (m6);
		
		
		
	\end{tikzpicture}
	\caption{A factor graph encoding the SLAM problem and the \texttt{RelDyn} motion priors}
	\label{fig:factor_graph_RelDyn}
\end{figure}

Along with projection factors, denoted $\phi_{i}^{\mathrm{proj}}, \, i = 1,\ldots,r_k$, a factor graph of the SLAM problem with motion priors can be constructed, as conceptually illustrated in Figure~\ref{fig:factor_graph_RelDyn}.
In light of this goal, a brief discussion of the \texttt{RelDyn} factor is provided throughout this section.
Additionally, the reader is directed to the Appendix, where we provide further details of the formulation.
Specifically, we derive the equations of the relative motion of the spacecraft with respect to the small-body in Appendix~\ref{sec:EoM}, by detailing the kinematics and dynamics of the system, with the assumptions regarding the modeling of the forces acting on the spacecraft and the small-body.
We continue by deriving the subsequent continuous stochastic differential equations in Appendix~\ref{sec:EoM_Stochastic} by identifying the sources of noise in the modeling due to relevant perturbations and sensor input uncertainties.

To obtain the relevant kinematics and dynamics of the spacecraft-small-body system, as expressed in the $\fG$-frame coordinates, we collect equations~(\ref{eq:relativeRotation_kinematics2}),~(\ref{eq:relativeVelocity_in_G}),~(\ref{eq:newtoneuler_asteroid_in_G}),~(\ref{eq:relativeTransDyn2_in_G}), as derived  Appendix~\ref{sec:EoM}, to produce a system of four equations of motion.
Additionally, for the sake of readability, we hereafter define the shorthand notation $\xQ \triangleq R_{\fG\fS}$, $\xw \triangleq \xin{\vec{\omega}_{\fG\fI}}{\fG}$, $\xr \triangleq \xin{\vec{r}_{\oS\oG}}{\fG}$, $\xv \triangleq \xin{\vec{v}_{\oS\oG}}{\fG}$, $\xR \triangleq R_{\fI\fS}$, $\xs \triangleq \xin{\vec{\omega}_{\fS\fI}}{\fS}$, $\vec{d} \triangleq \xin{\vec{r}_{\oA\oI}}{\fI}$, $\vec{c} \triangleq \xin{\vec{r}_{\oA\oG}}{\fG}$, $C \triangleq R_{\fG\fA}$, $A \triangleq {}^{\oA}_{}\xin{J}{\fA}_a$, $\vec{\tau} \triangleq \xin{\vec{\tau}_{a}}{\fG}$, $\vec{g}(\vec{d}) \triangleq 1/m_s \xin{\vec{F}_{\mathrm{SRP}}}{\fI}(\vec{d})$, $\vec{f} \triangleq 1/m_s \xin{\vec{F}_s}{\fS}$, which allow us to restate the equations of motion as 
\begin{align}
	\dot{\xQ} &= \xQ \sk{\xs} - \sk{\xw}\xQ, \label{eq:dQdt}\\
	\dot{\xw} &= M^{-1} \left(\vec{\tau} - \sk{\xw}  M \xw\right), \label{eq:dwdt}\\
	\dot{\xr} &= \xv - \sk{\xw}\xr, \label{eq:drdt}\\
	\dot{\xv} &= \left( \sk{\xw}\sk{\xw} +\sk{M^{-1}  \left(\vec{\tau} - \sk{\xw}  M \xw\right)}  \right) \vec{c}  \nonumber \\
	& - \sk{\xw}  \xv  - \left(\frac{\mu_a}{\|\xr - \vec{c}\|^3} +\frac{\mu_{\odot}}{\|\vec{d}\|^3}\right)\left(\xr-\vec{c}\right) \nonumber \\  
	&+ \xQ\xR^\top \vec{g}\left(\vec{d}\right) + \xQ\vec{f}, \label{eq:dvdt}
\end{align}
with $M\triangleq C A C^\top$.
Recall from Section~\ref{sec:Bayesian_description} that the \texttt{RelDyn} factor function encodes the motion prior by means of a likelihood or prior distribution function predicated on the state and the known accumulated observations at discrete times. 
Therefore, we first need to identify and quantify the probability distribution which relates to the dispersion of the state evolving over time. 
We wish to be able to evaluate this distribution at a sequence of a priori unknown discrete times along the trajectory.

We assume that the dispersion of the state at any given time is due to the accumulated effect of exogenous perturbations on the trajectory across time.
Specifically, we wish to exploit the stochastic form of the equations of motion to derive the distribution of trajectories as a function of time.
Furthermore, to be able to evaluate the distribution at any desired time, we require the stochastic differential equations to be in continuous form.
Finally, we desire to obtain the relevant stochastic differential equations directly from the equations of motion, as further detailed in
Appendix~\ref{sec:EoM}, by admitting realistic and physical perturbations through the input channels.

To this end,
let the 3-dimensional white noise Gaussian process $\vec{\nu}_{*}(t)$ (where ${*}=R,\vec{s},\vec{\tau},\vec{f}$), such that $\E{\vec{\nu}_{*}(t)} = \vec{0}$ and $\E{\vec{\nu}_{*}(t)\vec{\nu}_{*}^\top(\tau)} = W_{*}\delta(t-\tau)$, where $W_{*}$ is the covariance matrix of the noise $\vec{\nu}_{*}$. 
Additionally, we assume that $\vec{\nu}_{*}$ for all  $*=R,\vec{s},\vec{\tau}, \vec{f}$, are mutually uncorrelated.
We impose that the aforementioned noises affect the system through the channels of the input $u = (R,\vec{s},\vec{\tau},\vec{f})$, respectively, such that for any $\hat{u} \triangleq (\hat{\xR}, \hat{\xs}, \hat{\vec{\tau}}, \hat{\vec{f}}) \in U$, we have
\begin{align*}
	u = \left(\hat{\xR}\Exp{\vec{\nu}_R}, \hat{\xs} + \vec{\nu}_{\vec{s}}, \hat{\vec{\tau}} + \vec{\nu}_{\vec{\tau}},  \hat{\vec{f}} + \vec{\nu}_{\vec{f}} \right).
\end{align*}
Note that the first element of the input tuple $u$ may represent the filtered measurement value $\hat{R}$ of the spacecraft's \textit{inertial star tracker instrument} perturbed by its associated uncertainty $\vec{\nu}_{R}$ and the second element of $u$ may represent the filtered measurement value $\hat{\xs}$ of its \textit{rate gyro instrument}, perturbed by the associated uncertainty $\vec{\nu}_{\xs}$. 
This inclusion allows us to fuse the known filtered measurements of a star tracker system and a rate gyro into the overall navigation solution.
By this means, we avoid having to represent the inertial kinematics and dynamics of the spacecraft in our state space as separate equations.
In fact, attitude control inputs, such as actuation torques, which are required to propagate these equations, are typically not known a priori.

The resulting stochastic equations of motion, the derivation of which is detailed in Appendix~\ref{sec:EoM_Stochastic}, are stated as
\begin{align}
	\diff \vec{\kappa} &=  f_{\xQ}(x,\hat{u},p) \diff t  + \begin{bmatrix}
		0_{3\times3}  & L_{\vec{s}} & 0_{3\times 3} & 0_{3\times3}
	\end{bmatrix}\diff \vec{\varepsilon}, \nonumber \\
	\diff \xw &= f_{\xw}(x,\hat{u},p)\diff t + \begin{bmatrix}
		0_{3\times3}  & 0_{3\times 3} & L_{\vec{\tau}} & 0_{3\times3}
	\end{bmatrix}\diff \vec{\varepsilon}, \nonumber \\
	\diff \xr &= f_{\xr}(x)\diff t, \nonumber \\
	\diff \xv &= f_{\xv}(x,\hat{u},p)\, \diff t \nonumber \\
	&+ \begin{bmatrix}Q\sk{\hat{\xR}^\top \vec{g}(\vec{d})}L_R & 0 & -\sk{\vec{c}}M^{-1} L_{\vec{\tau}}& \xQ L_{\vec{f}} \end{bmatrix}\, \diff \vec{\varepsilon},
	\label{eq:stoch_EoMs_concise}
\end{align}
where $\hat{u} = (\hat{\xR}, \hat{\xs}, \vec{0}, \hat{\vec{f}}) \in U \triangleq \mathrm{SO}(3) \times \mathbb{R}^{3}\times \mathbb{R}^{3}\times \mathbb{R}^{3}$, $x = (\xQ,\xr,\xw,\xv) \in X \triangleq \mathrm{SO}(3) \times \mathbb{R}^{3}\times \mathbb{R}^{3}\times \mathbb{R}^{3}$, and $p \in P$ is a collection of parameters. 
We can rewrite~(\ref{eq:stoch_EoMs_concise}) in succinct form as
\begin{align}
	\diff x = f(x,\hat{u},p)\diff t + L(x, \hat{u}, p)  \, \diff \vec{\varepsilon}.
	\label{eq:stoch_concise}
\end{align}

Observe that in equation~(\ref{eq:stoch_dw}) in 
Appendix~\ref{sec:EoM_Stochastic}, the nominal external torque $\hat{\vec{\tau}}$ does not appear since it has been set to zero.
Indeed, the very large principal moments of inertia of the small-body body and the relative weakness of persistent external torque applied on the small-body support the assumption that the change in angular velocity due to the term $\hat{\vec{\tau}}$ is negligible to that of the quantity $- \sk{\xw}  C A C^\top \xw$ over the duration of the spacecraft's orbit around the small-body.
Nevertheless, we maintain the noise term $\vec{\nu}_{\vec{\tau}}$ for capturing the imprecision of this approximation in the formulation of the factor, and in doing so, sustain smoothability of the small-body angular velocity state, as further discussed in Section~\ref{sec:smoothability}.
\begin{figure}[ht]
	\centering
	\def\Elevation{25} 
	\def\Azimuth{125} 
	\includegraphics[width=3.2in]{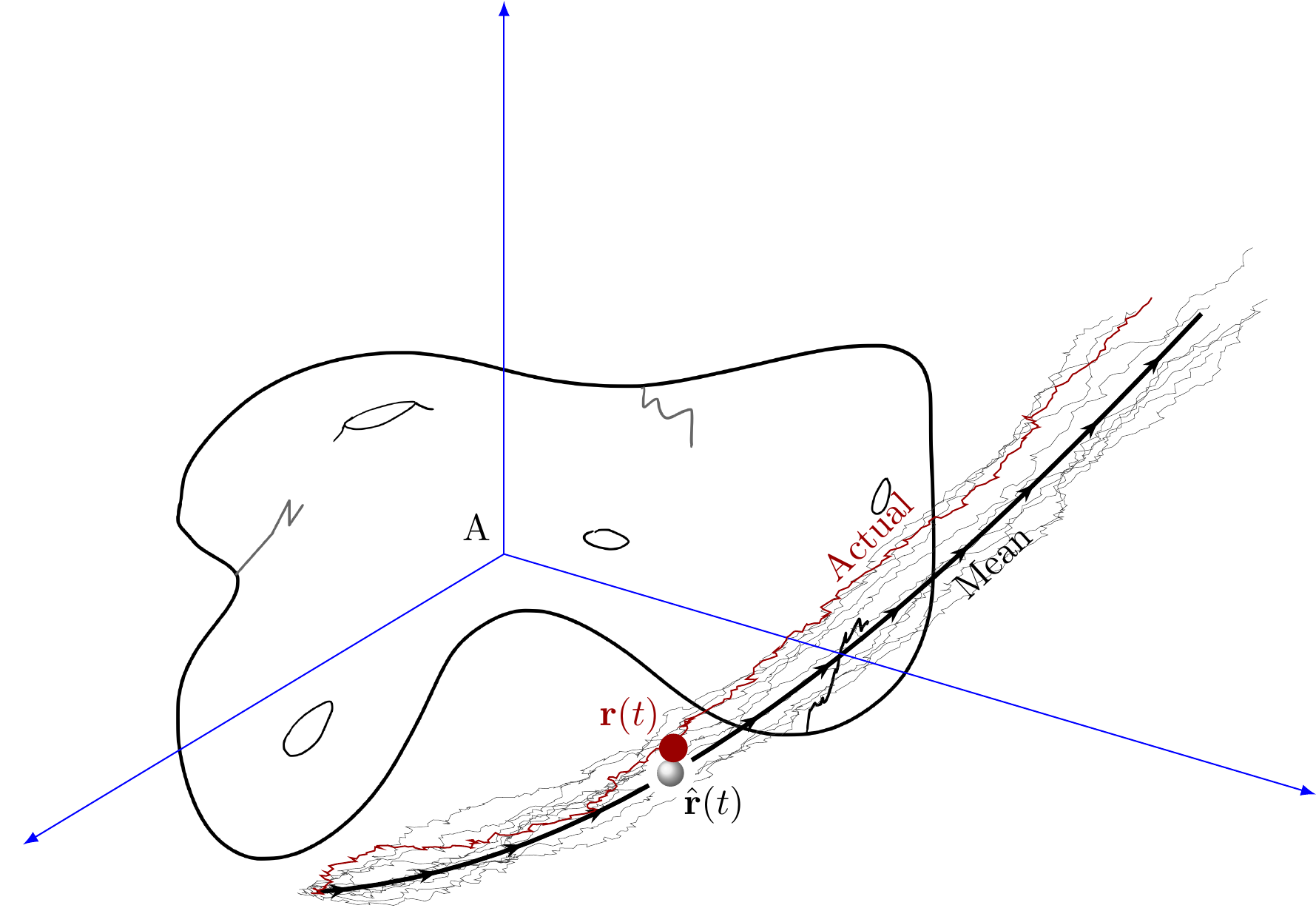}
	\caption{Illustration of stochastic system realizations and estimated solution.}
\end{figure}
Due to the non-linearity of the considered dynamical system in equation (\ref{eq:stoch_concise}), the distribution of the It\^{o} process $x(t) $ may require more than two moments to fully describe at any time. 
However, as a stepping stone to obtain a practical representation of the distribution $p(x(t)|x(t_0))$, we are interested in computing the first two moments $\hat{x}(t) = (\hat{\xQ}(t), \hat{\xw}(t),\hat{\xr}(t),\hat{\xv}) \triangleq \E{x(t)}$ and $\Sigma(t) \triangleq \E{\Delta_X(\hat{x}(t),x)\Delta_X^\top(\hat{x}(t),x)} $ at each time $t$.

Naturally, we want the relative dynamics factor to be proportional to the distribution $p(x_{k+1}|x_{k})$, as shown in the beginning of the Section~\ref{sec:reldyn_factor_formulation}.
Noting that the time step between $t_k$ and $t_{k+1}$ is not known a priori during the navigation segment, it is crucial to compute a process noise covariance $P_k$, which appropriately scales with the time step length, using a discretization scheme.
Such a treatment is similar to the time update step of an Extended Kalman Filter with first-order approximation assumptions and can be referenced in~\cite{sage1971estimation}.

First, for the continuous case, we define the \texttt{RelDyn} factor residual $\epsilon^{\mathrm{RelDyn}}$ such that
\begin{align*}
	\epsilon^{\mathrm{RelDyn}}(x(t) , x(t + \diff t)) \triangleq \diff x(t) - f(x(t), \hat{u}(t), p)\, \diff t.
\end{align*}
Then, the factor $\phi^{\mathrm{RelDyn}}$, as a function proportional to $\mathcal{N}\left(\vec{0}, \diff P(t)\right)$, may be simply defined as
\begin{align*}
	&\phi^{\mathrm{RelDyn}}(x(t) , x(t + \diff t)) \nonumber \\
	&\triangleq \exp\left( (\epsilon^{\mathrm{RelDyn}} (x(t) , x(t + \diff t)))^\top\diff P^{-1}(t) \epsilon^{\mathrm{RelDyn}}(*)\right).
\end{align*}

Let the partition $\left\lbrace t_k \right\rbrace_{k=0}^{N}$ of the time interval $\left[t_0, t_f\right]$ be such that $t_0 < t_1 < \ldots < t_k < \ldots < t_N = t_f$.  Given some $\hat{u}(t), \, t \in \left[t_0, t_f\right]$, we generate the sequence of predictions $\left\lbrace \hat{x}_k \right\rbrace_{k=0}^{N}$ such that 
\begin{align}    \label{eq:disc_f_of_x_sigma_u}
	\Delta_{X}(\hat{x}_{k+1}, \hat{x}_k) 
	= \int_{t_{k}}^{t_{k+1}} \hat{f}(\hat{x}(\tau),\hat{u}(\tau),p) \, \diff \tau, 
\end{align}
is a valid discretization of the first-order approximation $\dot{\hat{x}}(t) \approx \hat{f}(\hat{x}(t), \hat{u}(t),p)$ for all $k=1,\ldots,N-1$, and such that
\begin{align}      \label{eq:process_noise_discrete}
	P_k = \int_{t_k}^{t_{k+1}}  L(\hat{x}(\tau),\hat{u}(\tau),p)L^\top(\hat{x}(\tau),\hat{u}(\tau),p)\,  \diff t, 
\end{align}
characterizes the discrete process noise over the time-span $[t_k, t_{k+1}]$.
Then, using the definition in equation~(\ref{eq:def_Delta}) and the equation~(\ref{eq:disc_f_of_x_sigma_u}), we define the \texttt{RelDyn} residual $\epsilon_{k}^{\mathrm{RelDyn}}$ for the discrete case, as
\begin{align}
	&\epsilon_{k}^{\mathrm{RelDyn}}(x_{k}, x_{k+1}) \nonumber \\
	&\triangleq \begin{bmatrix}
		\log \left( \left(Q_{k}^\top Q_{k+1}\right)^\top \Omega_{k}^{\mathrm{prop}}\right)^\vee \\
		\vec{w}_{k+1} - \vec{w}_k - \int_{t_{k}}^{t_{k+1}} f_{\vec{w}}(x(\tau),\hat{u}(\tau),p) \diff \tau \\
		\vec{r}_{k+1} - \vec{r}_k - \int_{t_{k}}^{t_{k+1}} f_{\vec{r}}(x(\tau),p) \diff \tau \\
		\vec{v}_{k+1} - \vec{v}_k - \int_{t_{k}}^{t_{k+1}} f_{\vec{v}}(x(\tau),\hat{u}(\tau),p) \diff \tau 
	\end{bmatrix},
	\label{eq:deviation_reldyn_discrete}
\end{align}
where $ x(t_k) = \left(Q_k, \vec{w}_k, \vec{r}_k, \vec{v}_k\right)$ and 
where $\Omega_{k}^{\mathrm{prop}}$ is defined such that, for a partition $t_k=s_0<s_1<\ldots<s_n=t_{k+1}$,
\begin{align*}
	\Omega_{k}^{\mathrm{prop}} \triangleq \lim_{n\rightarrow\infty} \prod_{i=0}^{n-1}\exp \left(\int_{s_i}^{s_{i+1}} \sk{f_Q(x(\tau),\hat{u}(\tau),p)} \, \diff \tau\right), 
\end{align*}
and is dubbed the McKean-Gangolli injection method~\cite{chirikjian2011stochastic}, allowing for the stochastic process on the Lie group $\mathrm{SO}(3)$ to be written as a product integral. With appropriate correction factors, the limit can be truncated to numerically perform on-manifold integration with minimal error accumulation stemming from the approximation~\cite{andrle2013geometric}, as discussed in Section~\ref{sec:impl_RelDyn} below.

The factor $ \phi_{k}^{\mathrm{RelDyn}}$ for the discrete case may now be written as
\begin{align*}
	&\phi_{k}^{\mathrm{RelDyn}} (x_k,x_{k+1})  \nonumber \\
	&\triangleq \exp \left( (\epsilon_{k}^{\mathrm{RelDyn}}(x_{k}, x_{k+1}))^\top P_k^{-1} \epsilon_{k}^{\mathrm{RelDyn}}(x_{k}, x_{k+1})\right).
\end{align*}

\subsection{Smoothability of chosen state variables}
\label{sec:smoothability}

The MAP solution, which was discussed in Section~\ref{sec:Bayesian_description}, is shown to be equivalent to the solution obtained from optimal fixed-interval smoothing~\cite{sage1971estimation}.
Typically, the optimal smoothed state is defined as a linear combination of the state of the \textit{forward filter} and the state of the \textit{backward filter} at each time, with optimally chosen weights.
A state is said to be \textit{smoothable} if an optimal smoother provides a state estimate superior to that obtained when the final optimal filter estimate is extrapolated backwards in time~\cite{gelb1974applied}.
Furthermore, \cite{fraser1967new} has shown that, given linear forward and backward filters, only those states which are controllable by the noise driving the system state vector are smoothable.

In this context, we evaluate the relevance of our choice of state variables and the noise input and derived stochastic equations of motion in Appendix~\ref{sec:EoM_Stochastic} in terms of smoothability.
However, since the concept of smoothability pertains to linear smoothing, the direct application of the criterion to our problem may be misleading.
Instead, by demonstrating small-time local controllability through the noise by analyzing a local linearization of the equations~(\ref{eq:dQdt}) to~(\ref{eq:dvdt}) at some $(\hat{x}_k,\hat{u}_k$), we can assess the smoothability of our chosen state variables at that point.

Specifically, we can compute the local linearization matrices
\begin{align}
	F_k \triangleq \left. \frac{\partial f}{\partial x} \right\rvert_{\substack{x=\hat{x}_k\\u=\hat{u}_k}} = 
	\begin{bmatrix}
		\frac{\partial f_{\vec{Q}}}{\partial \vec{\kappa}} & \frac{\partial f_{Q}}{\partial \vec{w}} & 0 & 0 \\
		0 & \frac{\partial f_{\vec{w}}}{\partial \vec{w}} & 0 & 0 \\
		0 & \frac{\partial f_{\vec{r}}}{\partial \vec{w}} & \frac{\partial f_{\vec{r}}}{\partial \vec{r}} & \frac{\partial f_{\vec{r}}}{\partial \vec{v}} \\
		\frac{\partial f_{\vec{v}}}{\partial \vec{\kappa}} & \frac{\partial f_{\vec{v}}}{\partial \vec{w}} & \frac{\partial f_{\vec{v}}}{\partial \vec{r}} & \frac{\partial f_{\vec{v}}}{\partial \vec{v}}
	\end{bmatrix}_{\substack{x=\hat{x}_k\\u=\hat{u}_k}},
\end{align}
where
{\small 
	\begin{align*}
		\frac{\partial f_{Q}}{\partial \vec{\kappa}} &= \sk{\xQ^\top \xw}, \, \sk{\vec{\kappa}} \in T_{Q}\mathrm{SO}(3),\, \frac{\partial f_{Q}}{\partial \vec{w}} = -\xQ^\top, \nonumber \\
		\frac{\partial f_{\vec{w}}}{\partial \vec{w}} &= M^{-1}\left(\sk{M\xw} - \sk{\xw}M\right), \nonumber \\
		\frac{\partial f_{\vec{r}}}{\partial \vec{w}} &= \sk{\xr}, \, \frac{\partial f_{\vec{r}}}{\partial \vec{r}} = -\sk{\xw}, \, \frac{\partial f_{\vec{r}}}{\partial \vec{v}} = \mathrm{I}_3, \nonumber \\
		\frac{\partial f_{\vec{v}}}{\partial \vec{\kappa}} &= -\xQ\sk{\vec{f}} - \xQ\sk{R^\top\vec{g}(\vec{d})}, \nonumber \\
		\frac{\partial f_{\vec{v}}}{\partial \vec{w}} &= -\sk{\vec{c}}M^{-1}\left(\sk{M\xw} - \sk{\xw}M\right) \nonumber \\
		& - \sk{\xw}\sk{\vec{c}} - \sk{\sk{\vec{w}}\vec{c} } + \sk{\xv}, \nonumber \\
		\frac{\partial f_{\vec{v}}}{\partial \vec{r}} &= \mu_{a}/ \|\vec{z}\|^5 \left(3\vec{z}\vec{z}^\top - \|\vec{z}\|^2 \mathrm{I}_3\right) \nonumber \\
		&- \mu_{\odot}/\|\vec{d}\|^3\mathrm{I}_3, \, \vec{z} \triangleq \xr - \vec{c}, \nonumber \\
		\frac{\partial f_{\vec{v}}}{\partial \vec{v}} &= -\sk{\xw},
	\end{align*}
}%
and $L_k = L(\hat{x}_k, \hat{u}_k,p)$, where $L$ is as defined in Appendix~\ref{sec:EoM_Stochastic}.

We now apply the PBH controllability test~\cite{williams2007linear, fraser1967new} to show that omitting a noise term in equation~(\ref{eq:stoch_dw}) will lead to situations where the pair $(F_k, L_k)$ is uncontrollable. 
For example, take the specific case where $M = \mathrm{diag}(m_{1}, m_{2}, m_{3})$ and where $\hat{\xw}_k = [0, \, 0,\, \omega]^\top$ has only one non-zero component. 
This scenario is similar to the principal axis-rotator small-body case with frames $\fG$ and $\fA$ aligned.
Then,
\begin{align*}
	\left.\frac{\partial f_{\vec{w}}}{\partial \vec{w}} \right\rvert_{\substack{x=\hat{x}_k\\u=\hat{u}_k}} &= M^{-1}\left(\sk{M\hat{\xw}_k} - \sk{\hat{\xw}_k}M\right) \nonumber \\
	&= {\small \begin{bmatrix}
		0 & (m_{2}-m_{3})/m_{1}\omega & 0 \\
		(m_{3}-m_{1})/m_{2}\omega & 0 & 0 \\
		0 & 0 & 0
	\end{bmatrix}}.
    \renewcommand{\arraystretch}{1.0}
\end{align*}
By inspection, such a choice leads to a row of zeros in $F_k$, and hence $\lambda=0$ will be an eigenvalue of $F_k$.
Subsequently, we write out the PBH rank test matrix for $\lambda=0$. 
We note that if $L_{\tau}=0$, then $\mathrm{rank}\left[F_k \quad L_k\right]<12$, and therefore the smoothability condition of the state $\vec{w}$ is not satisfied.
Given the concatenation of several full rank matrices in the other rows of the PBH test matrix, it is unlikely that the rank is less than 12 for any other eigenvalue of $F_k$, so long as $L_{\tau}\neq 0$.
Therefore, it is crucial to include appropriate and physically justified noise terms in the considered dynamics to support smoothability.
Furthermore, the inclusion of such noise avoids the degeneracy of the discrete process noise Gaussian distribution, with covariance $P_k$, as computed in Section~\ref{sec:reldyn_factor_formulation}.
This argument supports the inclusion of a noise term through the $\vec{\tau}$ channel in the derivation of the stochastic equations of motion in Appendix~\ref{sec:EoM_Stochastic}.

\section{Algorithm and implementation details}
\label{sec:methods}

We are ready now to finally state the full solution to the problem and provide the key implementation details.
Equations (\ref{eq:disc_f_of_x_sigma_u}) and (\ref{eq:process_noise_discrete}) constitute constraints on the evolution of the mean system state and its associated system noise covariance between time instances $t_k$ and $t_{k+1}$.
We wish to enforce these constraints.
Practically, at each time instance $t_k,\, k=1\ldots N$, 
we let $\tilde{x}_k$ be the current best guess of $\hat{x}_k$ and $\tilde{P}_k$ be the current best guess of $P_k$.
Given $\tilde{x}_{k}$, $\tilde{x}_{k+1}$ and $\hat{u}(t), \, t \in [t_k, t_{k+1}]$, we compute the residual due to equation (\ref{eq:disc_f_of_x_sigma_u}), then the guess matrix $\tilde{P}_k$, and finally the value of the factor $\phi_{k}^{\mathrm{RelDyn}} (\tilde{x}_k,\tilde{x}_{k+1})$.
Note that the integrals in equations~(\ref{eq:deviation_reldyn_discrete}) and (\ref{eq:process_noise_discrete}) are computed numerically using an on-manifold integration scheme, detailed in Section~\ref{sec:methods}.

\begin{figure}[h]
	\centering
	\begin{tikzpicture}[xscale=0.7, yscale=0.6]
		\node[shape=circle, draw=black, minimum size = 0.3in] (T0) at (-1.8 in,0 in) {$T_0$};
		
		\node[shape=circle, draw=black, minimum size = 0.3in] (T1) at (-0.5 in,0 in) {$T_1$};
		\node[shape=circle, draw=black, minimum size = 0.3in, text=black] (T2) at (0.8 in,0 in) {$T_2$};

		\node[shape=circle, draw=black, minimum size = 0.3in, text=blue, below = 0.1in of T0] (w0) {$\bm{\mathrm{w}}_0$};
		\node[shape=circle, draw=black, minimum size = 0.3in, text=forestgreen, below = 0.1in of w0] (v0)  {$\bm{\mathrm{v}}_0$};
		
		\node[shape=circle, draw=black, minimum size = 0.3in, text=blue, below = 0.1in of T1] (w1) {$\bm{\mathrm{w}}_{1}$};
		\node[shape=circle, draw=black, minimum size = 0.3in, text= forestgreen, below = 0.1in of w1] (v1) {$\bm{\mathrm{v}}_{1}$};

		\node[shape=circle, draw=black, minimum size = 0.3in, text=blue, below = 0.1in of T2] (w2) {$\bm{\mathrm{w}}_{2}$};
		\node[shape=circle, draw=black, minimum size = 0.3in, text= forestgreen, below = 0.1in of w2] (v2) {$\bm{\mathrm{v}}_{2}$};
		
		\node (ell2) at ($(w2)+(0.8in,0)$) {$\cdots$};

		\node[shape=circle, draw=black, minimum size = 0.3in] (x2) at (-1.6in , 1.5 in) {$\ell_1$};
		\node[shape=circle, draw=black, minimum size = 0.3in] (x3) at (-0.9 in, 1.5 in) {$\ell_3$};
		\node[shape=circle, draw=black, minimum size = 0.3in, text=black] (x4) at (0.0 in, 1.5 in) {$\ell_4$};
		\node (ell1) at (0.8in , 1.5 in) {$\cdots$};

		\node[shape=circle, fill=black, label=left:$\phi_1^{\text{proj}}$] (m1) at ($ 0.5*(T0) + 0.5*(x2) $ ) {};
		\node[shape=circle, fill=black, label=above:$\phi_2^{\text{proj}}$] (m2) at ($ 0.5*(x2) + 0.5*(T1) $ ) {};
		\node[shape=circle, fill=black, label=below :$\phi_3^{\text{proj}}$] (m3) at ($ 0.5*(T0) + 0.5*(x3) $ ) {};
		\node[shape=circle, fill=black,, label=below:$\phi_5^{\text{proj}}$] (m5) at ($ 0.75*(x4) + 0.25*(T0) $ ) {};
		\node[shape=circle, fill=black, label=below:$\phi_4^{\text{proj}}$] (m4) at ($ 0.5*(x3) + 0.5*(T1) $ ) {};
		\node[shape=circle, fill=black, label=below:$\phi_6^{\text{proj}}$] (m6) at ($ 0.5*(x4) + 0.5*(T1) $ ) {};
		\node[shape=circle, fill=black, label=above left:$\phi_7^{\text{proj}}$] (m7) at ($ 0.5*(x4) + 0.5*(T2) $ ) {};
		
		\node[shape=circle, fill=black, label=above:$\phi^{\mathrm{prior}}(T_0)$, left=0.15in of T0] (pT0) {};
		\node[shape=circle, fill=black, label=above:$\phi^{\mathrm{prior}}(\vec{w}_0)$, left=0.15in of w0] (pw0) {};
		\node[shape=circle, fill=black, label=above:$\phi^{\mathrm{prior}}(\vec{v}_0)$, left=0.15in of v0] (pv0) {};

		\path[-] (pw0) edge (w0);
		\path[-] (pv0) edge (v0);

		\node[shape=circle, fill=black, label=above:$\phi^{\text{RelDyn}}_1$] (f1) at ($ 0.5*(T0) + 0.5*(v1) + (0in, 0.0in)$ ) {};
		
		\path[-] (f1) edge (T0);
		
		\path[-] (f1) edge (w0);
		\path[-] (f1) edge (v0);

		\path[-] (f1) edge (T1);
		
		\path[-] (f1) edge (w1);
		\path[-] (f1) edge (v1);
		
		\node[shape=circle, fill=black, label=above:$\phi^{\text{RelDyn}}_2$] (f2) at ($ 0.5*(T1) + 0.5*(v2) + (0in, 0.0in)$ ) {};
		
		\path[-] (f2) edge (T1);
		
		\path[-] (f2) edge (w1);
		\path[-] (f2) edge (v1);

		\path[-] (f2) edge (T2);
		
		\path[-] (f2) edge (w2);
		\path[-] (f2) edge (v2);

		\path[-] (pT0) edge (T0);
		\path[-] (x2) edge (m1);
		\path[-] (x3) edge (m3);
		\path[-] (x2) edge (m2);
		
		\path[-] (m1) edge (T0);
		\path[-] (m3) edge (T0);
		\path[draw=black,-] (m5) edge (T0);
		\path[draw=black,-] (m5) edge (x4);
		\path[-] (m2) edge (T1);
		\path[-] (m4) edge (T1);
		\path[draw=black,-] (m6) edge (T1);
		\path[draw=black,-] (x4) edge (m7);
		\path[draw=black,-] (m7) edge (T2);
		\path[-] (x3) edge (m4);
		\path[draw=black,-] (x4) edge (m6);
		
	\end{tikzpicture}
	\caption{Full \texttt{AstroSLAM} factor graph.}
	\label{fig:factor_graph_RelDyn_v2}
\end{figure}
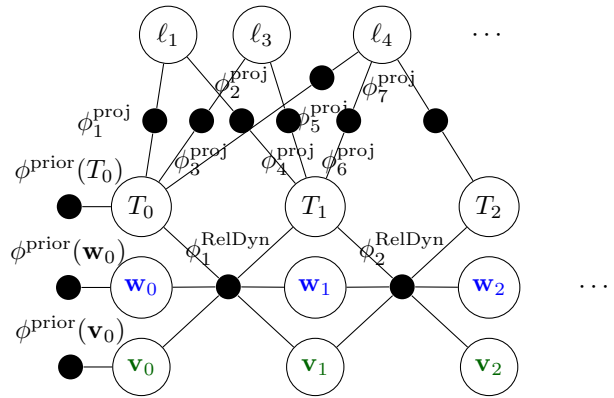

Recalling the development in Section~\ref{sec:reldyn_factor_formulation}, and the definitions in Section~\ref{sec:Bayesian_description}, we can now write the maximizing posterior as 
\begin{align}
	\max_{\substack{\tilde{x}_n \in X,\, n=1,\ldots ,N,\\ \tilde{\ell}_m \in \mathbb{R}^3, m=1,\ldots, M}} &\phi^{\mathrm{prior}}(\tilde{x}_0)\left(\prod_{k=0}^{N-1} \phi_{k}^{\mathrm{RelDyn}}(\tilde{x}_{k}, \tilde{x}_{k+1})\right) \nonumber \\
	&\times \left(\prod_{{(k,i)} \in \mathcal{J}_n} \phi_{*}^{\mathrm{proj}}(\tilde{x}_k, \tilde{\ell}_i) \right),
	\label{eq:posterior_developped}
\end{align}
where $\phi_{*}^{\mathrm{proj}}$ is the projection factor, $\mathcal{J}_n$ is the set of landmark-frame data-association index pairs, and $\phi^{\mathrm{prior}}$ is the prior distribution factor. 
As both the prior and projection factors are developed in detail in other works (see~\cite{dellaert2017factor}, for example), we simply restate them here using our notation, such that
\begin{align*}
	\phi^{\mathrm{prior}}(x_0) &\triangleq \exp \left( \Delta_X(\hat{x}_0, x_0)^\top \Sigma_0^{-1} \Delta_X(\hat{x}_0, x_0)\right), \\
	\phi_{*}^{\mathrm{proj}}(x_k,\ell_i) &\triangleq \exp \left( (\vec{y}_{ik}^{m}-\vec{y}_{ik})^\top (\Sigma_{\vec{y}}^{m})^{-1} (\vec{y}_{ik}^{m}-\vec{y}_{ik}) \right),
\end{align*}
where each $\vec{y}_{ik}^{\mathrm{m}} \in \mathcal{Y}_k$ is obtained as per the relationship in equation~(\ref{eq:landmark_projection}).
Note that, as a consequence of independence, $\phi^{\mathrm{prior}}(x_0) = \phi^{\mathrm{prior}}(T_0)\phi^{\mathrm{prior}}(\vec{w}_0)\phi^{\mathrm{prior}}(\vec{v}_0)$.
The structure of the cost in~(\ref{eq:posterior_developped}) may now be fully represented by a factor graph, an example of which is presented in Figure~\ref{fig:factor_graph_RelDyn_v2},
where $T_k$ is constructed using $Q_k$ and $\vec{r}_k$, as per relationship~\eqref{eq:SE3_def}.

Since each of the considered factors is proportional to a Gaussian distribution, by taking the negative logarithm of~(\ref{eq:posterior_developped}), we can write an equivalent minimization problem, constituting a sparse non-linear least squares problem~\cite{dellaert2017factor}, stated as
\begin{align}
	&\min_{\substack{\tilde{x}_n \in X,\, n=1,\ldots ,N,\\ \tilde{\ell}_m \in \mathbb{R}^3, m=1,\ldots, M}} 	(\epsilon^{\mathrm{prior}}(\tilde{x}_0))^\top\Sigma_{0}^{-1}\epsilon^{\mathrm{prior}}(\tilde{x}_0)) \nonumber \\ &+\sum_{k=0}^{N-1} (\epsilon_{k}^{\mathrm{RelDyn}}(\tilde{x}_{k}, \tilde{x}_{k+1}))^{\top} \tilde{P}_k^{-1} \epsilon_{k}^{\mathrm{RelDyn}}(\tilde{x}_{k}, \tilde{x}_{k+1}) \nonumber \\
	&+ \sum_{ (k,i) \in \mathcal{J}} (\epsilon_{*}^{\text{proj}}(\tilde{x}_k, \tilde{\ell}_i))^{\top} (\Sigma_{\vec{y}}^{\mathrm{m}})^{-1} \epsilon_{*}^{\text{proj}}(\tilde{x}_n, \tilde{\ell}_i).
	\label{eq:full_min_least_squares}
\end{align}


Next, we provide the implementation details of the algorithm, 
while underlining the specificities of the asteroid navigation problem.
We first discuss the initialization steps of the \texttt{AstroSLAM} algorithm, in which we insert prior factors for initial poses and kinematic variables of both the spacecraft and small-body, as well as generate an initial estimate for the map $\Psi_1$. 

Our processing pipeline consists of a front-end system and a back-end system predicated on the iSAM2 engine and GTSAM library, along with an initialization step and a loop closure detection step.

\subsection{State initialization}

We leverage estimates from pre-encounter Earth-based measurements and approach phase sensor measurements to perform the initialization of both the spacecraft and small-body states.
Inertial position  measurements $\vec{r}_{\oS_k\oI}^{\fI,\mathrm{m}}$ of the spacecraft, modelled as $\vec{r}_{\oS_k\oI}^{\fI,\mathrm{m}} \triangleq \vec{r}_{\oS_k\oI}^\fI + \vec{\nu}_{\vec{r}},$ $\vec{\nu}_{\vec{r}} \sim \mathcal{N} \left(\vec{0}_{3\times 1}, \Sigma_{\vec{r}}^{\mathrm{m}}\right)$, are based on Earth-relative radiometric ranging and bearing measurements, a method of localization widely practiced in deep space mission spacecraft tracking using communication station networks, such as the DSN.
Acurate ground-based navigation estimates using Earth-relative range and bearing measurements, such as uplink-downlink pulse ranging and Delta-Differenced One-Way Ranging (DDOR), a type of Very Long Baseline Interferometry (VLBI)~\cite{Miller2019, miller1977application}, are available to be used for initialization.
It is important to note that our algorithm requires DSN-type measurements only during the initialization phase, to anchor the initial pose as described below.

Inertial position estimates of the small-body, denoted $\hat{\vec{r}}_{\oG_k\oI}^{\fI}$, are predicated on orbit determination (OD) performed using Earth-based telescopic measurements.
An initial relative position vector $\hat{\vec{r}}_{\oS_0\oG}$ estimate may be obtained by differencing spacecraft and small-body inertial position estimates.
Alternatively, relative optical navigation (OpNAV) performed using pre-encounter sensor measurements, accurate in the order of several hundreds of meters for large small-bodies and tens of meters for small small-bodies~\cite{bhaskaran2011small}, may be used to provide such an initial prior on the relative position.

To further simplify the initialization, we assume that the small body is a stable single axis rotator.
Thus, we align the arbitrary frame $\fG$ such that $\vec{\omega}_{\fG_0\fI}  = w_0 \uvec{g}_3$, where $w_0$ is the magnitude of the angular velocity vector of the small-body.
The unit vector $\uvec{g}_{3}^{\fI}$ is classically parameterized by the spin pole tilt angles relative to the J2000 ecliptic plane.
If sufficient pre-encounter observations are available, a prior for these angles can be estimated and their associated uncertainty computed~\cite{thomas1997vesta}.

Typically, a prominent and salient feature on the surface is hand-picked as the prime meridian direction, thus fixing the $\uvec{g}_1$ axis.
In our case, given the initial prior $\vec{r}_{\oS_0\oG}^{\fI}$ discussed earlier, the prime meridian $\uvec{g}_1$ may be initialized by first computing $\uvec{g}_2$ following $\uvec{g}_{2}^{\fI} = \sk{\uvec{g}_{3}^{\fI}} {\vec{r}_{\oS_0\oG}^{\fI}}/{\|\vec{r}_{\oS_0\oG}^{\fI}\|}$,
and then computing $\uvec{g}_{1}^{\fI} = \sk{\uvec{g}_{2}^{\fI}}\uvec{g}_{3}^{\fI}$.
Consequently, $R_{\fI\fG_0} = \begin{bmatrix}
	\uvec{g}_{1}^{\fI} & \uvec{g}_{2}^{\fI} & \uvec{g}_{3}^{\fI}
\end{bmatrix}$ is determined.
Note that, for this method to work, the spacecraft relative position at initialization time cannot be coincident with the small-body spin axis.
In practice, an on-board star tracker system is leveraged to obtain an orientation measurement $R_{\fI\fS_k}^{\mathrm{m}}$, modelled as $R_{\fI\fS_k}^{\mathrm{m}} \triangleq R_{\fI\fS_k}\Exp{\vec{\nu}_R}, \, \vec{\nu}_R\sim \mathcal{N} \left(\vec{0}_{3\times 1}, \Sigma_{\mathrm{R}}^{\mathrm{m}}\right)$. 
Orientation measurements are usually known with very good accuracy and little uncertainty.

We may now establish the prior factor for relative pose $T_{\fG_0\fS_0}$ by combining the information we have available at initialization.
We first define the relative measurement at time $k=0$, given by
\begin{align*}
	T_{\fG_0\fS_0}^{\mathrm{m}} &= \begin{bmatrix}
		\hat{R}_{\fI\fG_0}^{\top}R_{\fI\fS_0}^{\mathrm{m}} & \hat{R}_{\fI\fG_0}^{\top}\vec{r}_{\oS_0\oG}^{\fI, \mathrm{m}} \\
		0_{1\times 3} & 1
	\end{bmatrix} \\
	&= \begin{bmatrix}
		\hat{R}_{\fI\fG_0}^{\top}R_{\fI\fS_0}\Exp{\vec{\nu}_{R}} & \vec{r}_{\oS\oG}^{\fG_0} +  \hat{R}_{\fI\fG_0}^{\top}\vec{\nu}_{\vec{r}}\\
		0_{1\times 3} & 1
	\end{bmatrix} \nonumber \\
	&= \begin{bmatrix}
		\hat{R}_{\fI\fG_0}^{\top}R_{\fI\fS_0}& \vec{r}_{\oS\oG}^{\fG_0} \\
		0_{1\times 3} & 1
	\end{bmatrix}\begin{bmatrix}
		\Exp{\vec{\nu}_{R}} & \hat{R}_{\fI\fS_0}^{\top}\vec{\nu}_{\vec{r}}\\
		0_{1\times 3} & 1
	\end{bmatrix},
\end{align*}
where $\vec{r}_{\oS_0\oG}^{\fI,\mathrm{m}}$ is the relative position vector at time $k=0$ obtained by differencing inertial position estimates of the spacecraft and the small-body, or by means of an OpNav 
solution.
It then follows that 
\begin{align}
	T_{\fG_0\fS_0}^{-1} T_{\fG_0\fS_{0}}^\mathrm{m} &= \begin{bmatrix}
		\Exp{\vec{\varepsilon}_1} & \vec{\varepsilon}_2\\ 0_{1\times 3} & 1
	\end{bmatrix},
	\label{eq:tgitis}
\end{align}
where $\vec{\varepsilon}_1 = \vec{\nu}_R$ and $\vec{\varepsilon}_2 = R_{\fI\fS_k}^\top\vec{\nu}_{\vec{r}}$. 
Using the logarithm map of the SE(3) group of homogeneous transformations at the identity and the \textit{vee} operator, as detailed in Section~\ref{sec:notation}, we can write
\begin{align}
	\Log{ T_{\fG_0\fS_0}^{-1}T_{\fG_0\fS_{0}}^\mathrm{m}} = \begin{bmatrix}
		\vec{\varepsilon}_1 \\ \vec{\varepsilon}_2
	\end{bmatrix}  \sim \left(\vec{0}_{6\times 1}, \Sigma_{\mathrm{T},0}^\mathrm{m}\right), 
	\label{eq:prior_T0}
\end{align}
where, by first-order linear approximation, we have $
\Sigma_{\mathrm{T},0}^\mathrm{m} = J_\mathrm{R} \Sigma_{\mathrm{R}}^{\mathrm{m}} J_\mathrm{R}^\top + J_{\vec{r}}(\hat{R}_{\fI\fG_0}^{\top}R_{\fI\fS_0}^{\mathrm{m}}) \Sigma_{\vec{r} }^{\mathrm{m}} J_{\vec{r}}^\top(\hat{R}_{\fI\fG_0}^{\top}R_{\fI\fS_0}^{\mathrm{m}})
$, with Jacobians
\begin{equation*}
	J_{\mathrm{R}} = \begin{bmatrix}\mathrm{I}_3 \\ 0_{3\times 3}\end{bmatrix}, 
	\qquad
	J_{\vec{r}}(R) = \begin{bmatrix}0_{3\times 3} \\ R^\top\end{bmatrix},
\end{equation*}
where $R \in \mathrm{SO}(3)$.
A prior factor $\phi_0^{\text{prior}}(T_{\fG_0\fS_0})$ is emplaced in the graph. This factor encodes the residual between the pose $T_{\fG_0\fS_0} \in \mathrm{SE}(3)$ and the measurement $T_{\fI\fS_0}^\mathrm{m}$, with covariance $\Sigma_{\mathrm{T},0}^\mathrm{m}$,
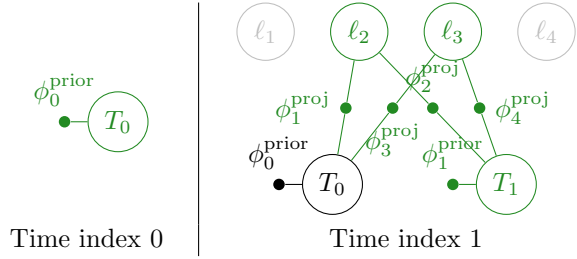
\begin{figure}[htpb]
	\begin{center}
		\begin{tabularx}{\linewidth}{>{\centering\arraybackslash}m{1.0in}|>{\centering\arraybackslash}m{2.0in}}

			\begin{tikzpicture}[xscale=0.7,yscale=0.8]
				\node[shape=circle, fill=dark-green, draw=none,minimum size = 0.06in, inner sep = 0.01in, label=above:\color{dark-green}$\phi_0^{\text{prior}}$] (p0) at (-2.2in,0in) {};
				\node[shape=circle, draw=dark-green, minimum size = 0.3in, text=dark-green] (T0) at (-1.8 in,0 in) {$T_0$};
				
				\path[dark-green,-] (p0) edge (T0);
				
			\end{tikzpicture}
			& 
			
			\begin{tikzpicture}[xscale=0.7, yscale=0.8]
				\node[shape=circle, fill=black, draw=none,minimum size = 0.06in, inner sep = 0.01in, label=above:$\phi_0^{\text{prior}}$] (p0) at (-2.2in,0in) {};
				\node[shape=circle, draw=black, minimum size = 0.3in] (T0) at (-1.8 in,0 in) {$T_0$};
				\node[shape=circle, fill=dark-green, draw=none,minimum size = 0.06in, inner sep = 0.01in, label=above:\color{dark-green}$\phi_1^{\text{prior}}$] (p1) at (-0.9in,0in) {};
				\node[shape=circle, draw=dark-green, minimum size = 0.3in, text=dark-green] (T1) at (-0.5 in,0 in) {$T_1$};

				\node[shape=circle, draw=lightgray, minimum size = 0.3in, text=lightgray] (x1) at (-2.3 in, 1.0 in) {$\ell_1$};
				\node[shape=circle, draw=dark-green, minimum size = 0.3in, text=dark-green] (x2) at (-1.6in , 1.0 in) {$\ell_2$};
				\node[shape=circle, draw=dark-green, minimum size = 0.3in, text=dark-green] (x3) at (-0.9 in, 1.0 in) {$\ell_3$};
				\node[shape=circle, draw=lightgray, minimum size = 0.3in, text=lightgray] (x4) at (-0.2 in, 1.0 in) {$\ell_4$};

				\node[shape=circle, fill=dark-green, draw=none,minimum size = 0.06in, inner sep = 0.01in, label=left:\color{dark-green}$\phi_1^{\text{proj}}$] (m1) at ($ 0.5*(T0) + 0.5*(x2) $ ) {};
				\node[shape=circle, fill=dark-green, draw=none,minimum size = 0.06in, inner sep = 0.01in, label=above:\color{dark-green}$\phi_2^{\text{proj}}$] (m2) at ($ 0.5*(x2) + 0.5*(T1) $ ) {};
				\node[shape=circle, fill=dark-green, draw=none,minimum size = 0.06in, inner sep = 0.01in, label=below :\color{dark-green}$\phi_3^{\text{proj}}$] (m3) at ($ 0.5*(T0) + 0.5*(x3) $ ) {};
				
				\node[shape=circle, fill=dark-green, draw=none,minimum size = 0.06in, inner sep = 0.01in, label=right:\color{dark-green}$\phi_4^{\text{proj}}$] (m4) at ($ 0.5*(x3) + 0.5*(T1) $ ) {};

				\path[draw = black,-] (p0) edge (T0);
				\path[draw = dark-green,-] (p1) edge (T1);
				\path[draw = dark-green,-] (x2) edge (m1);
				\path[draw = dark-green,-] (x3) edge (m3);
				\path[draw = dark-green,-] (x2) edge (m2);
				
				\path[draw = dark-green,-] (m1) edge (T0);
				\path[draw = dark-green,-] (m3) edge (T0);
				
				\path[draw = dark-green,-] (m2) edge (T1);
				\path[draw = dark-green,-] (m4) edge (T1);
				
				\path[draw = dark-green,-] (x3) edge (m4);

				
			\end{tikzpicture}
			\\ 
			Time index 0 &  Time index 1
		\end{tabularx}
	\end{center}
	\caption{Initialization steps of the SLAM problem.}
	\label{fig:init_steps_SLAM}
	
\end{figure}

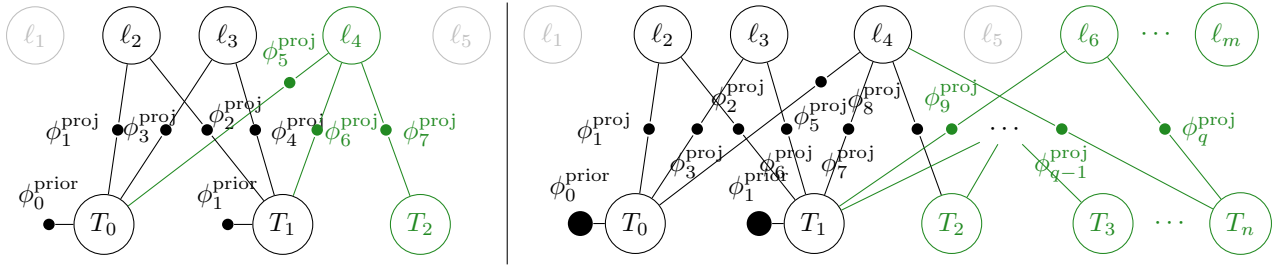
\begin{figure*}[t]
	\centering
	\begin{tabular}{c|c}
		
		\begin{tikzpicture}[xscale=0.72, yscale=0.66]
			\node[shape=circle, fill=black,draw=none,minimum size = 0.06in, inner sep = 0.01in, label=above:$\phi_0^{\text{prior}}$] (p0) at (-2.2in,0in) {};
			\node[shape=circle, draw=black, minimum size = 0.3in] (T0) at (-1.8 in,0 in) {$T_0$};
			\node[shape=circle, fill=black,draw=none,minimum size = 0.06in, inner sep = 0.01in, label=above:$\phi_1^{\text{prior}}$] (p1) at (-0.9in,0in) {};
			\node[shape=circle, draw=black, minimum size = 0.3in] (T1) at (-0.5 in,0 in) {$T_1$};
			\node[shape=circle, draw=dark-green, minimum size = 0.3in, text=dark-green] (T2) at (0.5 in,0 in) {$T_2$};

			\node[shape=circle, draw=lightgray, minimum size = 0.3in, text=lightgray] (x1) at (-2.3 in, 1.5 in) {$\ell_1$};
			\node[shape=circle, draw=black, minimum size = 0.3in] (x2) at (-1.6in , 1.5 in) {$\ell_2$};
			\node[shape=circle, draw=black, minimum size = 0.3in] (x3) at (-0.9 in, 1.5 in) {$\ell_3$};
			\node[shape=circle, draw=dark-green, minimum size = 0.3in, text=dark-green] (x4) at (0.0 in, 1.5 in) {$\ell_4$};
			\node[shape=circle, draw=lightgray, minimum size = 0.3in, text=lightgray] (x5) at (0.8in , 1.5 in) {$\ell_5$};

			\node[shape=circle, fill=black,draw=none,minimum size = 0.06in, inner sep = 0.01in, outer sep = 0.01in, label=left:$\phi_1^{\text{proj}}$] (m1) at ($ 0.5*(T0) + 0.5*(x2) $ ) {};
			\node[shape=circle, fill=black,draw=none,minimum size = 0.06in, inner sep = 0.01in, outer sep = 0.01in, label={[label distance=-0.1in]above right:$\phi_2^{\text{proj}}$}] (m2) at ($ 0.5*(x2) + 0.5*(T1) $ ) {};
			\node[shape=circle, fill=black,draw=none,minimum size = 0.06in, inner sep = 0.01in, outer sep = 0.01in, 
			label={[label distance=-0.2in]above left:$\phi_3^{\text{proj}}$}]  (m3) at ($ 0.5*(T0) + 0.5*(x3) $ ) {};
			\node[shape=circle, fill=dark-green,draw=none,minimum size = 0.06in, inner sep = 0.01in, outer sep = 0.01in, label=above:\color{dark-green}$\phi_5^{\text{proj}}$] (m5) at ($ 0.75*(x4) + 0.25*(T0) $ ) {};
			\node[shape=circle, fill=black,draw=none,minimum size = 0.06in, inner sep = 0.01in, label=right:$\phi_4^{\text{proj}}$] (m4) at ($ 0.5*(x3) + 0.5*(T1) $ ) {};
			\node[shape=circle, fill=dark-green,draw=none,minimum size = 0.06in, inner sep = 0.01in,,outer sep = 0.01in, label={ [label distance=-0.05in]right:\color{dark-green}$\phi_6^{\text{proj}}$}] (m6) at ($ 0.5*(x4) + 0.5*(T1) $ ) {};
			\node[shape=circle, fill=dark-green,draw=none,minimum size = 0.06in, inner sep = 0.01in, outer sep = 0.01in, label={right:\color{dark-green}$\phi_7^{\text{proj}}$}] (m7) at ($ 0.5*(x4) + 0.5*(T2) $ ) {};

			\path[-] (p0) edge (T0);
			\path[-] (p1) edge (T1);
			\path[-] (x2) edge (m1);
			\path[-] (x3) edge (m3);
			\path[-] (x2) edge (m2);
			
			\path[-] (m1) edge (T0);
			\path[-] (m3) edge (T0);
			\path[draw=dark-green,-] (m5) edge (T0);
			\path[draw=dark-green,-] (m5) edge (x4);
			\path[-] (m2) edge (T1);
			\path[-] (m4) edge (T1);
			\path[draw=dark-green,-] (m6) edge (T1);
			\path[draw=dark-green,-] (x4) edge (m7);
			\path[draw=dark-green,-] (m7) edge (T2);
			\path[-] (x3) edge (m4);
			\path[draw=dark-green,-] (x4) edge (m6);

			
		\end{tikzpicture}
		&
		\begin{tikzpicture}[xscale=0.72, yscale=0.66]
			\node[shape=circle, fill=black, label=above:$\phi_0^{\text{prior}}$] (p0) at (-2.2in,0in) {};
			\node[shape=circle, draw=black, minimum size = 0.3in] (T0) at (-1.8 in,0 in) {$T_0$};
			\node[shape=circle, fill=black, label=above:$\phi_1^{\text{prior}}$] (p1) at (-0.9in,0in) {};
			\node[shape=circle, draw=black, minimum size = 0.3in] (T1) at (-0.5 in,0 in) {$T_1$};
			\node[shape=circle, draw=dark-green, minimum size = 0.3in, text=dark-green] (T2) at (0.5 in,0 in) {$T_2$};
			\node[shape=circle, draw=dark-green, minimum size = 0.3in, text=dark-green] (T3) at (1.6 in,0 in) {$T_3$};
			
			\node[draw = none, text = dark-green] (dots) at (2.1 in,0 in) {$\cdots$};
			\node[shape=circle, draw=dark-green, minimum size = 0.3in, text=dark-green] (Tn) at (2.6 in,0 in) {$T_n$};
			
			\node[shape=circle, draw=lightgray, minimum size = 0.3in, text=lightgray] (x1) at (-2.4 in, 1.5 in) {$\ell_1$};
			\node[shape=circle, draw=black, minimum size = 0.3in] (x2) at (-1.6in , 1.5 in) {$\ell_2$};
			\node[shape=circle, draw=black, minimum size = 0.3in] (x3) at (-0.9 in, 1.5 in) {$\ell_3$};
			\node[shape=circle, draw=black, minimum size = 0.3in] (x4) at (0.0 in, 1.5 in) {$\ell_4$};
			\node[shape=circle, draw=lightgray, minimum size = 0.3in, text=lightgray] (x5) at (0.8in , 1.5 in) {$\ell_5$};
			\node[shape=circle, draw=dark-green, minimum size = 0.3in, text=dark-green] (x6) at (1.5in , 1.5 in) {$\ell_6$};
			\node[draw=none, minimum size = 0.3in, text=dark-green] (x7) at (2.0 in, 1.5 in) {$\cdots$};
			\node[shape=circle, draw=dark-green, minimum size = 0.3in, text=dark-green] (x7) at (2.5 in, 1.5 in) {$\ell_m$};

			\node[shape=circle, fill=black,minimum size = 0.06in, inner sep = 0.01in, outer sep = 0.01in, label=left:$\phi_1^{\text{proj}}$] (m1) at ($ 0.5*(T0) + 0.5*(x2) $ ) {};
			\node[shape=circle, fill=black,minimum size = 0.06in, inner sep = 0.01in, outer sep = 0.01in, label=above:$\phi_2^{\text{proj}}$] (m2) at ($ 0.5*(x2) + 0.5*(T1) $ ) {};
			\node[shape=circle, fill=black,minimum size = 0.06in, inner sep = 0.01in, outer sep = 0.01in, label=below :$\phi_3^{\text{proj}}$] (m3) at ($ 0.5*(T0) + 0.5*(x3) $ ) {};
			\node[shape=circle, fill=black,minimum size = 0.06in, inner sep = 0.01in, outer sep = 0.01in, label=below:$\phi_5^{\text{proj}}$] (m5) at ($ 0.75*(x4) + 0.25*(T0) $ ) {};
			\node[shape=circle, fill=black,minimum size = 0.06in, inner sep = 0.01in, outer sep = 0.01in, label=below:$\phi_6^{\text{proj}}$] (m4) at ($ 0.5*(x3) + 0.5*(T1) $ ) {};
			\node[shape=circle, fill=black,minimum size = 0.06in, inner sep = 0.01in, outer sep = 0.01in, label=below:$\phi_7^{\text{proj}}$] (m6) at ($ 0.5*(x4) + 0.5*(T1) $ ) {};
			\node[shape=circle, fill=black,minimum size = 0.06in, inner sep = 0.01in, outer sep = 0.01in, label=above left:$\phi_8^{\text{proj}}$] (m7) at ($ 0.5*(x4) + 0.5*(T2) $ ) {};
			\node[shape=circle, fill=dark-green,minimum size = 0.06in, inner sep = 0.01in, outer sep = 0.01in, label=above:\color{dark-green}$\phi_9^{\text{proj}}$] (m8) at ($ 0.5*(x6) + 0.5*(T1) $ ) {};

			\node[shape=circle, fill=dark-green,minimum size = 0.06in, inner sep = 0.01in, outer sep = 0.01in, label=below:\color{dark-green}$\phi_{q-1}^{\text{proj}}$] (mqminus1) at ($ 0.5*(x4) + 0.5*(Tn) $ ) {};
			\node[shape=circle, fill=dark-green,minimum size = 0.06in, inner sep = 0.01in, outer sep = 0.01in, label=right:\color{dark-green}$\phi_{q}^{\text{proj}}$] (mq) at ($ 0.5*(x6) + 0.5*(Tn) $ ) {};
			\node[draw = none] (mdots) at ($ 0.5*(m8) + 0.5*(mqminus1) $ ) {$\cdots$};

			\path[-] (p0) edge (T0);
			\path[-] (p1) edge (T1);
			\path[-] (x2) edge (m1);
			\path[-] (x3) edge (m3);
			\path[-] (x2) edge (m2);
			
			\path[-] (m1) edge (T0);
			\path[-] (m3) edge (T0);
			\path[-] (m5) edge (T0);
			\path[-] (m5) edge (x4);
			\path[-] (m2) edge (T1);
			\path[-] (m4) edge (T1);
			\path[-] (m6) edge (T1);
			\path[-] (x4) edge (m7);
			\path[-] (m7) edge (T2);
			\path[draw=dark-green,-] (m8) edge (T1);
			\path[-] (x3) edge (m4);
			\path[-] (x4) edge (m6);
			\path[draw=dark-green,-] (x6) edge (m8);
			\path[draw=dark-green,-] (x4) edge (mqminus1);
			\path[draw=dark-green,-] (mqminus1) edge (Tn);
			\path[draw=dark-green,-] (x6) edge (mq);
			\path[draw=dark-green,-] (mq) edge (Tn);
			\path[draw=dark-green,-] (T1) edge (mdots);
			\path[draw=dark-green,-] (T2) edge (mdots);
			\path[draw=dark-green,-] (T3) edge (mdots);

			
		\end{tikzpicture}
	\end{tabular}
	\caption{Front-end pipeline factor-graph following the initialization of SLAM.}
	\label{fig:frontend_pipeline}
\end{figure*}

\subsection{Map initialization}

Initialization of the map is delayed until time index $k=1$, at which point at least two images of the target with sufficient parallax are captured.
Local image features $\Upsilon_0$ and $\Upsilon_1$ are extracted and undergo data association, with outlier rejection, producing a set of 2D-2D correspondences. 
A strict outlier rejection criterion is used to obtain a subset of reliable correspondences.
It is now possible to apply a typical 8-point algorithm~\cite{hartley2004multiple} using the inlier 2D-2D correspondences to find a guess pose transformation $\hat{T}^\circ \in \mathrm{SE}(3)$ such that
\begin{align*}
	\hat{T}^{\circ} \triangleq \begin{bmatrix}
		\hat{R}^{\circ} & \hat{\vec{t}}^{\circ} \\
		0 & 1
	\end{bmatrix},
\end{align*}
where $\hat{R}^{\circ} \in \mathrm{SO}(3)$ and $\hat{\vec{t}}^{\circ} \in \mathbb{R}^{3}$, describe the estimated change in relative orientation and relative position between the poses of the camera frame at time indices $k=0$ and $k=1$. 
From the frame kinematics and composition rules, and knowing that $\hat{R}^{\circ} = \hat{R}_{\fC_0\fC_1}$, it follows that $\hat{R}_{\fG_1\fS_1} = \hat{R}_{\fG_1\fG_0}\hat{R}_{\fG_0\fS_0}R_{\fS\fC}\hat{R}^{\circ}R_{\fC\fS}$.
We note that $\hat{R}_{\fG_0\fG_1} \approx \Exp{-\hat{\vec{\omega}}_{\fG_0\fI}^{\fI} \Delta t_0}$. 
It follows that $\hat{R}_{\fG_1\fS_1} \approx \Exp{\hat{\vec{\omega}}_{\fG_0\fI}^{\fG_0} \Delta t_0}\hat{R}_{\fG_0\fI}\hat{R}_{\fI\fS_0}^{\mathrm{m}}R_{\fS\fC}\hat{R}^{\circ}R_{\fC\fS}$.
Additionally, we know that  
\begin{align}     \label{eq:scale_amb}
	\vec{t}^{\circ} &= \lambda \vec{r}_{\mathrm{C}_1\mathrm{C}_0}^{\fC_0} \nonumber \\
	&=  \lambda R_{\fC\fS}(\vec{r}_{\mathrm{C}\mathrm{S}}^{\fS}  + R_{\fS_0\fG_0}R_{\fG_0\fG_1}\vec{r}_{\mathrm{S}_1\mathrm{G}}^{\fG_1} - (\vec{r}_{\mathrm{C}\mathrm{S}}^{\fS}  + R_{\fS_0\fG_0}\vec{r}_{\mathrm{S}_0\mathrm{G}}^{\fG_0} )) \nonumber \\
	&= \lambda R_{\fC\fS}\hat{R}_{\fS_0\fG_0}\left(\Exp{\hat{\vec{\omega}}_{\fG_0\fI}^{\fG_0} \Delta t_0}\vec{r}_{\mathrm{S}_1\mathrm{G}}^{\fG_1} - \vec{r}_{\mathrm{S}_0\mathrm{G}}^{\fG_0}\right),
\end{align} 
with $\lambda>0$, an unknown scaling factor.

Thus, in order to establish reliable relative pose estimates $\hat{T}_0$ and $\hat{T}_1$, a good prior knowledge of the small-body inertial angular velocity is required.
Additionally, the ambiguity in the scale parameter $\lambda$ still remains.
Typically, the scale can be established using off-nadir altimeter measurements over a range of orbital configurations in a multi-arc solution, as performed by the OSIRIS-REx mission around asteroid Bennu~\cite{goossens2021mass}.
However, the latter solution, in its current format, is not amenable to on-the-fly autonomy, as the batch estimation process requires a compilation of data from multiple mission arcs.
Furthermore, altimeter measurements have to be used at close range to the target small-body, again restricting autonomy since judiciously pre-designed maneuvers have to be executed to first obtain the necessary altimeter measurements at close range.
If a good initial estimate of the angular velocity $\xw_0$ of the small-body is available, the map scale can be determined by combining DSN-like inertial position measurements with the orientation matrix and position vectors extracted from the relative poses $T_0$ and $T_1$, as shown in Equation~(\ref{eq:scale_amb}).
The availability of such prior knowledge currently depends heavily on Earth-based and pre-encounter measurements of the target small-body. 
Regarding this matter, as detailed in \cite{goossens2021mass}, the ambiguity in scale due to the parameter $\lambda$ is determining in the estimation of an appropriate spherical gravity parameter term $\mu_a$.
An error in $\mu_a$ will then affect the propagation step of the smoothing process.
For the purpose of this work, we have assumed that the scale parameter $\lambda$ is well known.
Future work will develop novel autonomous methods to estimate the scale.

Having an estimate of poses $\hat{T}_0$ and $\hat{T}_1$, we compute the triangulation of the landmarks $\Psi_1$ using measurements $\Upsilon_0$ and $\Upsilon_1$.
We then generate guess values for the estimated landmark positions $\left\lbrace \vec{r}_{\mathrm{L}\oG}^\fG\right\rbrace_{\mathrm{L} \in \Psi_1}$.
At this step, all appropriate factors are inserted based on the 2D-to-3D correspondences, resulting in the factor graphs illustrated in Figure~\ref{fig:init_steps_SLAM}, where the shorthands $\ell_i \triangleq \vec{r}_{\mathrm{L}_i\oG}^{\fG}, \, \mathrm{L}_i\in \Psi_k$ and $T_k \triangleq T_{\fG_k \fS_k}, \, k=0,\ldots,n$ are used for brevity.

\subsection{Front-end}

The front-end system includes feature detection and matching and encodes the structure of the SLAM problem in a factor graph using data structures based on the definitions in Section \ref{subsec:asteroid_nav_defs}. 

We use ORB features~\cite{Rublee2011}, as they perform well in practice and they are fast to compute. 
These are also good placeholders for more robust automatic features to be implemented in the future. 
For matching, we use brute force nearest neighbor search based on the Hamming distance for binary feature descriptors. 
We reject 2D-2D matches with a distance ratio greater than 0.85~\cite{lowe2004distinctive}. 
We also enforce a essential matrix constraint as a geometric check for matched features.
We track landmarks by comparing the current frame $n$ to previous frame 2D-to-2D correspondences against past frame 2D-to-3D correspondences.
The feature-landmark matched pairs then undergo a reprojection error test, after which the surviving pairs establish new projection factors to be inserted into the graph, relating to the most recent frame $n$. 
If there are enough tracked features from time index $k=n-1$, then visual tracking is successful and the well-known P$n$P algorithm~\cite{lepetit2009epnp} is used to guess the camera pose $\Tilde{T}_{\fG_n\fC_n}$ value from matched correspondences.
Alternatively, when \texttt{RelDyn} factors are included and if the P$n$P solution is of poor quality, we use the prediction from the motion model at time $t_n$ to guess a new camera pose $\Tilde{T}_{\fG_n\fC_n}$.
Leveraging the latest guess pose $\Tilde{T}_{\fG_n\fC_n}$, a guess value for 3D position for each of the newly detected features is generated using triangulation.
Additionally, we delay the insertion of new landmarks into the graph based on whether the number of times the associated landmark has been seen is above a predetermined threshold of 3.
The front-end system produces incrementally growing factor graphs, as is illustrated in 
Figure~\ref{fig:frontend_pipeline}.

\subsection{Back-end}

The underlying structure of the navigation problem is thus captured by encoding visual SLAM measurement constraints, star tracker orientation measurement constraints, Earth-relative inertial position measurement constraints as factors and \texttt{RelDyn} factors in a single factor graph.
The choice of this graph formulation is predicated on the fact that a factor graph, as an undirected graph, readily explains the relationships between unknowns since its incidence matrix directly relates to the $R$ matrix in the $QR$ factorization of the square root information matrix~\cite{kaess2012isam2}.
In addition, the graph lends itself naturally to incremental growth.
For every new frame inserted, the graph is incrementally augmented with the new variables and factors by the front-end. 

In practice, the minimization associated with the small-body relative navigation problem is performed incrementally in the factor graph framework using the iSAM2~\cite{kaess2012isam2} algorithm.
The iSAM2 algorithm integrates these new measurement constraints as new factors and performs inexpensive Givens rotations~\cite{kaess2012isam2} on the existing $R$ matrix to perform further variable elimination without recalculating the full variable elimination.
The back-end system evaluates the losses at each factor, computes the associated Jacobians at the guess values $\lbrace \tilde{x}_k\rbrace_{k=0}^{n}$ and $\lbrace \tilde{\ell}_i\in \mathcal{L}_n \rbrace$, and performs the minimization for inference~\cite{dellaert2017factor}, yielding a new estimate solution.
Note that since the process noise covariance related to the \texttt{RelDyn} factor is not fixed, but is rather the solution to a dynamical equation, the out-of-the-box GTSAM factor template had to be modified to allow for a variable process noise covariance to be incorporated accordingly.

\subsection{Loop closure}
\label{sec:loop-closure}
To perform loop closure, we leverage the bag-of-words representation developed in~\cite{galvezTRO2012}. 
We convert each image $i=1,\ldots,N$ to a bag-of-words vector $\vec{v}_i$ and compute the similarity metric
\begin{align}
	s(\mathbf{v}_i,\mathbf{v}_j) \triangleq 1 - \frac{1}{2} \bigg|  \frac{\mathbf{v}_i}{|\mathbf{v}_i|}-\frac{\mathbf{v}_j}{|\mathbf{v}_j|} \bigg|.
\end{align}
We compare all prior images that are at least $10$ frames away from the current frame. 
When the similarity score for two images is greater than a threshold $\eta$, we perform an additional geometric check and then add a factor between the poses corresponding to the detected loop. 

\subsection{Computation of the \emph{\texttt{RelDyn}} factor residual}
\label{sec:impl_RelDyn}

We perform on-manifold discrete Crouch-Grossman geometric integration as detailed in~\cite{andrle2013geometric} to obtain accurate predictions which also respect the constraints of the relative rotation matrix $Q(t) \in \mathrm{SO}(3)$. 
Given $\left\lbrace t_k \right\rbrace_{k=0}^{N}$, 
for every $k=0,\ldots,N-1$, let $\Delta t_k \triangleq t_{k+1} - t_k$, we use an $N_s$-stage Crouch-Grossman geometric integration scheme and compute the propagated quantities
\begin{align}
	\tilde{\Omega}_{k}^{\mathrm{prop}}&\triangleq \prod_{i=1}^{N_s}
	\exp\left(\sk{\bi{i} \Delta t_k \tilde{f}_{Q}^{(i)} }\right),\\
	\begin{bmatrix}
		\Delta \tilde{\vec{w}}_{k}^{\mathrm{prop}}\\
		\Delta \tilde{\vec{r}}_{k}^{\mathrm{prop}}\\
		\Delta \tilde{\vec{v}}_{k}^{\mathrm{prop}}\\
	\end{bmatrix}
	&\triangleq
	\sum_{i=1}^{N_s} \bi{i} \Delta t_k
	\begin{bmatrix}
		\tilde{f}_{\vec{w},k}^{(i)}  \\
		\tilde{f}_{\xr,k}^{(i)}  \\
		\tilde{f}_{\xv,k}^{(i)}
	\end{bmatrix}
\end{align},
whereby the shorthands
\begin{align}
	\tilde{f}_{*,k}^{(i)} &\triangleq f_{*}\left(\tilde{x}_k^{(i)}, \hat{u}_{k}^{(i)}, p\right),\quad *=Q,\vec{r},\vec{w},\vec{v}, \\
	\tilde{x}_k^{(i)} &\triangleq \left(\tilde{Q}_{k}^{(i)}, \tilde{\vec{w}}_{k}^{(i)}, \tilde{\vec{r}}_{k}^{(i)}, \tilde{\vec{v}}_{k}^{(i)}\tilde{}\right), \\
	\hat{u}_k^{(i)} &\triangleq \hat{u}(t_k + \ci{i}\Delta t_k), \\
	\txQki{i} &\triangleq \txQk{k}\prod_{j=1}^{i-1} \exp\left(\sk{ \aij{i}{j} \Delta t_k  \tilde{f}_{Q,k}^{(j)} }\right),\\
	\begin{bmatrix}
		\txwki{i} \\
		\txrki{i} \\
		\txvki{i}
	\end{bmatrix}
	& \triangleq
	\begin{bmatrix}
		\txwk{k} \\
		\txrk{k} \\
		\txvk{k}
	\end{bmatrix} + \sum_{j=1}^{i-1}\aij{i}{j}\Delta t_k  
	\begin{bmatrix}
		\tilde{f}_{\vec{w},k}^{(j)}  \\
		\tilde{f}_{\xr,k}^{(j)}  \\
		\tilde{f}_{\xv,k}^{(j)}
	\end{bmatrix}
\end{align}
are defined and computed for every $i=1,\ldots,N_s$.
The coefficients $\aij{i}{j}, \bi{i}, \ci{i}, \, j< i=1,\ldots,N_s$ are obtained from an appropriate Butcher table~\cite{andrle2013geometric}, an example of which is given in 
Table~\ref{table:cg3_butcher_table} below.
\begin{table}[h]
	\caption{Crouch-Grossman 3rd Order 3-Stage Butcher Table.}
	\label{table:cg3_butcher_table}
	\centering
	\begin{tabular}{c|cccc}
		$c_1$ & $a_{11}$ & $a_{12}$ & $\ldots$ & $a_{1,N_s}$ \\ 
		$c_2$ & $a_{21}$ & $a_{22}$ & $\ldots$ & $a_{2,N_s}$ \\ 
		$\vdots$ & $\vdots$ & $\vdots$ & $\ddots$ & $\vdots$ \\ 
		$c_{N_s}$ & $a_{N_{s}1}$ & $a_{N_{s}2}$ & $\ldots$ & $a_{N_s,N_s}$ \\ \hline
		$ $ & $b_1$ & $b_2$ &$\ldots$ & $b_{N_{s}}$
	\end{tabular}
	\\
	$\big\uparrow$
	\\
	\begin{tabular}{c|cccc}
		$0$     & $ $       & $ $      & $ $ \\ 
		$3/4$   & $3/4$     & $ $      & $ $ \\ 
		$17/24$ & $119/216$ & $17/108$ & $ $ \\ \hline
		$ $     & $13/51$   & $-2/3$   & $24/17$
	\end{tabular}

\end{table}

The \texttt{RelDyn} residual $\epsilon_{k}^{\mathrm{RelDyn}}(\tilde{x}_{k}, \tilde{x}_{k+1})$ is now computed such that
\begin{align}
	&\epsilon_{k}^{\mathrm{RelDyn}}(\tilde{x}_{k}, \tilde{x}_{k+1}) \nonumber \\
	&\triangleq \begin{bmatrix}
		\log \left( \left(\tilde{Q}_{k}^\top \tilde{Q}_{k+1}\right)^\top \tilde{\Omega}_{k}^{\mathrm{prop}}\right)^\vee \\
		\tilde{\vec{w}}_{k+1} - \tilde{\vec{w}}_k - \Delta \tilde{\vec{w}}_{k}^{\mathrm{prop}}\\
		\tilde{\vec{r}}_{k+1} - \tilde{\vec{r}}_k - \Delta \tilde{\vec{r}}_{k}^{\mathrm{prop}} \\
		\tilde{\vec{v}}_{k+1} - \tilde{\vec{v}}_k - \Delta \tilde{\vec{v}}_{k}^{\mathrm{prop}}
	\end{bmatrix}.
\end{align}

\begin{figure}[ht]
	\centering
	\def\Elevation{25} 
	\def\Azimuth{125} 
	\includegraphics[width=3.5in]{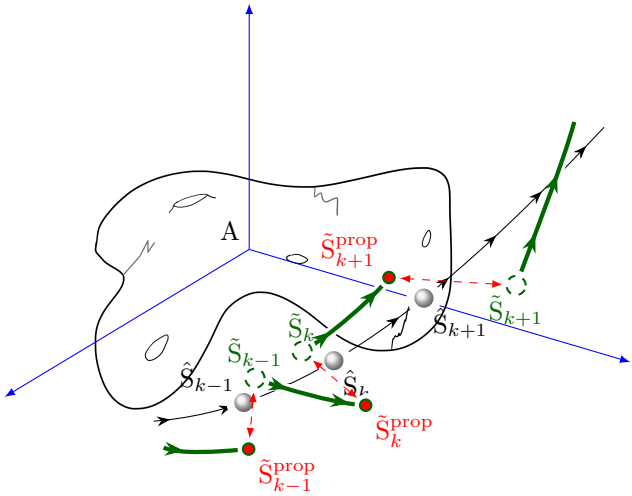}
	\caption{Illustration of the \texttt{RelDyn} factor residual.}
\end{figure}

\subsection{Process noise propagation}
\label{subsec:process_noise_impl}

To compute $P_k$ we introduce the matrix $\Lambda_k$, such that, given any pair $(\hat{x}_k,\hat{u}_k)$,
\begin{align*}
	&\Delta_X (x(t_{k+1}), \hat{x}_k) \nonumber \\
	&=  \Phi(t_{k+1}, t_k) \Delta_X (x(t_k), \hat{x}_{k}) \nonumber \\ 
	&+ \int_{t_k}^{t_{k+1}}\Phi(t_{k+1},\tau) L(\hat{x}(\tau),p)\diff \vec{\varepsilon}(\tau), \quad \hat{x}(t_k) = \hat{x}_k, \, \hat{u}(t_k) = \hat{u}_k \\
	&= \Phi(t_{k+1}, t_k) \Delta_X (x(t_k), \hat{x}_{k}) + \Lambda_k \varepsilon_k,
\end{align*}
where $\vec{\varepsilon}_k \sim \mathcal{N}(\vec{0}, \mathrm{I}_{12})$ and $\dot{\Phi}(t,t_k) = F(\hat{x}(t), \hat{u}(t), p)\Phi(t,t_k), \, \Phi(t_k,t_k) = \mathrm{I}_{12}$. 
We may assume that $P_k = \Lambda_k \Lambda_k^\top$.
It follows that
\begin{align*}
	P_k = \int_{t_k}^{t_{k+1}} \Phi(t_{k+1}, \tau)L(\hat{x}(\tau),p)L^\top(\hat{x}(\tau),p)\Phi^\top(t_{k+1}, \tau)\diff \tau, \nonumber \\
	\quad \hat{x}(t_k) = \hat{x}_k, \, \hat{u}(t_k) = \hat{u}_k.
\end{align*}
We further approximate the linearized system by assuming that the matrices $F, L$ shown in Section~\ref{sec:smoothability} are piecewise constant over the interval $[t_k, t_{k+1}]$, i.e., $F_k \triangleq F(\hat{x}(t_k), \hat{u}(t_k), p)$, $L_k \triangleq L(\hat{x}(t_k),\hat{u}(t_k),p)$.
Then, we construct the following matrix~\cite{vanLoanMatrixExponential}
\begin{align*}
	\begin{bmatrix}
		F_k & L_k L_k^\top \\
		0_{12\times 12}   & -F_k^{\top}
	\end{bmatrix},
\end{align*}
take the matrix exponential, yielding
\begin{align*}
	\mathrm{exp}
	\left(
	\begin{bmatrix}
		F_k & L_k L_k^\top \\
		0_{12\times 12}   & -F_k^{\top}
	\end{bmatrix} \Delta t_k
	\right)
	= \begin{bmatrix}
		\Phi_k & \Gamma_k \\
		0_{12\times 12}  & \Phi_k^\top
	\end{bmatrix},
\end{align*}
and we extract from the right hand side the desired submatrices, i.e., the discrete state transition matrix $\Phi_k$ of the piecewise-constant linearization, as well as the covariance of the discretized process noise $\Lambda_k \vec{\varepsilon}_k$, given as $P_k = \Gamma_k \Phi_k^\top$.

\section{Validation on legacy mission imagery}
\label{sec:experiments}

In this section, we discuss the design of the validation process for the proposed algorithm. 
We provide the details of a validation test case using imagery and data pertaining to a previously flown mission.
Since access to real ground-truth data for a flown mission is impossible, in the next session we also test our algorithm using data generated in a controlled lab environment, allowing us to compare the results of the estimation problem against actual ground-truth. 


\subsection{PDS asteroid imagery dataset}
\label{sec:valid_PDS}

\begin{figure*}[tpb]
	\centering
	\includegraphics[width=6in, clip=true, trim=0.4in 0.7in 0.3in 0.5in]{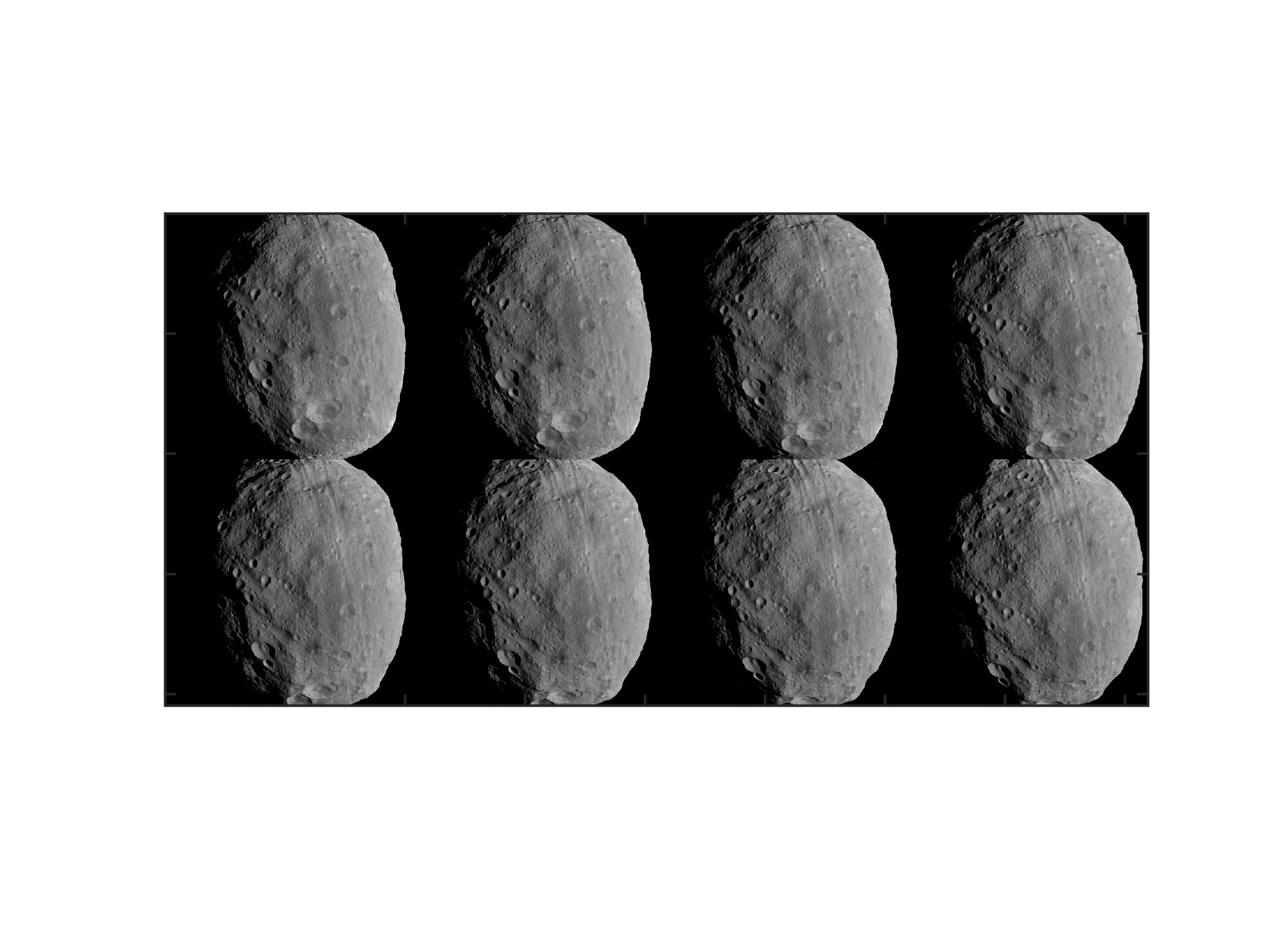}
	\caption{Eight time-consecutive images from the RC3 observation phase of the DAWN mission.}
	\label{fig:sample_images}
\end{figure*}

We use real imagery~\cite{vestaImages} of Asteroid (4) Vesta acquired during the Rotation Characterization 3 (RC3) observation phase of the Dawn mission~\cite{russell2011}, and archived in the Small Bodies Node of the NASA Planetary Data System (PDS), to validate the algorithm.
In the chosen sequence, the $1024 \times 1024$ images were captured, while the spacecraft performed one apparent revolution around Vesta in the asteroid body-fixed frame, with a mean orbital radius of $5480$ km. 
The images thus provide a spatial resolution of $0.5\ \text{km/pixel}$ of the surface (see Figure~\ref{fig:sample_images} for sample images). 
This sequence, therefore, enables possible loop closures to be tested as well.
Two tests were conducted for the PDS dataset, each with the same prior uncertainty on poses $T_{\fG_0\fS_0}$ and $T_{\fG_1\fS_1}$, i.e., the covariance $\Sigma_{\mathrm{T},0}$ of the initial pose measurements, but with the \texttt{RelDyn} factors excluded in the first case and included in the second.
The values are chosen as $\Sigma_{\mathrm{R}}^{\mathrm{m}} = \sigma_{\mathrm{R}}^2 \mathrm{I}_3$ and $\Sigma_{\mathrm{r}}^{\mathrm{m}} = \sigma_{\mathrm{r}}^2 \mathrm{I}_3$, with  $\sigma_\mathrm{R} = 1 \times 10^{-5}\,\mathrm{rad}$ and $\sigma_\mathrm{r} = 0.05\,\mathrm{km}$.
Note that, for simplicity, the pose discrepancy $T_{\fA\fG}$ is assumed to be known in these trials.
Additionally, as explained in Section~\ref{sec:introduction}, the dynamical parameters, mainly the spherical gravity term $\mu_a$ and the initial spin state $\hat{\vec{w}}_0$, which affect the relative motion, are assumed to be known with relatively low uncertainty from pre-encounter modeling.
The intrinsic parameters of the Dawn Framing Camera (FC) were taken to be those computed during calibration~\cite{russell2011}.
In the conducted tests, the algorithm was set to extract and describe 1400 ORB features where 8 scale pyramid levels are explored.

To validate the estimated spacecraft relative trajectory, we use the archival SPICE kernel datasets maintained by NASA's Navigation and Ancillary Information Facility (NAIF). 
For the missions archived, SPICE kernels may be queried to provide \textbf{S}pacecraft ephemeris and asteroid (\textbf{P}lanetary) ephemeris as a function of time, as well as \textbf{I}nstrument descriptive data, \textbf{C}amera orientation matrix data and \textbf{E}vents information, such as mission phases. 

To validate the reconstructed map, we use an archival shape model of Vesta~\cite{preusker2016dawn} courtesy of the PDS Small Bodies Node. 
This shape model was derived using stereo photogrammetry (SPG) from a subset of DAWN mission Framing Camera 2 (FC2) images captured during the High-Altitude Mapping Orbit (HAMO) mission segment.
The model comprises approximately 100k vertices and 197k triangular faces and is shown in Figure~\ref{fig:sample_images}.

\subsection{Quantitative evaluation of trajectory estimation result and reconstructed map}
\label{sec:Section6.2}

Although knowledge about the spacecraft's relative position in the asteroid body-fixed frame, given by $\xin{\vec{r}_{\oS\oA}}{\fA}$, is ultimately sought after, analyzing the errors in the spacecraft's  local-horizontal-local-vertical frame $\mathcal{L}$ better reveals the performance of the algorithm. 
By virtue of the assumed true position vector $\vec{r}_{\oS\oA}$ and true velocity $\vec{v}_{\oS\oA}$ vector, define the frame $\mathcal{L} \triangleq ( \oS; \left\lbrace \uvec{\ell}_i\right\rbrace_{i=1}^{3}) $ such that $\uvec{\ell}_3 \triangleq {\vec{r}_{\oS\oA}}/{\|\vec{r}_{\oS\oA}\|},\, \uvec{\ell}_2 \triangleq ({\vec{r}_{\oS\oA} \times \vec{v}_{\oS\oA}})/{\|\vec{r}_{\oS\oA} \times \vec{v}_{\oS\oA}\|},\, \uvec{\ell}_1 = \uvec{\ell}_2 \times \uvec{\ell}_3$, 
leading to the definition of the rotation $R_{\mathcal{A}\mathcal{L}} = \begin{bmatrix}
	\xin{\uvec{\ell}_1}{\fA} & \xin{\uvec{\ell}_2}{\fA} & \xin{\uvec{\ell}_3}{\fA}
\end{bmatrix}$. For each time index $k= 1,\ldots,n$ we compute the position error $\delta\vec{r}_k \triangleq R_{\mathcal{A}_k\mathcal{L}_k}^\top (\xin{\hat{\vec{r}}_{\oS_k\oA_k}}{\fA_k} -  \xin{\vec{r}_{\oS_k\oA_k}}{\fA_k})$,
where $\xin{\hat{\vec{r}}_{\oS_k\oA_k}}{\fA_k} = \hat{R}_{\fA\fG}(\xin{\hat{\vec{r}}_{\oS_k\oG_k}}{\fG_k} - \xin{\hat{\vec{r}}_{\oA\oG}}{\fG})$, with constant parameters $\hat{R}_{\fA\fG}$ and $\xin{\hat{\vec{r}}_{\oA\oG}}{\fG}$ assumed to be known.
The error in relative orientation is then better described by making use of the SE(3) logarithm map $\log : \mathrm{SE}(3) \rightarrow \mathrm{se}(3)$, as detailed in Section~\ref{sec:notation}, to produce the error between the ground-truth NAIF SPICE pose $T_{\fS_k\fA_k}$ and the estimated pose $\tilde{T}_{\fS_k\fA_k}$. 
The results of this evaluation are presented in Section~\ref{sec:result} for the DAWN RC3 sequence and in Section~\ref{sec:eval_map} for the in-lab experiment sequence.

Given the map $\Psi_N$ at the final time $k=N$ and the set of ground truth 3D shape model vertices $\mathcal{V}$, we evaluate the quality of the estimated landmark by computing the distances $\left\lbrace d(\mathrm{L},\mathcal{V})\right\rbrace_{\mathrm{L}\in \Psi_N}$, where
\begin{equation} \label{eq:dd}
	d(\mathrm{L},\mathcal{V}) \triangleq \min_{\mathrm{V} \in \mathcal{V}} \|\vec{r}_{\mathrm{L}\oA}^\fA - \vec{r}_{\mathrm{V}\oA}^\fA\|_2,
\end{equation}
which in our case minimizes the 2-norm.
The results of this evaluation are presented in Figure \ref{fig:DAWN_seqence_landmarks}, where the estimated landmarks are colored as a function of their distance to the closest point in the ground-truth set of vertices.
Note that this distance metric is one-sided.
Thus, choosing to instead search over the set $\Psi_N$ would yield different distance values $\left\lbrace d(\mathrm{V},\Psi_N)\right\rbrace_{\mathrm{V} \in \mathcal{V} }$.
Nevertheless, given the much higher vertex density of the ground truth shape model as compared to the estimated landmarks, we deem the described point-to-set distance $d(\mathrm{L},\mathcal{V}), \, \mathrm{L}\in \Psi_N$ to be an appropriate measure of the deviation of our solution landmarks from the 3D shape.

\subsection{NASA PDS imagery dataset results}
\label{sec:result}

\begin{figure}[ht]
	\centering
	\includegraphics[width=3.40in, clip=true, trim=0.1in 0.8in 0.1in 0.8in]{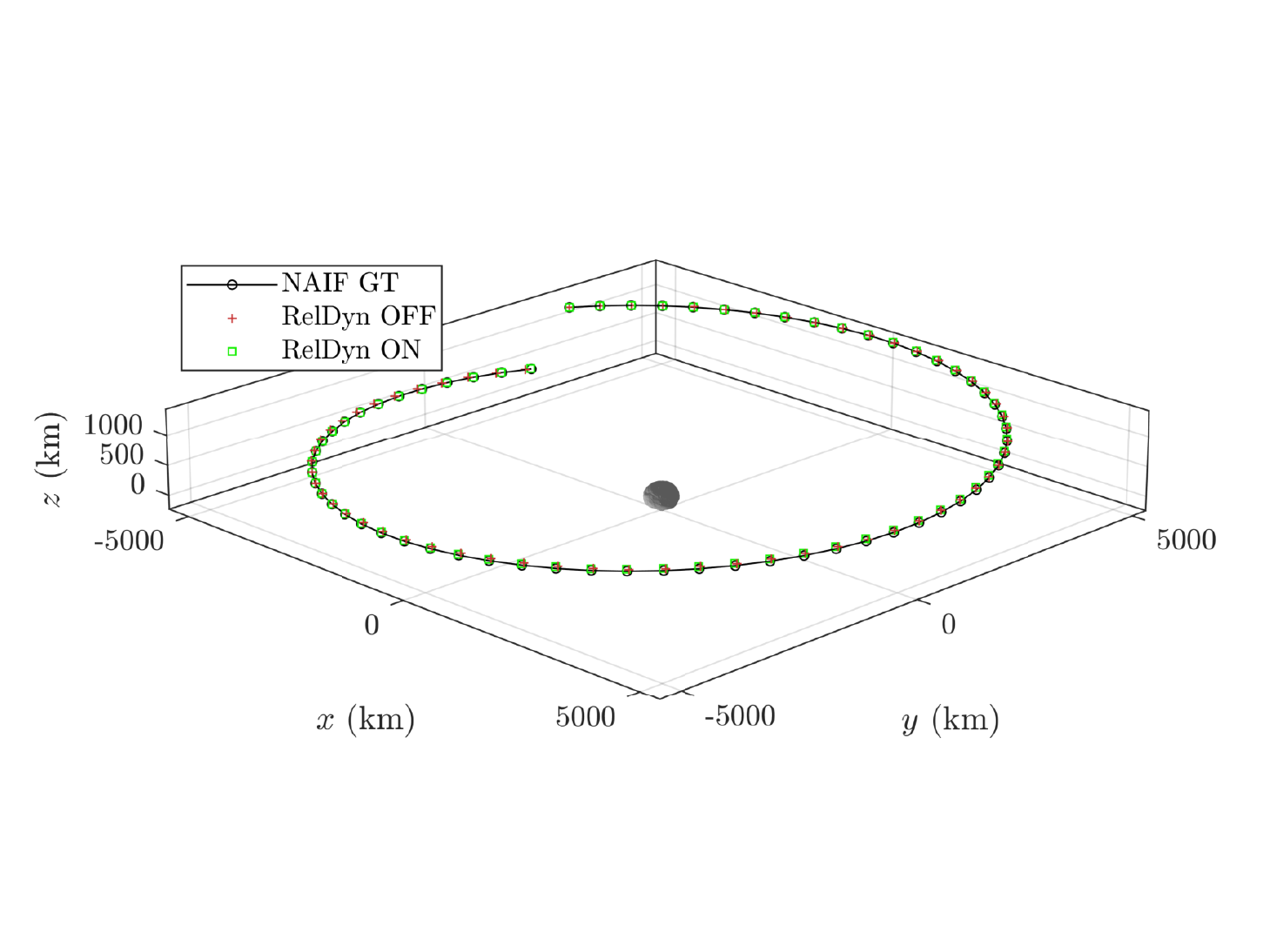}
	\caption{3D trajectory estimate vs NAIF SPICE kernel ground truth for DAWN RC3 sequence.}
	\label{fig:DAWN_sequence_3d_plot}
\end{figure}

\begin{figure}[ht]
	\centering
	\includegraphics[width=3.3in, clip=true, trim=0.3in 0.0in 3.3in 0.0in]{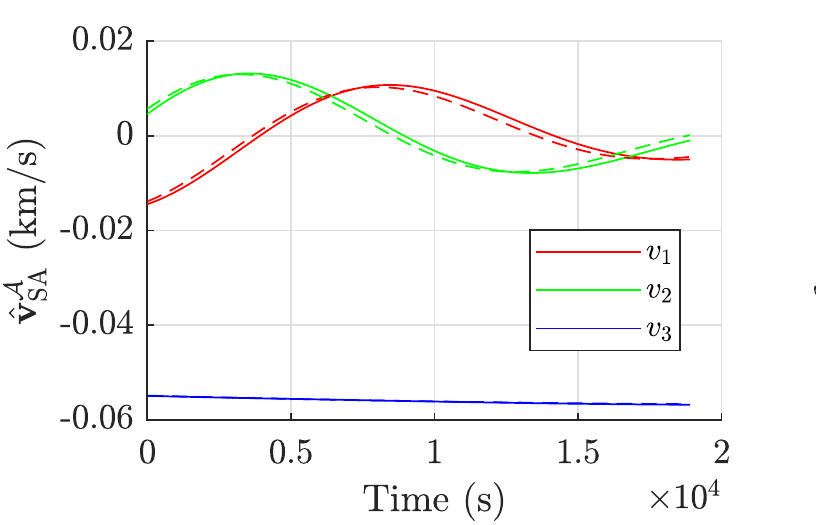}
	\includegraphics[width=3.3in, clip=true, trim=3.3in 0.0in 0.3in 0.0in]{figures/velocities_DAWN_RC3.pdf}
	\caption{Relative velocity estimates vs NAIF SPICE kernel ground truth for DAWN RC3 sequence.}
	\label{fig:DAWN_sequence_velocities}
\end{figure}

In this section, we illustrate and discuss the results of the real-mission dataset.
We first visualize in Figure~\ref{fig:DAWN_sequence_3d_plot} the 3D trajectory derived from the relative poses $\lbrace \hat{T}_{ \fA_{k}\fS_{k} }\rbrace_{k=0}^{N} $ overlayed on the trajectory extracted from the DAWN mission SPICE kernel poses, assumed to be the true poses $\lbrace T_{ \fA_{k}\fS_{k} }\rbrace_{k=0}^{N}$ for the DAWN legacy mission case.
When the \texttt{RelDyn} factors are included, we additionally obtain estimates of the linear and angular velocities, which we show in overlay to ground truth NAIF SPICE velocity estimates in  Figure~\ref{fig:DAWN_sequence_velocities}.

\begin{figure}[ht]
	\centering
	\includegraphics[width=3.4in, clip=true, trim=0.3in 0.5em 0in 0.6em]{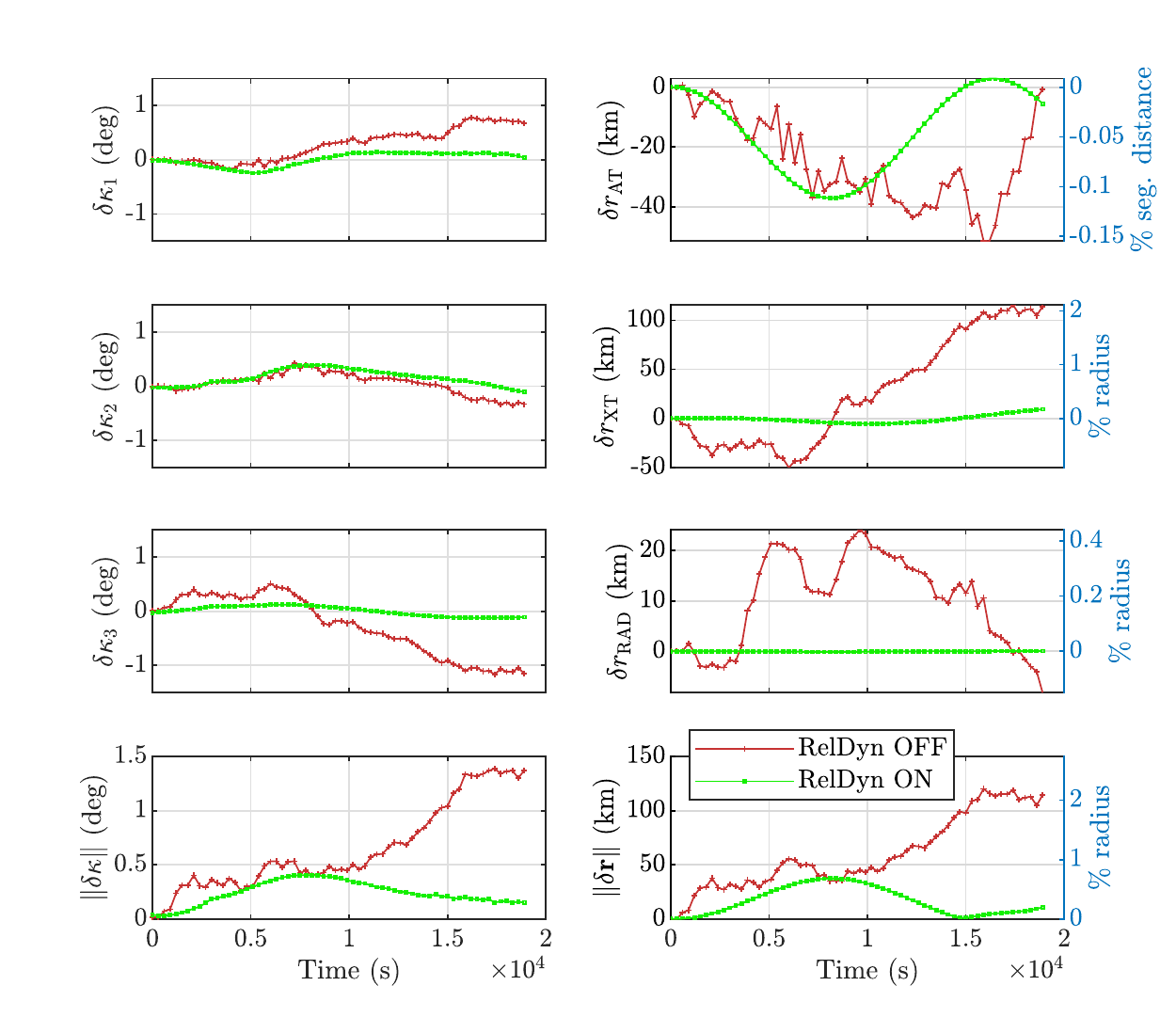}
	\caption{Estimated pose errors with respect to NAIF SPICE kernel ground truth for DAWN RC3 sequence.}
	\label{fig:DAWN_sequence_time_history}
\end{figure}

Considering that the orbit radius is roughly 5470 km throughout the sequence, the error in the radial direction  ($\delta r_{\mathrm{RAD}}$), as shown in Figure~\ref{fig:DAWN_sequence_time_history}, yields an error of 0.2\% (10~km) on average, with a worst case of 0.4\% (19.2 km), while for the case including motion priors, it yields an accuracy of 0.0004\% (50~m) on average and a worst case of 0.003\% (140~m). Note that the estimated and ground-truth trajectories are close, as illustrated in Figure~\ref{fig:DAWN_sequence_3d_plot}. 
We observe, in Figure~\ref{fig:DAWN_sequence_time_history}, a significant improvement, specifically, in terms of radial navigation error $\delta r_{\mathrm{RAD}}$ and cross-track error $\delta r_{\mathrm{XT}}$, when the \texttt{RelDyn} motion priors are included in the problem.
The improvements can be easily explained by considering the corrective nature of the motion prior on components of the estimated relative position which are inherently uncertain in SLAM alone.
Indeed, the baseline SLAM solution (\texttt{RelDyn} OFF), produced solely by imposing visual cue constraints, is uncertain in the boresight direction of the camera, which in our case also corresponds to the radial direction.
Furthermore, the large cross-track error of the baseline SLAM solution (\texttt{RelDyn} OFF) can be explained by the fact that the sequence images provide little change in perspective in the actual cross-track direction, leading to poor estimate correction in that direction.
In fact, as can also be seen in Figure~\ref{fig:DAWN_sequence_time_history}, we observe that this component is the most significant contributor to the overall navigation error norm.
Therefore, by constraining the solution using \texttt{RelDyn} factors, encourages a solution where the modulus of the radius changes little, due to the large distance to the asteroid during the approach phase, and where the motion of the spacecraft is quasi-planar, due to a quasi-Keplerian configuration of the orbit, we improve the error significantly in both of these directions.

\begin{figure}[ht]
	\centering
	\includegraphics[width=3.4in, clip=true, trim=0.3in 0.0in 3.3in 0.0in]{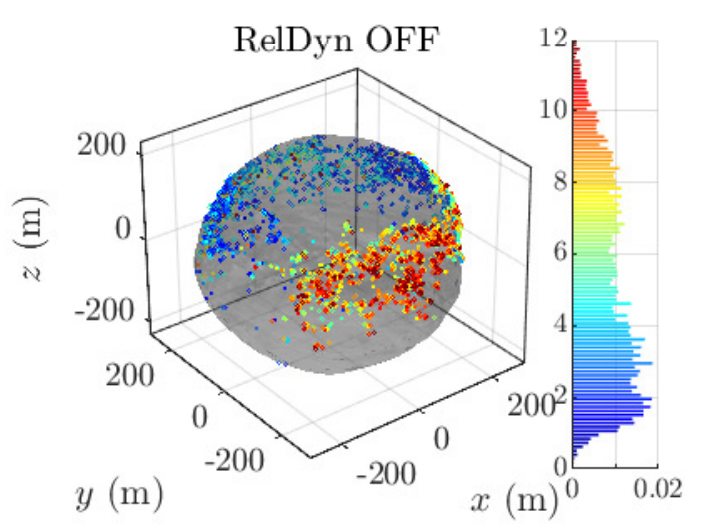}
	\includegraphics[width=3.4in, clip=true, trim=3.1in 0.0in 0.4in 0.0in]{figures/landmark_errors_DAWN_RC3.pdf}
	\caption{Estimated landmark positions and errors with respect to SPC-derived PDS shape model.}
	\label{fig:DAWN_seqence_landmarks}
\end{figure}

The landmark errors in Figure~\ref{fig:DAWN_seqence_landmarks} also show significant improvement between the case with and without the incorporation of motion priors.
It should be noted, however, that the excellent performance of the algorithm, shown in Figure~\ref{fig:DAWN_sequence_time_history}, is predicated on the low-uncertainty pre-encounter knowledge of the dynamical model parameters, including the spherical gravity term $\mu_a$, the modeled solar radiation pressure force $\vec{F}_{\mathrm{SRP}}$, as well as the initial spin state vector $\vec{\omega}_{\fG\fI, 0}$.
In practice, these have to be estimated in-situ with additional measurement modalities.
Future work should focus on incorporating such in-situ estimation into the algorithm.


\section{Experimental validation using in-lab generated imagery}
\label{sec:valid_astros}

To perform additional validation of the proposed AstroSLAM algorithm, we use image and ground-truth data generated at the \textbf{A}utonomous \textbf{S}pacecraf\textbf{t} \textbf{R}obotic \textbf{O}perations in \textbf{S}pace (ASTROS) experimental facility~\cite{cho20095}, located at the Dynamics and Control Systems Laboratory of the Georgia Institute of Technology. 

\subsection{Experimental setup}

The ASTROS facility houses an eponymous 5 degree-of-freedom spacecraft simulator test-bed, a 7 degree-of-freedom robotic manipulator system (RMS) consisting of a Schenck\texttrademark{} linear stage and a Universal Robots\texttrademark{} UR10e robotic arm, a 12-camera VICON\texttrademark{} motion capture system, as well as a dedicated control room.


\begin{figure*}[th]
	\centering
	\begin{subfigure}[t]{0.29\textwidth}
		\centering
		\includegraphics[height=2.0in, trim=0px 0px 0px 0px, clip=true]{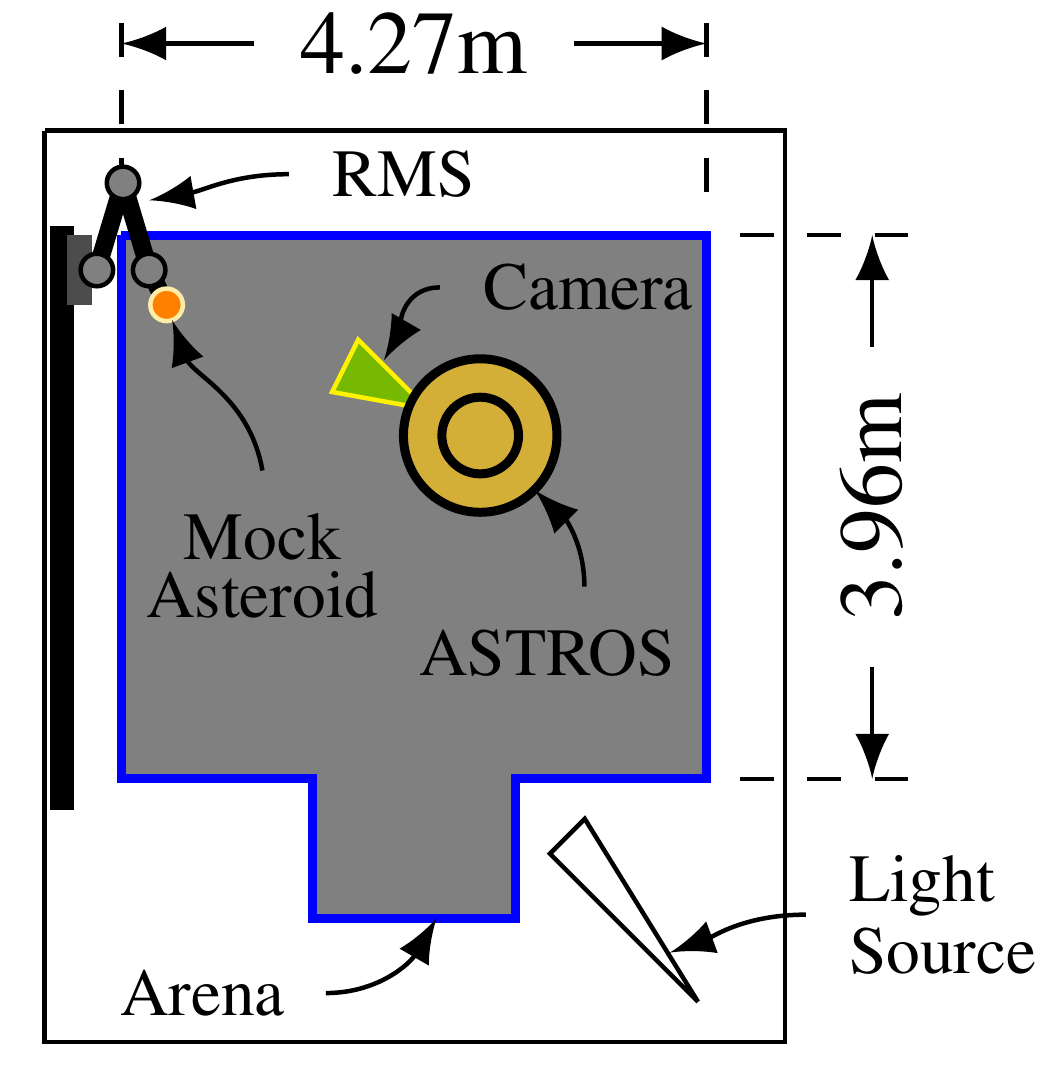}
		\caption{Disposition of Devices in Arena}
	\end{subfigure}
	\begin{subfigure}[t]{0.22\textwidth}
		\centering
		\includegraphics[height=2.0in, trim=25px 0px 0px 60px, clip=true]{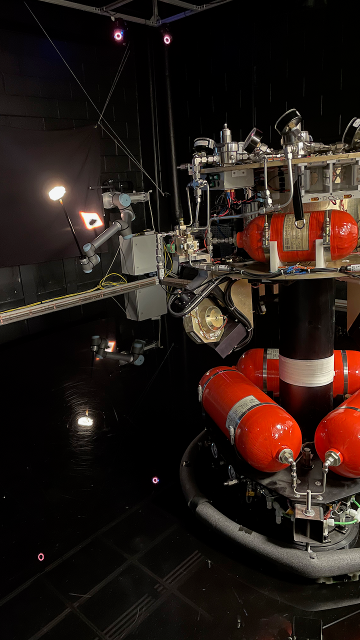}
		\caption{ASTROS in Arena}
	\end{subfigure}
	\begin{subfigure}[t]{0.46\textwidth}
		\centering
		\includegraphics[height=2.0in, trim=50px 0px 50px 0px, clip=tru]{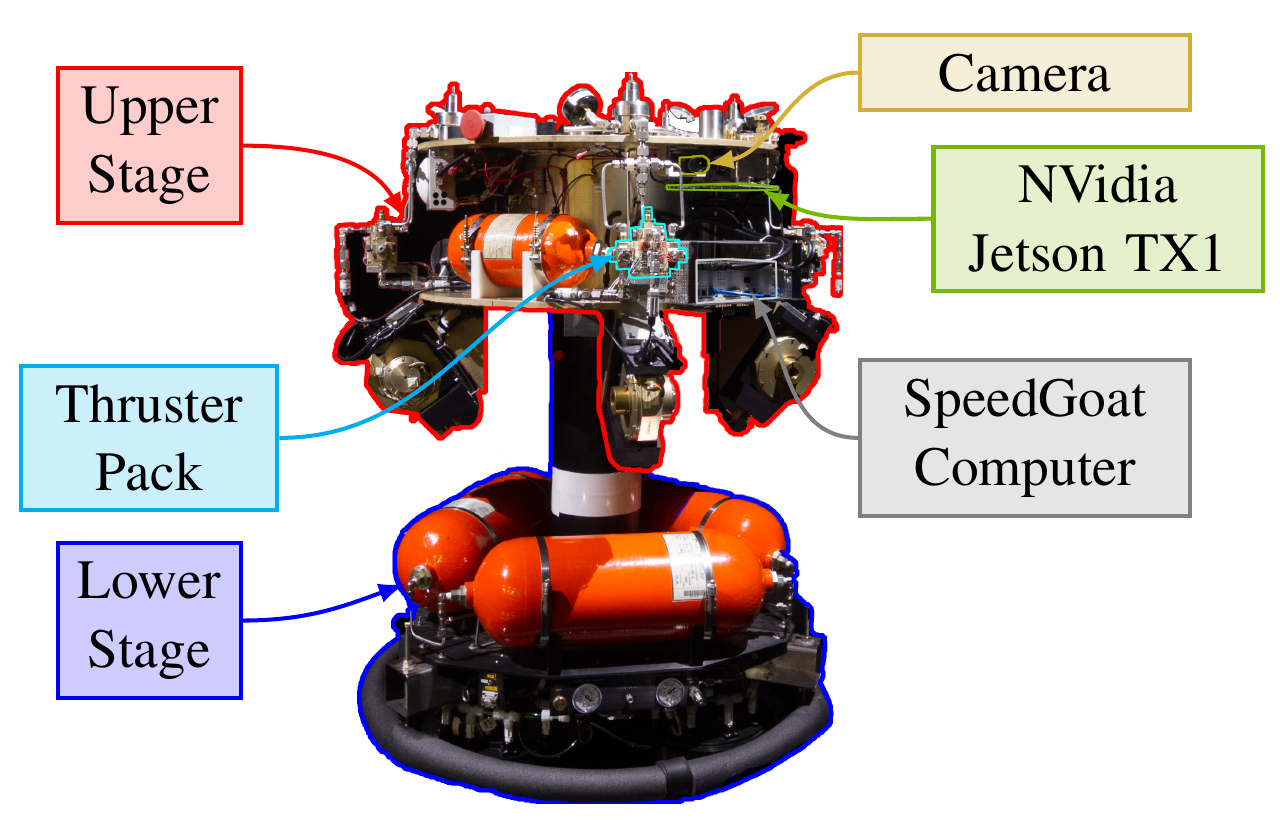}
		\caption{Main Components of ASTROS Test Bed}
	\end{subfigure}
	
	\begin{subfigure}[t]{0.37\textwidth}
		\centering
		\includegraphics[height=1.8in, trim=50px 50px 100px 48px, clip=true]{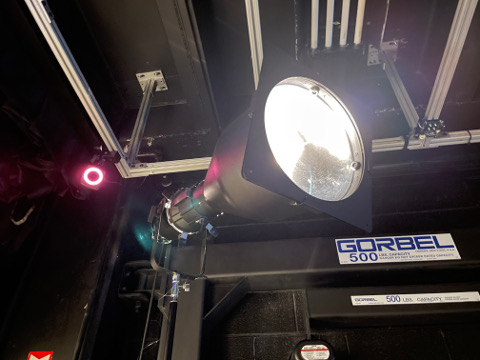}
		\caption{Collimated-Beam Light Source and~VICON Motion Capture System}
	\end{subfigure}
	\begin{subfigure}[t]{0.37\textwidth}
		\centering
		\includegraphics[height=1.8in, trim=30px 550px 760px 30px, clip=true]{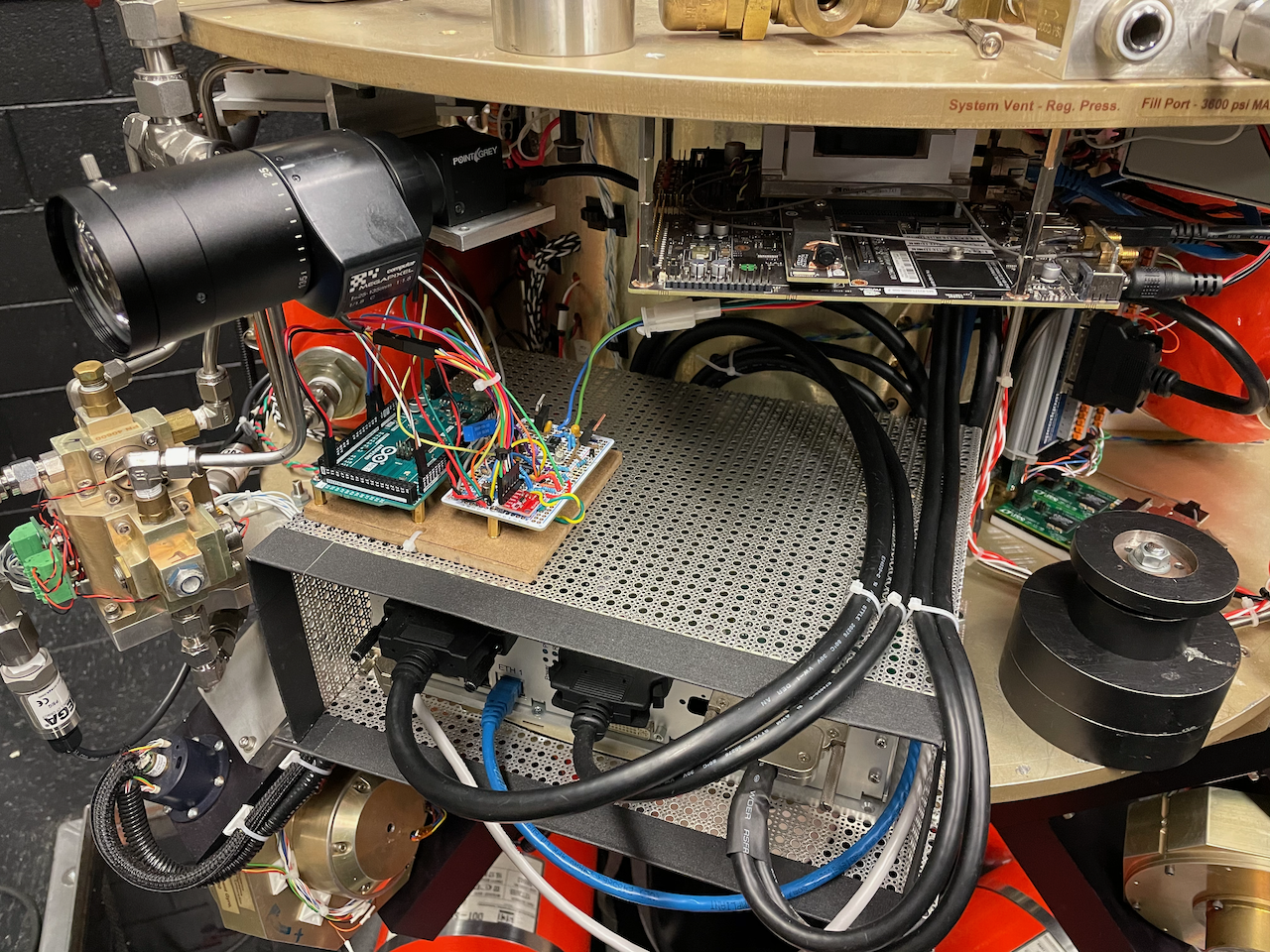}
		\caption{Small FOV camera optics}
	\end{subfigure}
	\begin{subfigure}[t]{0.24\textwidth}
		\centering
		\includegraphics[height=1.8in, trim=7px 24px 7px 30px, clip=true]{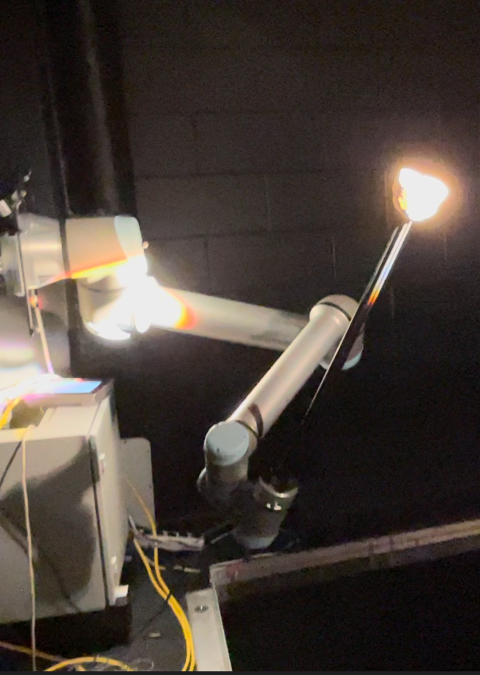}
		\caption{RMS and mock asteroid}
	\end{subfigure}
	\caption{\texttt{AstroSLAM} experiment setup.}
	\label{fig:my_label}
\end{figure*}

Mechanically, the ASTROS platform is composed of two structures, called the upper and lower stages. 
The motion of these two stages is restricted or rendered free by exploiting two pressurized-air bearing systems, allowing for frictionless motion in up to 5 degrees of freedom, 3 of which are of rotation and 2 of translation. 
A linear air-bearing system between the lower stage and the floor levitates its lower stage off the near-perfectly flat floor, providing two degrees of planar translation plus one degree of rotation (2+1 configuration). 
Additionally, a hemispherical air-bearing allows for free rotation of the upper stage around two perpendicular horizontal axes.
For the purposes of this experiment, the hemispherical joint is maintained fixed at a preset attitude, and hence the test-bed is in 2+1 configuration.

The platform is fitted with 12 cold-air gas thrusters which, when firing, generate forces and torques to allow it to actively maneuver in the test arena.
The ASTROS test-bed also possesses an inertial measurement unit and a rate gyro, which when paired with an extended Kalman filter, allow it to estimate the position, attitude, linear velocity and angular velocity of the upper stage.
The actuation of the thrusters is performed by dedicated power electronics in response to control computed on an embedded SpeedGoat\texttrademark{} computer.
The computer compiles and executes a program derived from a prototyped Simulink\texttrademark{} model incorporating sensor measurement acquisition, control computation, actuator allocation and input-output communications with devices on the platform in real-time.

A mock model asteroid is affixed to the RMS end-effector and its motion is scheduled
in open-loop control mode.
Using the linear stage and robotic arm joint encoder values only, we achieve sub-millimeter end-effector positioning error in the test arena with respect to an arbitrarily pre-defined inertial frame. 

Throughout the experiment, the motion capture system's UDP data stream was used to save the position and attitude of the ASTROS upper stage at a frequency of 100~Hz.
The frame number index of the data stream was subsequently used to synchronize and time-stamp all signal histories across the multiple devices, thus providing a single clock baseline for all the acquired data.

\subsection{Simulating and tracking an idealized trajectory}

To emulate a true unforced orbital motion, idealized trajectories were generated and tracked in the ASTROS facility test arena.
These trajectories were re-scaled via non-dimensionalization and redimensionalization to fit the physical limits of the arena and the safe allowable velocities while achieving a reasonably long segment duration and speed-to-downrange-distance ratio which emulate the real mission scenario.

Let the desired camera frame, denoted by~$\mathcal{E}$, be associated to the real camera frame~$\mathcal{C}$, and the desired ASTROS upper stage frame, denoted by~$\mathcal{D}$, be associated to the real ASTROS upper stage frame~$\fS^{\circ}$ .
The camera is fixed on the ASTROS platform's upper stage, yielding a constant rotation matrix $R_{\fS^{\circ}\fC}$ ($R_{\mathcal{D}\mathcal{E}})$ and a constant translation vector $\vec{r}_{\mathrm{C}\mathrm{S}^{\circ}}^{\fS^\circ}$ ($\vec{r}_{\mathrm{E}\mathrm{D}}^{\mathcal{D}}$), which are both estimated along with the camera intrinsic parameters by non-linear calibration.
To respect the proportions of the emulated scenario, we assume that the modelled ideal spacecraft frame~$\fS$, as defined in Section~\ref{subsec:asteroid_nav_defs}, is coincident and aligned with the experimental camera frame~$\mathcal{C}$. 
\begin{figure*}[th]
	\centering
	\includegraphics[width=6.5in, clip=true, trim=0.5in 0.2in 0.1in 0.2in]{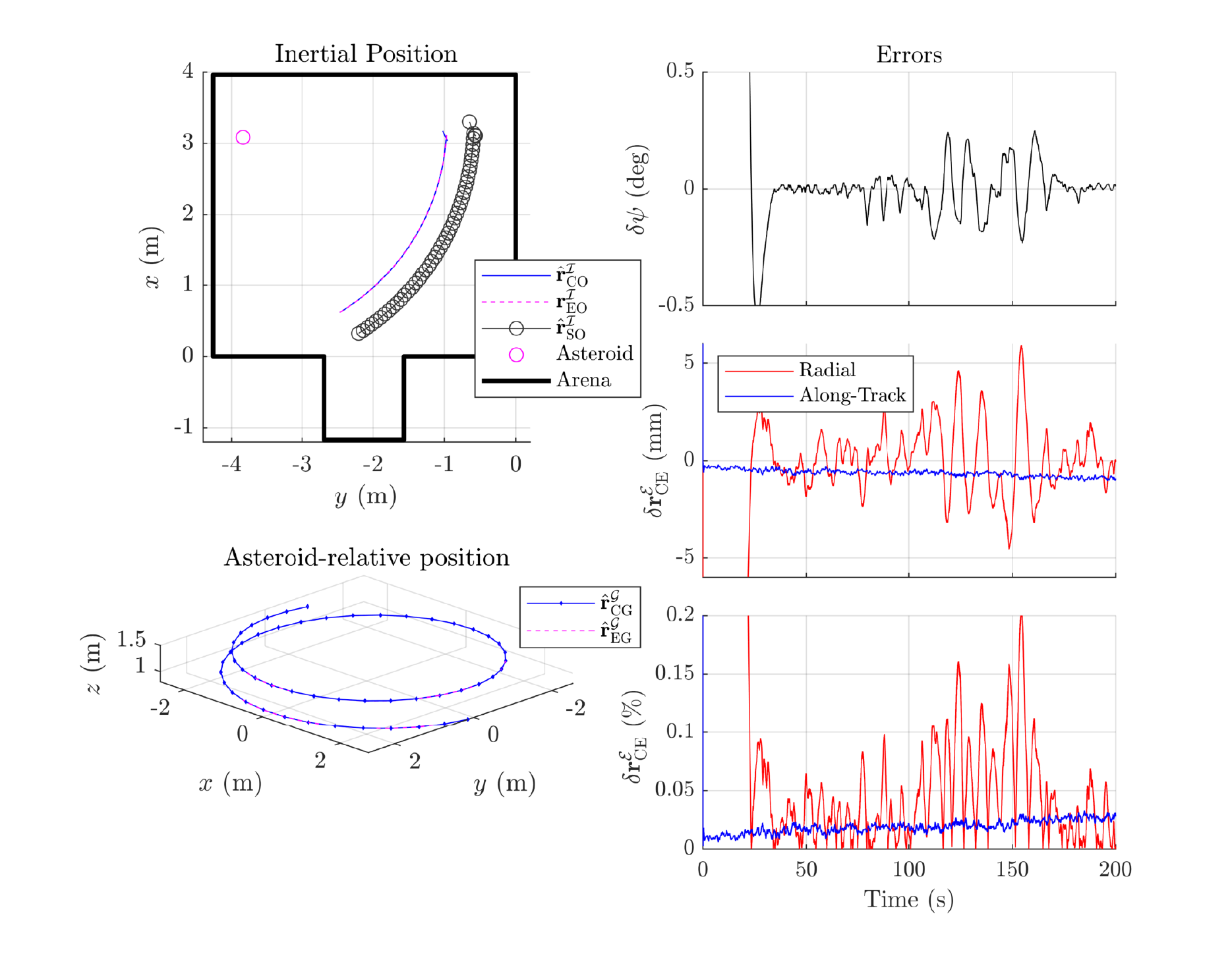}
	\caption{ASTROS lab experiment idealized trajectory and tracking errors.}
	\label{fig:idealized_traj}
\end{figure*}
We want the desired camera frame~$\mathcal{E}\triangleq \lbrace\mathrm{E}; \lbrace\uvec{e}_i\rbrace_{i=1}^{3}\rbrace$ to trace a trajectory described by the tuple $(R_{\fI\mathcal{E}}(t),\vec{r}_{\mathrm{E}\oA}^{\fI}(t), \vec{\omega}_{\mathcal{E}\fI}^{\fI}(t), \vec{v}_{\mathrm{E}\oA}^{\fI}(t)), \, t \in [t_0, t_f]$, such that
\begin{align*}
	\dvec{r}_{\mathrm{E}\oA}^{\fI}(t) &= \vec{v}_{\oE\oA}^{\fI}(t), \vec{r}_{\mathrm{E}\oA}^{\fI}(0) = \vec{r}_{\mathrm{E}\oA,0}^{\fI},\nonumber \\
	\dvec{v}_{\mathrm{E}\oA}^{\fI}(t) &= -\left(\frac{\mu_a }{\|\vec{r}_{\oE\oA}(t)\|^3} +\frac{\mu_{\odot} }{\|\vec{r}_{\oA\oI}(t)\|^3}\right) \xin{\vec{r}_{\oE\oA}}{\fI}(t) \nonumber \\
	&+ \frac{1}{m_s}\vec{F}_{\mathrm{SRP}}^{\fI}\left(\vec{r}_{\oE\oI}^{\fI}(t)\right) + \frac{1}{m_s}\vec{F}_{s}^{\fI}(t), \vec{v}_{\mathrm{E}\oA}^{\fI}(0) = \vec{v}_{\mathrm{E}\oA,0}^{\fI},
\end{align*}
is satisfied, while maintaining line-of-sight with the mock asteroid throughout the segment.
Assuming $\vec{F}_{s}^{\fI}(t)\equiv \vec 0$, we can find a solution $\vec{r}_{\mathrm{E}\oA}^{\fI}(t)$ and $\vec{v}_{\mathrm{E}\oA}^{\fI}(t)$ readily, without requiring $R_{\fI\mathcal{E}}(t)$. 
Subsequently, we pick a center pointing strategy, such that, at any time $ t \in [t_0, t_f]$, 
\begin{align*}
	\uvec{e}_3^{\fI}(t) &= {\vec{r}_{\mathrm{A}\oE}^{\fI}(t)}/{\|\vec{r}_{\mathrm{A}\oE}^{\fI}(t)\|}, \nonumber \\
	\uvec{e}_2^{\fI}(t) &= { \sk{\uvec{e}_3^{\fI}(t)}\vec{v}_{\oA\oE}^{\fI}(t) }/{\|\sk{\uvec{e}_3^{\fI}(t)}\vec{v}_{\oA\oE}^{\fI}(t)\|}, \nonumber \\
	\uvec{e}_{1}^{\fI}(t) &= \sk{\uvec{e}_2^{\fI}(t)}\uvec{e}_3^{\fI}(t),
\end{align*} 
and we construct $R_{\fI\mathcal{E}}(t) = \begin{bmatrix} \uvec{e}_{1}^{\fI}(t) &  \uvec{e}_{2}^{\fI}(t) &  \uvec{e}_{3}^{\fI}(t)\end{bmatrix}$ and $\vec{\omega}_{\mathcal{E}\fI}^{\fI}(t) = \log(\dot{R}_{\fI\mathcal{E}}(t)R_{\fI\mathcal{E}}^\top(t))^\vee$.
Finally, we obtain
\begin{align*}
	R_{\fI\mathcal{D}}(t) &= R_{\fI\mathcal{E}}(t)R_{\mathcal{E}\mathcal{D}}(t), \nonumber \\
	\vec{r}_{\mathrm{D}\oI}^{\fI}(t) &= \vec{r}_{\mathrm{E}\oA}^{\fI}(t) + \vec{r}_{\mathrm{A}\oI}^{\fI}(t) - R_{\fI\mathcal{D}}(t)\vec{r}_{\mathrm{E}\mathrm{D}}^{\mathcal{D}}(t), \nonumber \\
	\vec{\omega}_{\mathcal{D}\mathcal{I}}^{\mathcal{I}}(t) &= \vec{\omega}_{\mathcal{E}\mathcal{I}}^{\mathcal{I}}(t), \nonumber \\
	\vec{v}_{\mathrm{D}\oI}^{\fI} &= \vec{v}_{\mathrm{E}\oA}^{\fI}(t) - \sk{\vec{\omega}_{\mathcal{D}\mathcal{I}}^{\mathcal{I}}(t)}R_{\fI\mathcal{D}}(t)\vec{r}_{\mathrm{E}\mathrm{D}}^{\mathcal{D}}(t).
\end{align*}
The tuple $(R_{\fI\mathcal{D}}(t), \vec{r}_{\mathrm{D}\oI}^{\fI}(t), \vec{\omega}_{\mathcal{D}\mathcal{I}}^{\mathcal{I}}(t), \vec{v}_{\mathrm{D}\oI}^{\fI}(t)), t \in \left[t_0, t_f\right]$ constitutes the reference trajectory, which is tracked by the tuple $(R_{\fI\mathcal{S}^{\circ}}, \vec{r}_{\mathrm{S}^{\circ}\oI}^{\fI}, \vec{\omega}_{\mathcal{S}^{\circ}\mathcal{I}}^{\mathcal{I}}, \vec{v}_{\mathrm{S}^{\circ}\oI}^{\fI})$ of frame $\mathcal{S}^{\circ}$ by means of a static feedback controller.

For simplicity, we fix the position of the asteroid in the test arena, thus $\vec{r}_{\oA\oI}^{\fI}(t) = \vec{r}_{\oA\oI,0}^{\fI}$, and we rotate it at a constant angular velocity around a single body axis, hence $\vec{\omega}_{\fA\fI}^{\fA}(t) = \vec{\omega}_{\fA\fI,0}^{\fA} = \begin{bmatrix}0 & 0 & \omega_a\end{bmatrix}$, starting from some initial orientation $R_{\fI\fA,0}$.
To further simplify the planned maneuver in the ASTROS arena, we devise a planar orbital trajectory.
We restrict the motion of the test-bed to the 2+1 planar case, by fixing the rotation of the upper stage and freeing the lower stage to move along the inertial $x$-$y$ directions, and rotate around the inertial $z$ direction.
We impose that the vertical component of the asteroid's inertial position corresponds to the vertical component of the inertial position of the camera frame, expressed in inertial frame coordinates, or simply put, $\vec{r}_{\mathrm{C}\mathrm{A}}(t)\cdot \uvec{n}_{3} \equiv  0$.
This guarantees that, once the upper stage attitude is locked, the camera frame can only move in an $x$-$y$ aligned plane and that this plane always intersects the center of the asteroid, itself manipulated by the robotic arm.
It can now be assumed that the desired camera frame mimics the motion of the spacecraft frame in an emulated orbital motion.
We mention the parameters of an example tracked trajectory in Table~\ref{tab:idealized_traj}, and we present, in Figure~\ref{fig:idealized_traj}, sub-millimeter tracking error in the along-track direction and sub-5-millimeter tracking error in the radial direction.
\begin{table}[thb]
	\vspace{1em}
	\centering
	\caption{Parameters of the tracked ideal trajectory.}
	\label{tab:idealized_traj}
	\begin{tabular}{c|c}
		\toprule
		Parameter & Values \\ \midrule
		$\vec{r}_{\mathrm{E}\oI,0}^{\fI}$ & $\begin{bmatrix}3.1150 &  -0.9719 &  -1.2228 \end{bmatrix}^\top$ m  \\ \midrule 
		$\vec{v}_{\mathrm{E}\oI,0}^{\fI}$ & $\begin{bmatrix} -0.0165 & -0.0010 & 0.0000 \end{bmatrix}^\top$ m/s  \\ \midrule  
		$\vec{r}_{\mathrm{A}\oI,0}^{\fI}$ & $\begin{bmatrix} 3.083 &  -3.839 &  -1.223 \end{bmatrix}^\top$ m \\ \midrule 
		$\vec{\omega}_{\fA\fI,0}^{\fA}$ & $\begin{bmatrix} 0 & 0& \frac{4}{180}\pi \end{bmatrix}^\top$ rad \\ \midrule  
		$R_{\fI\fA,0}$ & $\begin{bmatrix}
			-0.8761 &   0.0084  & -0.4820\\
			-0.1041 &   0.9730  &  0.2061\\
			0.4707 &   0.2308 &  -0.8516\end{bmatrix}$ \\ \midrule 
		$\vec{r}_{\mathrm{E}\mathrm{D}}^{\mathcal{D}}$ & $\begin{bmatrix}0.3744  & -0.1278  & -0.1898\end{bmatrix}^\top$ \\ \midrule 
		$ R_{\mathcal{D}\mathcal{E},0}$ & $\begin{bmatrix} 0.1088 &  -0.0016  &  0.9940 \\
			0.9940 &   0.0038 &  -0.1088\\
			-0.0036  &  1.0000  &  0.0020\end{bmatrix}$ \\ \bottomrule
	\end{tabular}
\end{table}

To achieve the level of precision required to simulate orbital motion, all of the parameters relevant to the experiment were estimated using accurate calibration schemes. 
Specifically, we carried out the accurate determination of the asteroid mounting boom length, the estimation of the RMS home position and attitude, the simultaneous calibration of the camera position and attitude relative to the upper stage and of the camera intrinsic parameters, using 2D-to-3D correspondences induced by taking images of a known 3D calibration target.

\subsection{Lab experiment image dataset}

To capture images, we used a Teledyne FLIR\texttrademark{} Flea3 visible-spectrum global-shutter camera along with a MegaPixel 25-135~mm tele-objective lens.
Set at around 100~mm focal length, the lens produces a field-of-view angle of about 5\textdegree~to mimic the navigation camera of a typical asteroid surveying mission.
The size of the mock asteroid and working distance were chosen accordingly to produce an apparent size of the mock asteroid in the image corresponding to 700-800 pixels in the horizontal direction.
An on-board NVidia TX1 computer acquired images of the mock asteroid as the ASTROS platform maneuvered in the arena.

\begin{figure*}[bht]
	\centering
	\includegraphics[width=2.1in]{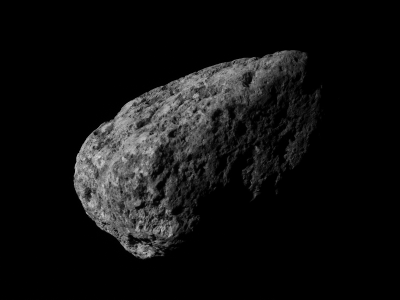}
	\includegraphics[width=2.1in]{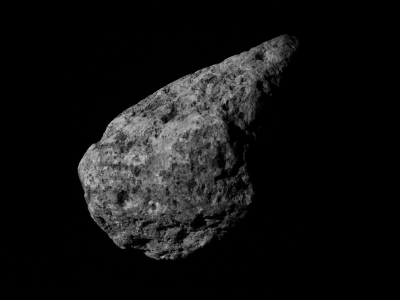}
	\includegraphics[width=2.1in]{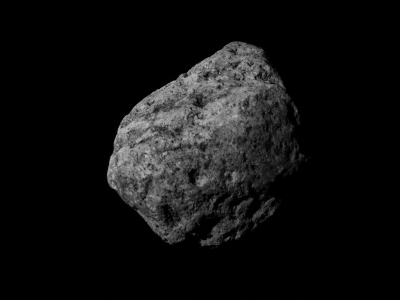}\\[0.02in]
	\includegraphics[width=2.1in]{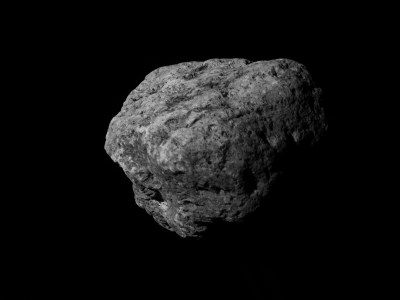}
	\includegraphics[width=2.1in]{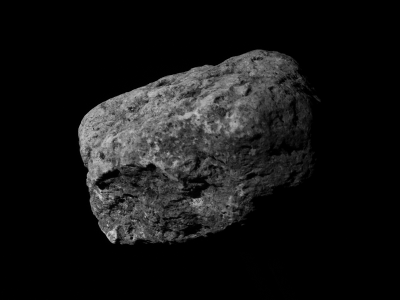}
	\includegraphics[width=2.1in]{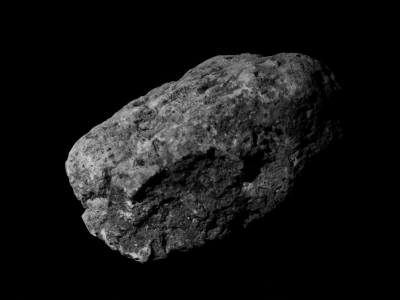}
	\caption{Captured images from the ASTROS experiment sequence.}
	\label{fig:in_lab_imagery}
\end{figure*}

To emulate space-like lighting, typically characterized by collimated light rays arriving from a source infinitely far away, a tight-beam stage lighting source was used.
The light source, a Source 4\texttrademark{} Ellipsoidal with a 5\textdegree~beam angle constrained by dedicated optics, illuminates the target mock asteroid throughout the experiment.
Note that the light source is fixed inertially.
This mimics the scenario in space where, for the short duration of the navigation segment, there is negligeable angular change in sunlight direction, when viewed inertially.
Although the challenge of light back-scattering still persists due to the presence of atmosphere in the facility, the tight-beam light source produces very crisp and harsh shadowing in the captured images.
A sample of the image sequence captured in the ASTROS experiment can be viewed in Figure~\ref{fig:in_lab_imagery}.

\subsection{Quantitative evaluation results}
\label{sec:eval_map}

\begin{figure}[htb]
	\centering
	\includegraphics[width=3.3in, clip=true, trim=0.3in 0.3in 0.3in 0.3in]{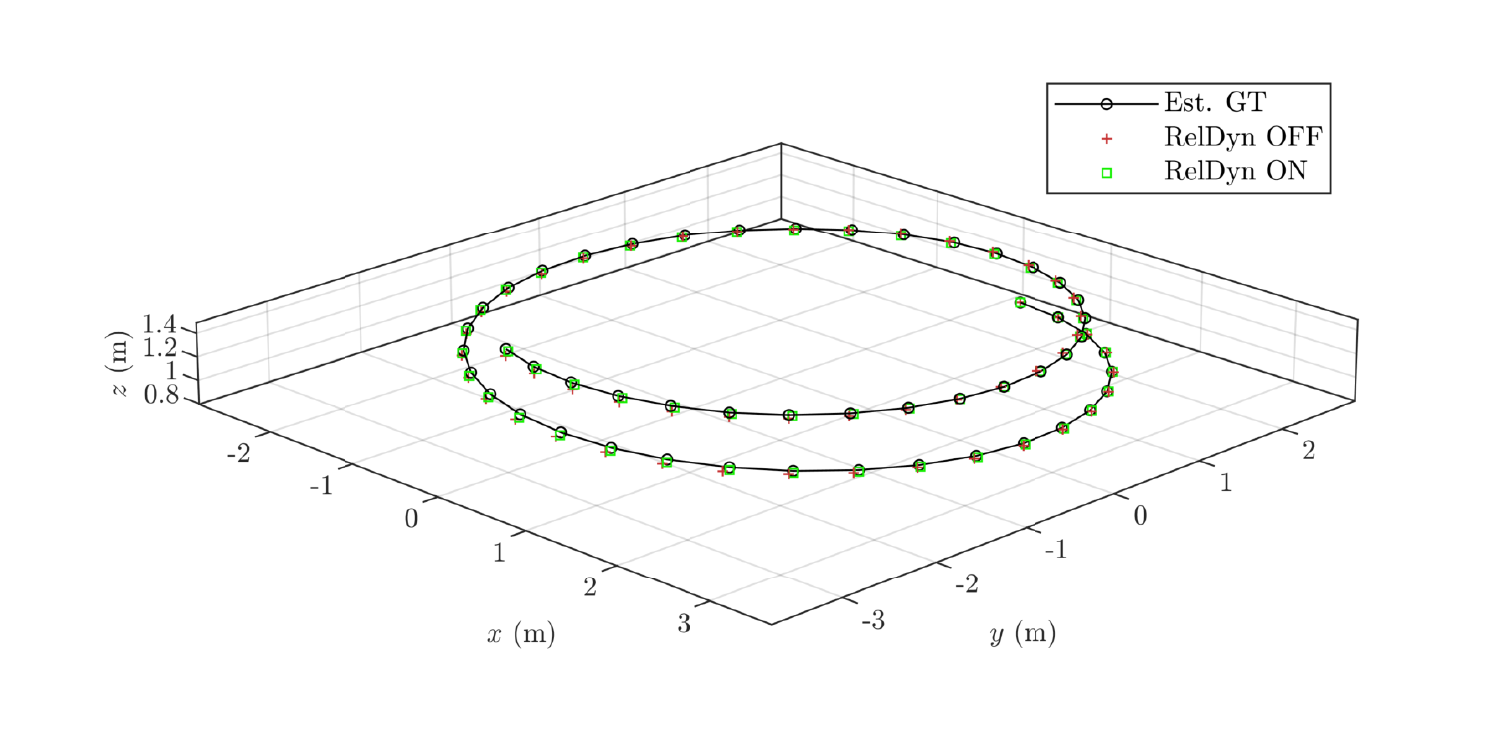}
	\caption{3D trajectory estimate vs estimated ground truth for ASTROS sequence.}
	\label{fig:ASTROS_sequence_3d_plot}
\end{figure}

\begin{figure}[htb]
	\centering
	\includegraphics[width=3.3in, clip=true, trim=0.4in 0.0em 0.1in 0.6em]{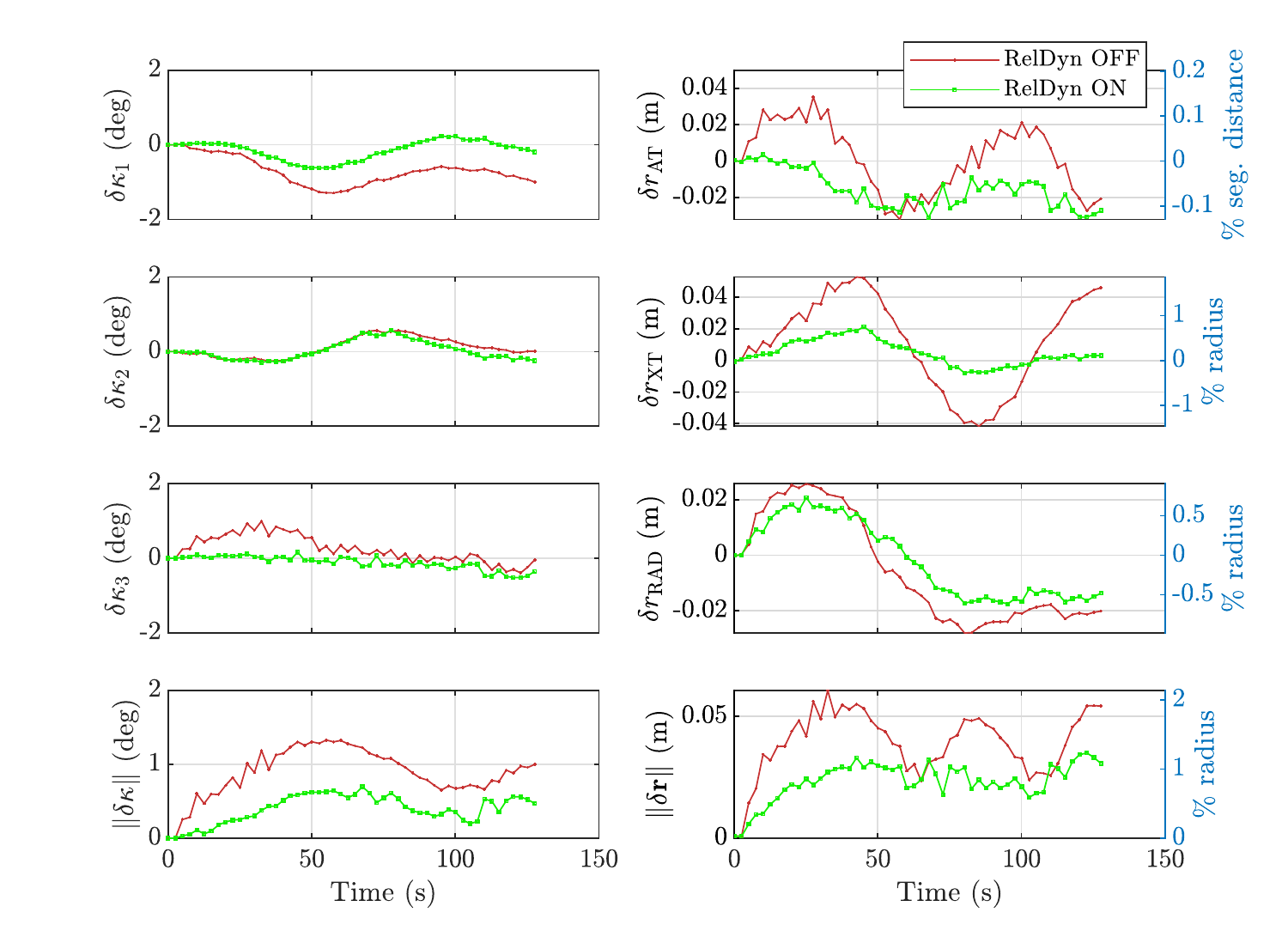}
	\caption{Estimated Pose Errors with respect to Estimated Ground Truth for ASTROS Sequence.}
	\label{fig:ASTROS_sequence_time_history}
\end{figure}

\begin{figure}[htb]
	\centering
	\includegraphics[width=3.3in, clip=true,trim=0.4in 3.2em 3.1in 0.5in]{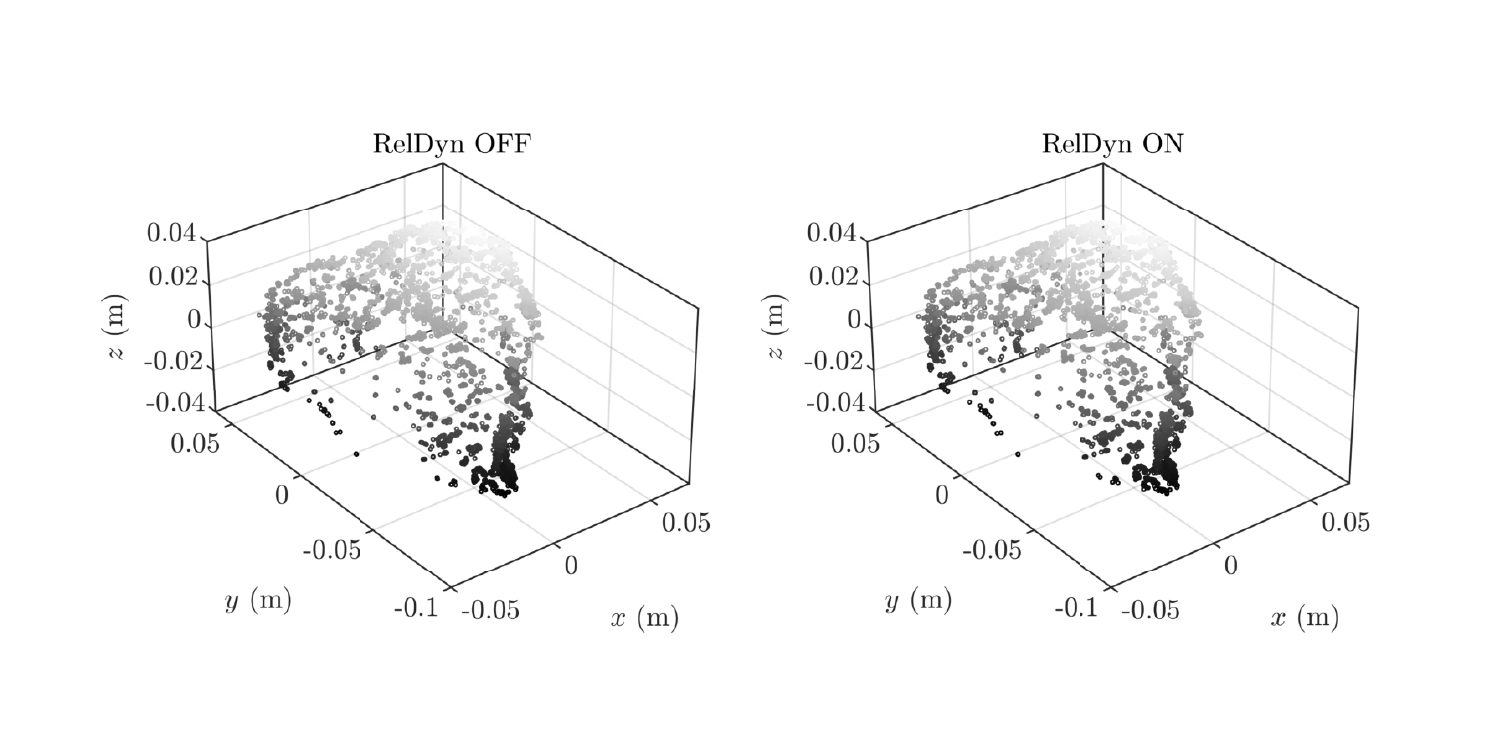}\\
	\includegraphics[width=3.3in, clip=true,trim=3.0in 3.5em 0.4in 0.5in]{figures/landmark_positions_lab_experiment_20211118T205341.pdf}
	\caption{Estimated Landmark Positions for ASTROS Experiment.}
	\label{fig:ASTROS_sequence_landmarks}
\end{figure}

We visualize in Figure~\ref{fig:ASTROS_sequence_3d_plot} the trajectory of the \texttt{AstroSLAM}-estimated 3D position $\tilde{\vec{r}}_{\mathrm{S}\mathrm{A}, k}^{\fA}, \, k=1,\ldots,N$, as derived from the  solution $\lbrace \tilde{T}_{ \fA_{k}\fS_{k} }\rbrace_{k=0}^{N}$ and relating to the idealized simulated spacecraft frame $\fS_k, \, k=1,\ldots,N$, overlayed on top of the trajectory of the EKF-estimated 3D position $\hat{\vec{r}}_{\mathrm{C}\mathrm{A},k}^{\fA}$, relating to the ASTROS camera frame.
We thus assume that the estimated pose $\lbrace \hat{T}_{ \mathcal{A}_k\mathcal{C}_k }\rbrace_{k=0}^{N}$ of the experimental camera corresponds to the ground truth simulated spacecraft pose $\lbrace T_{ \mathcal{A}_k\mathcal{S}_k }\rbrace_{k=0}^{N}$, allowing us to compare the \texttt{AstroSLAM} solution to the millimeter precision ground truth estimate.
The procedure used here to obtain the quantitative evaluation of the trajectory error is similar to the one detailed in Section~\ref{sec:Section6.2}.
We note that, as shown in Figure~\ref{fig:ASTROS_sequence_time_history}, when the \texttt{RelDyn} factors are included, the navigation errors are significantly improved in the cross track component of the position, and to a lesser extent, in the radial component of the position.
Finally, Figure~\ref{fig:ASTROS_sequence_landmarks}
illustrates the obtained map of the landmarks for the ASTROS sequence, demonstrating impressive reconstruction ability.

\begin{figure}[htb]
	\centering
	\includegraphics[width=3.3in, clip=true, trim=0.3in 0.0in 0.5in 0.0in]{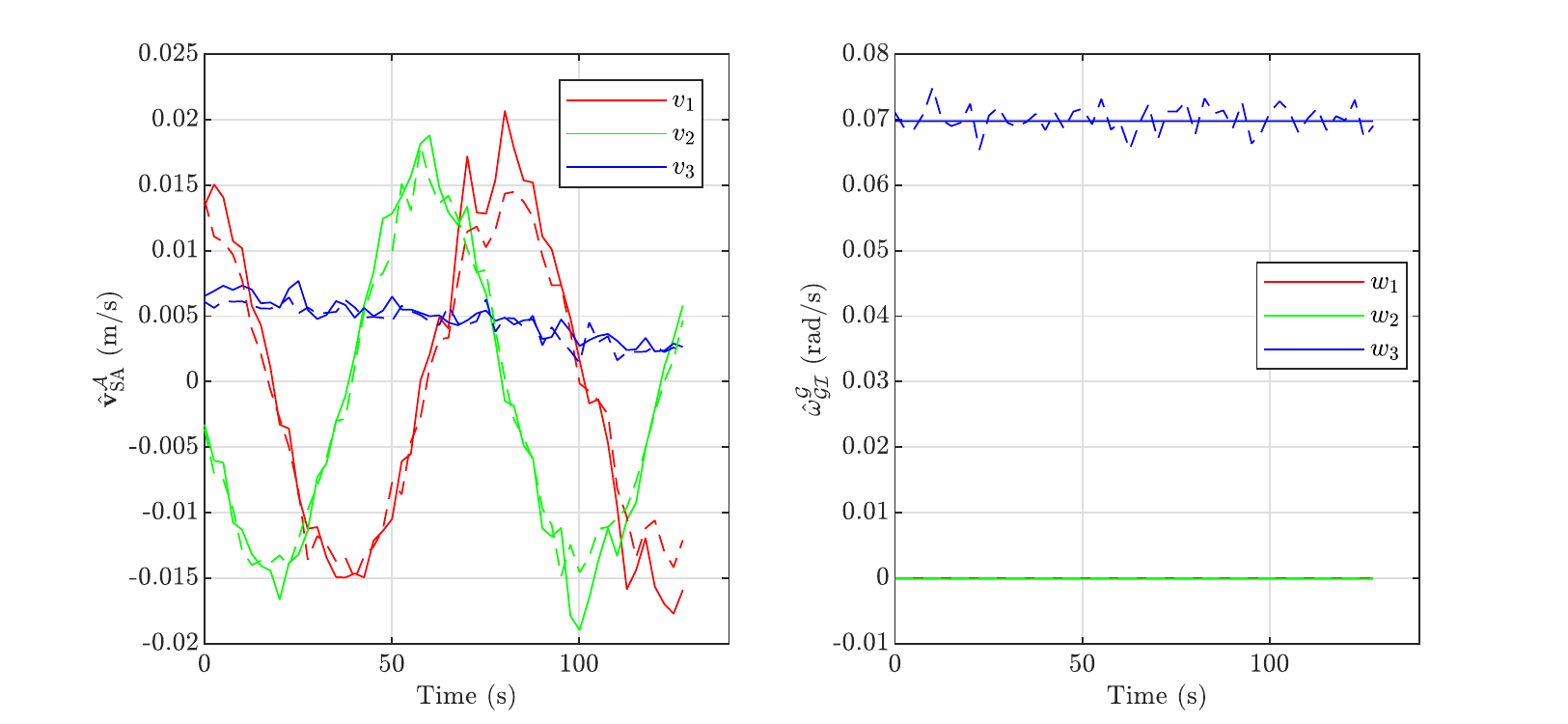}
	\caption{Relative velocity estimates vs estimated ground truth for ASTROS sequence.}
	\label{fig:ASTROS_sequence_velocities}
\end{figure}

Finally, the estimated pose errors with respect to estimated ground truth,
as illustrated by the green overlay in Figure~\ref{fig:ASTROS_sequence_time_history}, indicate excellent performance.

\section{Conclusions}
\label{sec:conclusion}

A comprehensive  vision-based  relative navigation solution, called \texttt{AstroSLAM}, is proposed for the motion of a spacecraft in the vicinity of a celestial small-body.
\texttt{AstroSLAM} solves for the navigation solution of a spacecraft under motion in the vicinity of a small body by exploiting monocular SLAM, sensor fusion and relative motion priors.
The developed motion priors are based on the dynamics of the spacecraft-small-body-Sun system, incorporating realistic perturbing effects, which affect the motion of the spacecraft in a non-negligible manner.
We show that the appropriate inclusion of noise terms in the stochastic modelling can impact the smoothability of the state.
The algorithm utilizes the factor graph formalism to cast the vision-based navigation problem as a SLAM smoothing problem that is solved efficiently using the \texttt{iSAM2} solver and the \texttt{GTSAM} library.
The factor graph approach allows the incorporation of asynchronous measurements of diverse modalities, as well as the inclusion of kinematic and dynamic constraints, thus explicitly specifying the structure of the likelihood function.
The algorithm was tested and its performance  was validated using both real imagery and trajectory data sequence pertaining to the DAWN mission and in a controlled lab environment. 
For the DAWN mission, the results demonstrate a good baseline performance of \texttt{AstroSLAM} in a typical real-world mission scenario and also show a significant improvement in terms of navigation error and landmark map reconstruction with the additional incorporation of the \texttt{RelDyn} motion priors in the estimation problem.
The 
\texttt{AstroSLAM} algorithm was also tested against imagery and data generated using the ASTROS spacecraft simulation facility at Georgia Tech by emulating realistic lighting and motion conditions.
The results of the in-lab validation further support the claim that appropriate modelling of the forces affecting the spacecraft is key to exploiting the motion priors for corrective and smoothing effects once these are incorporated into the estimation problem.
The performances demonstrated are predicated on good pre-encounter knowledge of the dynamical model parameters, such as the small-body gravitational potential, the modelling of the solar radiation pressure forces and the initial spin state of the small-body.
A natural future work direction would therefore be to augment \texttt{AstroSLAM} to tackle the in-situ estimation of these parameters by incorporating additional measurement modalities, something that is easy to do using the factor graph framework.

\section{Acknowledgments}
This work was supported by the Early Stage Innovations (ESI) grant award 80NSSC18K0251 sponsored by the U.S. National Aeronautics and Space Administration (NASA).
The authors would like to thank the following individuals:  
Katherine Skinner for contributions in the implementation of the algorithm, Frank Dellaert for early discussions, as well as Andrew Liounis and Joshua Lyzhoft of the NASA Goddard Space Flight Center for many invaluable discussions, comments, and suggestions.

\bibliographystyle{apalike}
\bibliography{ms.bib}

\appendix 
\section*{Appendix}

\section{Equations of motion} \label{sec:EoM}
\setcounter{equation}{0}
\renewcommand{\theequation}{\Alph{section}.\arabic{equation}}

In this section the rotation and translation kinematics of the spacecraft relative to the asteroid are first derived, as expressed in the arbitrarily chosen frame $\mathcal{G}$. 
Since the frame $\fI$ is inertially fixed, and the frames $\fA$, $\fG$, $\fS$ are rotating, with associated inertial angular velocity vectors $\vec{\omega}_{\fA\fI}$, $\vec{\omega}_{\fG\fI}$, and $\vec{\omega}_{\fS\fI}$ and since frame $\fG$ is an asteroid body-fixed frame, it follows that $\vec{\omega}_{\fG\fA} = \vec{0}$, and thus $\vec{\omega}_{\fG\fI} = \cancelto{0}{\vec{\omega}_{\fG\fA}} \quad + \vec{\omega}_{\fA\fI} =  \vec{\omega}_{\fA\fI}$.
The relative angular velocity between the spacecraft frame $\fS$ and frame $\fG$ is $\vec{\omega}_{\fS\fG} = \vec{\omega}_{\fS\fI} - \vec{\omega}_{\fG\fI}$.
The orientation of the frame $\fS$ relative to the frame $\fG$ is encoded in the rotation matrix $R_{\fG\fS} \in \mathrm{SO}(3)$, which satisfies the kinematic relationship
\begin{align}
	\dot{R}_{\fG\fS} = \sk{\xin{\vec{\omega}_{\fS\fG}}{\fG}}R_{\fG\fS},
	\label{eq:relativeRotation_kinematics1}
\end{align}
where $\xin{\vec{\omega}_{\fS\fG}}{\fG} \in \mathbb{R}^3$ is the relative angular velocity, expressed in the $\fG$ frame. Rewriting (\ref{eq:relativeRotation_kinematics1}) in terms of the frame $\fS$ and frame $\fG$ angular velocities and using Fact~\ref{fact1} from Section~\ref{sec:notation}, we obtain
\begin{align}
	\dot{R}_{\fG\fS} &= \sk{R_{\fG\fS}\xin{\vec{\omega}_{\fS\fI}}{\fS} - \xin{\vec{\omega}_{\fG\fI}}{\fG}}R_{\fG\fS} \nonumber \\
	&= R_{\fG\fS}\sk{\xin{\vec{\omega}_{\fS\fI}}{\fS}} - \sk{\xin{\vec{\omega}_{\fG\fI}}{\fG}}R_{\fG\fS}.
	\label{eq:relativeRotation_kinematics2}
\end{align}

Hereafter, we distinguish the inertial \textit{relative velocity vector}, denoted $\vec{v}_{\oS\oG} \triangleq \fddt{\fI}{\vec{r}_{\oS\oG}}$, from the body-fixed relative velocity vector, in turn denoted $\dvec{r}_{\oS\oG} \triangleq \fddt{\fG}{\vec{r}_{\oS\oG}}$.
From the kinematic transport theorem~\cite{schaub2003analytical}, we obtain the relationship between the aforementioned velocities, given as $\fddt{\fG}{\vec{r}_{\oS\oG}} = \fddt{\fI}{\vec{r}_{\oS\oG}} - \vec{\omega}_{\fG\fI} \times \vec{r}_{\oS\oG}$, which, once expressed in the $\fG$-frame coordinates, yields
\begin{align}
	\xin{\dvec{r}_{\oS\oG}}{\fG} = \xin{\vec{v}_{\oS\oG}}{\fG} - \sk{\xin{\vec{\omega}_{\fG\fI}}{\fG}} \xin{\vec{r}_{\oS\oG}}{\fG}.
	\label{eq:relativeVelocity_in_G}
\end{align}
In the same fashion, the inertial \textit{relative acceleration} is denoted $\vec{a}_{\oS\oG} \triangleq \fddt{\fI}{\vec{v}_{\oS\oG}}$, while the relative acceleration as seen by the $\fG$ frame is denoted $\dvec{v}_{\oS\oG} \triangleq \fddt{\fG}{\vec{v}_{\oS\oG}}$.
We obtain the relationship between the two relative accelerations, as
\begin{align}
	\fddt{\fG}{\vec{v}_{\oS\oG}} = \fddt{\fI}{\vec{v}_{\oS\oG}} - \vec{\omega}_{\fG\fI} \times \vec{v}_{\oS\oG}.
	\label{eq:transTheo_v_SG}
\end{align}
To later exploit knowledge of the relative orbital dynamics in the spacecraft-asteroid-sun system, we decompose the relative position vector $\vec{r}_{\oS\oG}$ using the intermediary point $\oA$ such that $\vec{r}_{\oS\oG} = \vec{r}_{\oS\oA} + \vec{r}_{\oA\oG}$.
Taking the derivative in the inertial frame, and then using the transport theorem, yields
\begin{align}
	\vec{v}_{\oS\oG} &= \fddt{\fI}{\vec{r}_{\oS\oA}} + \fddt{\fI}{\vec{r}_{\oA\oG}} \nonumber \\
	&= \vec{v}_{\oS\oA} + \cancelto{0}{\fddt{\fG}{\vec{r}_{\oA\oG}}} + \vec{\omega}_{\fG\fI}\times \vec{r}_{\oA\oG}.
	\label{eq:relativeVelocity_decomp}
\end{align}
Substituting equation (\ref{eq:relativeVelocity_decomp}) into equation (\ref{eq:transTheo_v_SG}), we obtain the relative acceleration as seen in the $\fG$-frame, as given by
\begin{align}
	\dvec{v}_{\oS\oG} &= \fddt{\fI}{\vec{v}_{\oS\oA}}  + \fddt{\fI}{\vec{\omega}_{\fG\fI}}\times \vec{r}_{\oA\oG} \nonumber \\
	&+ \vec{\omega}_{\fG\fI}\times \fddt{\fI}{\vec{r}_{\oA\oG}} - \vec{\omega}_{\fG\fI} \times \vec{v}_{\oS\oG} \nonumber \\
	&= \vec{a}_{\oS\oA} + \vec{\alpha}_{\fG\fI}\times \vec{r}_{\oA\oG} \nonumber \\
	&+ \vec{\omega}_{\fG\fI}\times \left(\vec{\omega}_{\fG\fI}\times \vec{r}_{\oA\oG}\right) - \vec{\omega}_{\fG\fI} \times \vec{v}_{\oS\oG},
	\label{eq:relativeAcceleration_in_G}
\end{align}
where $\vec{\alpha}_{\fG\fI} = \fddt{\fI}{\vec{\omega}_{\fG\fI}}$. 
The relative acceleration as seen by the $\fG$ frame and developed in equation (\ref{eq:relativeAcceleration_in_G}) can now be expressed in the $\fG$-frame coordinates to produce 
\begin{align}
	\xin{\dvec{v}_{\oS\oG}}{\fG} &= \xin{\vec{a}_{\oS\oA}}{\fG} + \left(\sk{\xin{\vec{\omega}_{\fG\fI}}{\fG}}\sk{\xin{\vec{\omega}_{\fG\fI}}{\fG}} + \sk{\xin{\dvec{\omega}_{\fG\fI}}{\fG}}\right) \xin{\vec{r}_{\oA\oG}}{\fG} \nonumber \\
	&- \sk{\xin{\vec{\omega}_{\fG\fI}}{\fG}}  \xin{\vec{v}_{\oS\oG}}{\fG}.
	\label{eq:relativeAcceleration_in_G_2}
\end{align}

Next, we derive the absolute rotation dynamics of the asteroid and of the spacecraft. 
Let the angular momentum of the asteroid around its center of mass be ${}^{\oA}_{}\xin{{\vec{H}}}{\fI}_a \triangleq {}^{\oA}_{}\xin{J}{\fI}_a \xin{\vec{\omega}_{\fA\fI}}{\fI}$, where ${}^{\oA}_{}{J}_a^\fI$ is the asteroid inertia matrix about the center of mass point $\oA$ expressed in $\fI$  frame coordinates. 
Using the rotation $R_{\fI\fA} \in \mathrm{SO}(3)$, we recover the inertia matrix expressed in the asteroid frame $\fA$ as ${}^{\oA}_{}\xin{J}{\fI}_a  = R_{\fI\fA}{}^{\oA}_{}\xin{J}{\fA}_a R_{\fA\fI}$
yielding ${}^{\oA}_{}\xin{{\vec{H}}}{\fI}_a = R_{\fI\fA}{}^{\oA}_{}\xin{J}{\fA}_a \xin{\vec{\omega}_{\fA\fI}}{\fA}$. Taking the time derivative of the angular momentum, we obtain 
\begin{align}
	{}^{\oA}_{}\xin{\dot{\vec{H}}}{\fI}_a &= R_{\fI\fA}\sk{\xin{\vec{\omega}_{\fA\fI}}{\fA}}{}^{\oA}_{}\xin{J}{\fA}_a \xin{\vec{\omega}_{\fA\fI}}{\fA} \nonumber \\
	&+ R_{\fI\fA}{}^{\oA}_{}\xin{\dot{J}}{\fA}_a \xin{\vec{\omega}_{\fA\fI}}{\fA} + R_{\fI\fA}{}^{\oA}_{}\xin{J}{\fA}_a \xin{\dvec{\omega}_{\fA\fI}}{\fA}.
	\label{eq:timeDeriv_AstAngularMomentum}
\end{align}
We assume that the asteroid behaves as a rigid body, and thus ${}^{\oA}_{}\xin{\dot{J}}{\fA}_a = 0_{3\times 3}$. 
Additionally, from Euler's angular momentum equation we have that ${}^{\oA}_{}\xin{\dot{\vec{H}}}{\fI}_a = \xin{\vec{\tau}_{a}}{\fI}$, where $\xin{\vec{\tau}_{a}}{\fI}$ is the sum of external torques applied on the asteroid. Left multiplying by $R_{\fA\fI}$ and then isolating the angular acceleration in equation \eqref{eq:timeDeriv_AstAngularMomentum} yields the classical result
\begin{align}
	\xin{\dvec{\omega}_{\fA\fI}}{\fA} = \left( {}^{\oA}_{}\xin{J}{\fA}_a \right)^{-1} \left( \xin{\vec{\tau}_{a}}{\fA} - \sk{\xin{\vec{\omega}_{\fA\fI}}{\fA}}{}^{\oA}_{}\xin{J}{\fA}_a \xin{\vec{\omega}_{\fA\fI}}{\fA}\right).
	\label{eq:newtoneuler_asteroid}
\end{align}
Substituting $\xin{\vec{\omega}_{\fA\fI}}{\fA} = R_{\fA\fG}\xin{\vec{\omega}_{\fG\fI}}{\fG}$ and $\xin{\dvec{\omega}_{\fA\fI}}{\fA} = R_{\fA\fG}\xin{\dvec{\omega}_{\fG\fI}}{\fG}$ in equation (\ref{eq:newtoneuler_asteroid}) and using Fact \ref{fact1}, we get
\begin{align}
	\xin{\dvec{\omega}_{\fG\fI}}{\fG} &= R_{\fG\fA}\left( {}^{\oA}_{}\xin{J}{\fA}_a \right)^{-1} R_{\fA\fG} \nonumber \\ 
	&\times \left(\xin{\vec{\tau}_{a}}{\fG} -  \sk{\xin{\vec{\omega}_{\fG\fI}}{\fG}}R_{\fG\fA}{}^{\oA}_{}\xin{J}{\fA}_a R_{\fA\fG}\xin{\vec{\omega}_{\fG\fI}}{\fG} \right).
	\label{eq:newtoneuler_asteroid_in_G}
\end{align}

Finally, we derive the translational dynamics of the spacecraft relative to the asteroid. 
We assume herein that the spacecraft is subjected to the gravitation force of the asteroid, denoted by $\vec{F}_{sa}$, the gravitation of the Sun, denoted by $\vec{F}_{s\odot}$, the solar radiation pressure (SRP), denoted by $\vec{F}_{\mathrm{SRP}}$, as well as the spacecraft actuation thrust force, denoted by  $\vec{F}_{s}$.
Assuming that the spacecraft's known mass, denoted by $m_s$, is fixed, the linear acceleration of the spacecraft with respect to the Sun's origin $\oI$, is given by $	\vec{a}_{\oS\oI} = 1/m_s(\vec{F}_{sa} + \vec{F}_{s\odot} + \vec{F}_{\mathrm{SRP}} + \vec{F}_{s})$.
The asteroid's gravitation force is obtained by computing
\begin{align}
	\vec{F}_{sa} = \left.\frac{\partial U(\vec{r})}{\partial \vec{r}}\right|_{\vec{r} = \vec{r}_{\oS\oA} }
\end{align}
for an appropriate gravity field potential function $U(\vec{r})$. 
Assume that $U(\vec{r})$ is parameterized using spherical harmonics. 
Then, when the probe is relatively distant from the asteroid, the spherical term of the potential dominates, in which case the attraction is given by $\vec{F}_{sa} = -{\mu_a m_s}/{\|\vec{r}_{\oS\oA}\|^3}\vec{r}_{\oS\oA}$.
Consider the Sun as a point mass central body. Then, $\vec{F}_{s\odot} =  -{\mu_{\odot} m_s}/{\|\vec{r}_{\oS\oI}\|^3}\vec{r}_{\oS\oI}$,
where $\vec{r}_{\oS\oI}$ is the position vector of the spacecraft center of mass $\oS$ with respect to the Sun origin $\oI$. 
Assume that the solar radiation pressure is a function of its position vector with respect to the Sun, $\vec{F}_{\mathrm{SRP}} = \vec{F}_{\mathrm{SRP}}\left(\vec{r}_{\oS\oO}\right)$.

In turn, we assume that the mass of the asteroid, denoted $m_a$ is fixed, and that the only force acting on the asteroid is the Sun's gravitational force. 
Then, the linear acceleration of the asteroid center of mass with respect to the Sun's origin $\oI$, is given by $\vec{a}_{\oA\oI} = 1/m_a\vec{F}_{a\odot}$, 
where $\vec{F}_{a\odot} = -{\mu_{\odot} m_a}/{\|\vec{r}_{\oA\oI}\|^3}\vec{r}_{\oA\oI}$.
The relative dynamics of the spacecraft-asteroid system are obtained computing $\vec{a}_{\oS\oA} = \vec{a}_{\oS\oI} - \vec{a}_{\oA\oI}$, leading to the relationship
\begin{align}
	\vec{a}_{\oS\oA} &= -\frac{\mu_a }{\|\vec{r}_{\oS\oA}\|^3}\vec{r}_{\oS\oA} -\frac{\mu_{\odot} }{\|\vec{r}_{\oS\oI}\|^3}\vec{r}_{\oS\oI} + \frac{1}{m_s}\vec{F}_{\mathrm{SRP}}\left(\vec{r}_{\oS\oO}\right) \nonumber \\
	&+ \frac{1}{m_s}\vec{F}_{s} + \frac{\mu_{\odot} }{\|\vec{r}_{\oA\oI}\|^3}\vec{r}_{\oA\oI}. 
	\label{eq:relativeTransDyn1}
\end{align}
Assume that $\vec{r}_{\oS\oI} \approxeq \vec{r}_{\oA\oI}$ given the very large distance between the spacecraft-asteroid system and the Sun, and rewrite equation (\ref{eq:relativeTransDyn1}) by expressing it in the $\fG$-frame and making explicit in terms of the state variables and input variables of interest, yielding
\begin{align}
	\xin{\vec{a}_{\oS\oA}}{\fG} &= -\left(\frac{\mu_a }{\|\vec{r}_{\oS\oG} - \vec{r}_{\oA\oG}\|^3} +\frac{\mu_{\odot} }{\|\vec{r}_{\oA\oI}\|^3}\right)\left(\xin{\vec{r}_{\oS\oG}}{\fG} - \xin{\vec{r}_{\oA\oG}}{\fG}\right) \nonumber \\
	&+ \frac{1}{m_s}R_{\fG\fS}R_{\fS\fI}\xin{\vec{F}_{\mathrm{SRP}}}{\fI}\left(\xin{\vec{r}_{\oA\oO}}{\fI}\right) + \frac{1}{m_s}R_{\fG\fS}\xin{\vec{F}_{s}}{\fS}.
	\label{eq:relativeTransDyn2}
\end{align}
Finally we substitute equations (\ref{eq:newtoneuler_asteroid_in_G}) and (\ref{eq:relativeTransDyn2}) into equation (\ref{eq:relativeAcceleration_in_G_2}), and rearrange terms to obtain equation (\ref{eq:relativeTransDyn2_in_G}).

\begin{strip}
	\begin{align}
		\xin{\dvec{v}_{\oS\oG}}{\fG} &= \left(\sk{\xin{\vec{\omega}_{\fG\fI}}{\fG}}\sk{\xin{\vec{\omega}_{\fG\fI}}{\fG}} + \sk{ R_{\fG\fA}\left( {}^{\oA}_{}\xin{J}{\fA}_a \right)^{-1} R_{\fA\fG}\left(\xin{\vec{\tau}_{a}}{\fG} - \sk{\xin{\vec{\omega}_{\fG\fI}}{\fG}}R_{\fG\fA}{}^{\oA}_{}\xin{J}{\fA}_a R_{\fA\fG}\xin{\vec{\omega}_{\fG\fI}}{\fG}\right)}\right) 
		\xin{\vec{r}_{\oA\oG}}{\fG} - \sk{\xin{\vec{\omega}_{\fG\fI}}{\fG}} \xin{\vec{v}_{\oS\oG}}{\fG} \nonumber \\
		&-\left(\frac{\mu_a }{\|\vec{r}_{\oS\oG} - \vec{r}_{\oA\oG}\|^3} +\frac{\mu_{\odot} }{\|\vec{r}_{\oA\oI}\|^3}\right) \left(\xin{\vec{r}_{\oS\oG}}{\fG} - \xin{\vec{r}_{\oA\oG}}{\fG}\right) + \frac{1}{m_s}R_{\fG\fS}R_{\fS\fI}\xin{\vec{F}_{\mathrm{SRP}}}{\fI}\left(\xin{\vec{r}_{\oA\oO}}{\fI}\right) + \frac{1}{m_s}R_{\fG\fS}\xin{\vec{F}_{s}}{\fS}.
		\label{eq:relativeTransDyn2_in_G}
	\end{align}
\rule{\textwidth}{0.5pt}
\end{strip}

\section{Stochastic continuous motion model}
\label{sec:EoM_Stochastic}
\setcounter{equation}{0}

Define the state tuple $x \triangleq \left(\xQ, \xw,\xr, \xv \right) \in \mathrm{SO}(3) \times \mathbb{R}^3 \times \mathbb{R}^3 \times \mathbb{R}^3 \triangleq X$, the input tuple $u \triangleq (\xR, \xs,\vec{\tau}, \vec{f}) \in \mathrm{SO}(3) \times \mathbb{R}^3 \times \mathbb{R}^3 \times \mathbb{R}^3 \triangleq U$, and the parameter tuple $p\triangleq\left(\mu_a,\vec{c},C,A\right) \in \mathbb{R}_{>0} \times \mathbb{R}^3 \times \mathrm{SO}(3) \times \mathcal{J} \triangleq P$. 
Given any two states $x_2 = \left(Q_2, \xw_2, \xr_2, \xv_2\right) \in X$ and $x_1 = \left(Q_1, \xw_1, \xr_1, \xv_1\right) \in X$, we define the error $\Delta_X(x_1,x_2) \in T_{x_1}X$ between $x_1$ and $x_2$ and centered at $x_1$ \cite{speyer2008stochastic} such that
\begin{align}
	\Delta_X (x_1,x_2) \triangleq \begin{bmatrix}
		\log{\left(\xQ_1^\top \xQ_2\right)}^\vee \\
		\xw_2 - \vec{w}_1 \\
		\xr_2 - \vec{r}_1 \\
		\xv_2 - \vec{v}_1
	\end{bmatrix}.
	\label{eq:def_Delta}
\end{align}

To establish the governing stochastic differential equations of the system~\cite{sage1971estimation}, we first define the 3-dimensional Wiener process $\vec{\varepsilon}_{*}(t),~(*=R,\vec{s},\vec{\tau},\vec{f})$, such that
\begin{align}
	\vec{\varepsilon}_{*}(t) = \int_{0}^{t} \vec{\nu}_{*}(\tau)\diff \tau, \quad \vec{\varepsilon}_{*}(0) = \vec{0},
\end{align}
with the increment $\diff \vec{\varepsilon}_{*}(t) = \vec{\varepsilon}_{*}(t+\diff t) - \vec{\varepsilon}_{*}(t) = \vec{\nu}_{*}(t)\,\diff t$
and satisfying $\E{\vec{\varepsilon}_{*}(t)} = \vec{0}$, $\E{\left(\vec{\varepsilon}_{*}(t) - \vec{\varepsilon}_{*}(\tau)\right)\left(\vec{\varepsilon}_{*}(t) - \vec{\varepsilon}_{*}(\tau)\right)^\top} = W_{*}|t - \tau|$,
which when $\tau \rightarrow t$ yields the relationship $\E{\diff \vec{\varepsilon}_{*} \diff \vec{\varepsilon}_{*}^\top}  = W_{*} \diff t $.
Note also that $\diff \vec{\varepsilon}_{*}\diff t = \vec{0}$.

Consider the equations of motion given in equations (\ref{eq:dQdt})-(\ref{eq:dvdt}) and substitute in $u$, where for $\hat{u} \triangleq (\hat{\xR}, \hat{\xs}, \hat{\vec{\tau}}, \hat{\vec{f}}) \in U$, we have
\begin{align*}
	u = \left(\hat{\xR}\Exp{\vec{\nu}_R}, \hat{\xs} + \vec{\nu}_{\vec{s}}, \hat{\vec{\tau}} + \vec{\nu}_{\vec{\tau}},  \hat{\vec{f}} + \vec{\nu}_{\vec{f}} \right).
\end{align*} 
To obtain the resulting stochastic differential equations, we evaluate the stochastic increment $\diff x(t) = \Delta_X\left(x(t), x(t+\diff t)\right)$, we use Facts~\ref{fact2}~and~\ref{fact3} from Section~\ref{sec:notation} and we separate the equations, while dropping the time dependence for readability, yielding
\begin{align}
	\diff \vec{\kappa} &= \left(\xQ^\top \diff \xQ\right)^\vee = \left(\hat{\xs} - \xQ^\top\xw \right) \diff t + \diff \vec{\varepsilon}_{\vec{s}}, \\
\diff \xw &= -  M^{-1}  \sk{\xw}  M \xw \diff t + M^{-1} \diff \vec{\varepsilon}_{\vec{\tau}}, \label{eq:stoch_dw}
\end{align}
\begin{align}
	\diff \xr &=  \left(\xv - \sk{\xw}\xr\right) \diff t,  \\
	\diff \xv &= \left( \left( \sk{\xw}\sk{\xw} -\sk{M^{-1}  \sk{\xw}  M \xw}  \right) \vec{c} - \sk{\xw}  \xv \right. \nonumber \\
	& \left. - \left(\frac{\mu_a}{\|\xr - \vec{c}\|^3} +\frac{\mu_{\odot}}{\|\vec{d}\|^3}\right)\left(\xr-\vec{c}\right)  + \xQ\hat{\xR}^\top \vec{g}\left(\vec{d}\right) + \xQ\hat{\vec{f}}  \right) \diff t \nonumber \\  &+ \xQ\sk{\hat{R}\vec{g}(\vec{d})}\diff\vec{\varepsilon}_{R} - \sk{\vec{c}}M^{-1}\diff\vec{\varepsilon}_{\vec{\tau}} + \xQ\diff\vec{\varepsilon}_{\vec{f}}.
\end{align}
By decomposition of the covariances $ W_{*} = L_{*}L_{*}^{-1}, ~ (*=R, \vec{t}, \vec{s}, \vec{\tau}, \vec{f})$, and by defining the Wiener process $\vec{\varepsilon}$, where $\E{\diff\vec{\varepsilon}(t)} = \vec{0}$ and $\E{\diff\vec{\varepsilon}(t)\diff\vec{\varepsilon}(t)^\top} = \mathrm{I}_{12} \diff t$, obtain the system of stochastic differential equations~(\ref{eq:stoch_EoMs_concise}).

\end{document}